\documentclass[11pt]{article}
\usepackage{enumerate, fullpage}
\usepackage{graphicx}
\usepackage{multirow}
\usepackage{color}
\usepackage{amsmath}
\usepackage{amsfonts}
\usepackage{amssymb}
\usepackage{mathrsfs}
\usepackage{mathptmx}
\usepackage{eucal,bm}
\usepackage{algorithm}
\usepackage{smile}

\usepackage{algorithmic}

\def\cS{{\cal S}}

\newcommand{\citep}{\cite}

\usepackage{tcolorbox}

\usepackage{tocloft}
\usepackage[colorlinks=true,linkcolor=red,urlcolor=blue,citecolor=blue]{hyperref}

\newcommand{\QED}{\ \hfill\rule[-2pt]{6pt}{12pt} \medskip}

\numberwithin{equation}{section}

\def\cO{{\cal O}}

\def\norm#1{\|#1\|}

\newcommand{\beq}{\begin{equation}}
\newcommand{\eeq}{\end{equation}}

\newcommand{\beqnr}{\begin{eqnarray}}
\newcommand{\eeqnr}{\end{eqnarray}}

\newcommand{\benum}{\begin{enumerate}}
\newcommand{\eenum}{\end{enumerate}}

\newtheorem{DE}{Definition}[section]
\newtheorem{AS}[DE]{Assumption}

\newcommand{\sg}[1]{\begin{color}{black}#1\end{color}}
\newcommand{\lxd}[1]{\begin{color}{black}#1\end{color}}

\advance \topmargin by -\headheight
\advance \topmargin by -\headsep

\evensidemargin \oddsidemargin

\begin{document}

\title{\huge  {Data-Driven  Minimax Optimization \\ with Expectation Constraints}}
\date{}
\author{Shuoguang Yang\thanks{email: sy2614@columbia.edu {\tt  }}
~~~~Xudong Li\thanks{School of Data Science, Fudan University, Shanghai 200433, China; email:  lixudong@fudan.edu.cn  {\tt  }}
~~~~Guanghui Lan\thanks{H. Milton Stewart School of Industrial and Systems Engineering, Georgia Institute
of Technology, Atlanta, GA, 30332; email:george.lan@isye.gatech.edu {\tt  }}
}

\maketitle

\vskip 0.5cm

\begin{abstract}
Attention to data-driven optimization approaches, including the well-known stochastic gradient descent method, has grown significantly  over recent decades, but data-driven constraints have rarely been  studied, because of the computational challenges of projections onto the feasible set defined by these hard constraints. In this paper, we focus on the non-smooth convex-concave stochastic  minimax regime and formulate the data-driven constraints as expectation constraints. The minimax expectation constrained problem subsumes a broad class of real-world applications, including data-driven robust optimization, {optimization with misspecification}, and Area Under the ROC Curve (AUC) maximization with fairness constraints. We propose a class of efficient primal-dual algorithms to tackle the minimax expectation constrained problem, and show that our algorithms converge at the \emph{optimal} rate of $\mathcal O(1/{\sqrt{N}})$, \sg{where $N$ is the number of iterations}. We demonstrate the practical efficiency of our algorithms by conducting numerical experiments on large-scale real-world applications.
    \end{abstract}

\section{Introduction}
Over recent years, attention to stochastic gradient-type algorithms such as stochastic gradient descent (SGD) methods have  grown tremendously, {partially because of their simplicity and extreme efficiency in handling large-scale and streaming datasets}. Of the various different problems where SGD has performed well, stochastic minimax optimization is of particular importance, taking the following form
\begin{equation*}
\min_{x \in \cX} \max_{y \in \cY}\  \EE_\omega[ f(x,y,\omega) ].
\end{equation*}
\sg{Here $\cX\subset \RR^{d_x}$ and $\cY \subset \RR^{d_y}$ are nonempty closed convex sets, $\omega$ is a random vector
whose probability distribution $P$ is supported on set $\Omega \subset \RR^{d}$, and $f:\cX\times \cY \times \Omega \to \RR$. We assume for every $(x,y)\in \cX\times \cY$, the expectation 
\[ 
\EE_\omega[f(x,y,\omega)] = \int_{\Omega} f(x,y,\omega) dP(\omega)
\]
is well-defined}. Such class of problems targets minimizing the objective with respect to $x$ while simultaneously maximizing it with respect to $y$, which, under proper assumptions, returns the desired saddle point. It has important applications in robust optimization \citep{ben2001approximate,ben2002robust,bertsimas2003robust,bertsimas2004price}, AUC maximization \citep{hanley1982meaning,ying2016stochastic}, game theory \citep{myerson1997game,nouiehed2019solving}, and generative adversarial networks (GAN) \citep{goodfellow2014generative}. In many existing studies, various iterative algorithms~\citep{bertsekas99nonlinear} consisting of \emph{projected} stochastic gradient descent/ascent steps are used to solve the  minimax optimization problem described above. The validity of these algorithms depends on a key preliminary assumption, i.e., $\cX$ and $\cY$ should be simple closed convex sets in the sense that the projections onto them can be computed easily. 
When the constraints preserve non-linear structures, such as logarithmic functions, high-order polynomials, or more complex forms \citep{danskin2012theory, soland1973optimal, pearsall1976lagrange}, the projection within SGD steps cannot be  conducted easily.

This issue is especially exacerbated in the presence of data-driven constraints, i.e.,  constraints imposed by the use of real-world data. When large-scale data are available, the decision maker (DM) could impose data-driven constraints through Sample Average Approximation (SAA).
In online streaming applications where data are received sequentially, the explicit form of the constraints are not available. 
One avenue to handle these complex constraints is to formulate them as \emph{expectation  constraints}, which preserves  substantial practical importance when adopted into existing data-driven optimization frameworks.
Unfortunately, the absence of explicit expressions of such constraints makes the projections extremely challenging.
Thus far, efficient algorithms have rarely been developed to handle expectation constraints, preventing the employment of data-driven constraints in real-world decision-making processes.

To avoid the challenges within projections, a common practice in the machine learning community is to solve a relaxed problem, i.e., treating expectation-constraints as penalties or regularization terms in the objective, which avoids  challenging projections. After assigning proper weights to the $m_1 + m_2$ penalties or regularization terms, the resulted relaxed problem is solved by classic stochastic gradient-type methods. Specifically, if $\gamma_i > 0$ and $\lambda_j >0$ are the weights associated with constraints $h_i$ and $g_j$, respectively, 
the relaxation approach considers the following stochastic optimization problem
\begin{equation}\label{prob:minmaxpenaltied}
\min_{x \in \cX} \max_{y \in \cY} \   \EE_{\omega,\xi,\zeta} \Big \{   f(x,y,\omega) + \sum_{i=1}^{m_1} \gamma_i h_i(x,\xi) - \sum_{j=1}^{m_2} \lambda_j g_j(y,\zeta)\Big\}.
\end{equation}
Often, the associated weights $\{\gamma_i\}_{i=1}^{m_1}$ and $\{ \lambda_j\}_{j=1}^{m_2}$ are selected using computational demanding hyper-parameter optimizations.  

Unfortunately, this widely adopted approach suffers from two major challenges in real-world applications. The first challenge is  hard constraint optimization, which arises in  scenarios where certain constraints must be strictly satisfied. For instance, in risk management,  the DM is often forced by law or legally binding policies to control the systematic risk to be strictly less than a certain value. Violating such requirements would lead to unaffordable losses to the DM. Unfortunately, the above relaxation approach  cannot accommodate such practical concerns. The second challenge is the heavy computational burden incurred in 
hyper-parameter optimization, referred to as the curse of dimensionality. To be specific, in
a scenario where each regularization term preserves $d$ candidate weights, 
hyper-parameter optimization requires the examination of all possible $d^{m_1 + m_2}$ combinations of $\lambda_i,\gamma_j$'s. Thus, the computational cost grows exponentially regarding the number of regularization terms, making this approach unfavorable for many real-world applications.

To  deal with the above issues explicitly, in this paper, we focus on the following expectation constrained minimax stochastic optimization problem,
\begin{equation}\label{prob:sto_minmax}
\begin{split}
 \min_{x\in \cX }  \max_{y \in \cY}  \,\, & F(x,y) = \EE_{\omega} [f(x,y,\omega)], \\
\mbox{s.t. }& \EE_{\xi_i} [h_i(x,\xi_i)] \leq 0, \text{ for } i =1,2,\cdots, m_1, \\
& \EE_{\zeta_j} [g_j(y,\zeta_j)]  \leq  0, \text{ for } j = 1,2,\cdots, m_2.
\end{split}
\end{equation}
Here $\cX\subset \RR^{d_x}$ and $\cY \subset \RR^{d_y}$ are nonempty closed convex sets, $\omega$, $\xi_i$ and $\zeta_j$ are random vectors with associated probability distributions and support sets.  We assume the projection onto sets $\cX$ and $\cY$ can be  conducted easily. For instance, $\cX$ and $\cY$ can be polyhedrons consisting of linear constraints or Euclidean balls where a closed-form projection is available.
For any given $x\in \cX$ and $y\in \cY$, we denote by $h(x,\xi) := [h_1(x,\xi_1);\cdots; h_{m_1}(x,\xi_{m_1})] \in \RR^{m_1}$ and $g(x,\zeta) := [g_1(x,\xi_1);\cdots; g_{m_2}(x,\zeta_{m_2})] \in \RR^{m_2}$ for notational convenience, and we
assume that the expectations in the objective function $F(x,y) := \EE_{\omega}[f(x,y,\omega)] \in \RR$ and constraints $H(x) := \EE_\xi [h (x,
\xi)] \in \RR^{m_1}$ and $G(y)  :=  \EE_\zeta [g(y,\zeta)] \in \RR^{m_2}$  are  well-defined and preserve finite values. In this paper, we further assume that the expected value function $F$ is convex in $x$ and concave in $y$ and expected value functions $H_i$, $i = 1,\ldots, m_1$ and $G_j$, $j = 1,\ldots, m_2$ are convex functions. 
Clearly, in contrast to the relaxation approach \eqref{prob:minmaxpenaltied}, in model \eqref{prob:sto_minmax}, we  consider the expectation-constrained problem explicitly, allowing hard-constraint optimization and simultaneously avoiding the heavy computation burden of hyper-parameter optimization. {Denote $\widetilde \cX := \left\{ x\in \cX\mid H(x)\le 0 \right\}$ and $\widetilde \cY:= \left\{
y\in \cY \mid G(y) \le 0
\right\}$.
We say $(x^*,y^*)\in \widetilde\cX \times \widetilde\cY$} is a saddle point to problem \eqref{prob:sto_minmax} if for any $x\in \widetilde \cX$ and $y \in \widetilde \cY$, it holds that
\[
F(x^*, y) \le F(x^*,y^*) \le F(x, y^*). 
\]
We assume that the set of saddle points to problem \eqref{prob:sto_minmax} is nonempty.

\subsection{Motivating Applications} \label{sec:applications}
We start with five motivating applications.

\noindent (A). \emph{Data-Driven Robust Optimization.}  {Consider a paramaterized objective $f(x,\theta)$ where $x$ is the decision variable and $\theta$ is a vector of parameters,} robust optimization is an  approach widely adopted to handle {the scenario where  $\theta$ is unknown but belongs to
an uncertainty set $\Theta$ \citep{ben2001approximate,ben2002robust,BenTalElGhaouiNemirovski2009}, and the DM aims at finding the optimal decision $x^*$ against the worst case scenario $\theta \in \Theta$.} Assuming explicit knowledge of $\Theta$, the effect of robust optimization has been studied extensively \citep{bertsimas2003robust,bertsimas2004price}.

Apart from that, in a wide class of  real-world applications, the DM can construct data-driven uncertainty sets by 
imposing additional data-driven constraints, whose explicit forms are not available but could be reformulated in the form of expectations. In particular, the DM may consider the following,
\begin{equation}\label{prob:robust}
\begin{split}
\min_{x \in \cX} \max_{\theta \in \Theta}  \,\,\,\, f(x,\theta), 
\mbox{ s.t. }\EE_{\zeta_i}[ g_i(\theta,\zeta_i)] \leq 0, \ \ i = 1,2,\cdots, m,
\end{split}
\end{equation}
where $\EE_{\zeta_i}[g_i(\theta,\zeta_i)] \leq 0$ are convex constraints formulated by using real-world data. Within such a formulation, the DM may obtain more accurate knowledge of $\Theta$, and provide less conservative solutions to resist the worst-case scenario. 

The above data-driven robust optimization problem has broad applicability in finance, revenue management, and engineering. For instance, in revenue management, 
let $D(s,p,\theta,\xi)$ be a random demand function  dependent on the product feature $s \in \RR^d$, the product price $p\in \RR$, and an associated  parameter $\theta \in \RR^{d+1}$. 
We consider the optimal pricing problem \citep{gallego2019revenue}, where the DM  aims to find the optimal price that maximizes the total expected reward, against the worst case parameter $\theta$ within a known uncertainty $\Theta$. 
\sg{Under data-driven models, if the DM has access to historical sales information that the expected demands of products $\tilde s_1,\cdots, \tilde s_m$ priced at $\tilde p_1,\cdots, \tilde p_m$ are not less than $d_1,\cdots, d_m$, respectively, she can impose 
 additional constraints to the uncertainty set and consider the following:
\begin{equation}\label{prob:robust_pricing}
\begin{split}
 \max_{p\in \mathbb{R}}  \min_{\theta \in \Theta}\    \EE_{\xi} [ p D(s,p ,\xi ;\theta)], 
\mbox{ s.t. }  \EE_{ \zeta_i} [D(\tilde s_i,\tilde p_i , \zeta_i ;\theta)] \geq d_i, \ \ i=1,2,\cdots, m.
\end{split}
\end{equation}
Here, $\xi$ and $ \zeta_i$ are random variables representing the uncertainty. By imposing data-driven expectation constraints, the DM  would find a less conservative solution and achieve higher profits. For a linear demand function of form  $D(s,p,\xi ;\theta) = s^\top \theta_{1:d}  + \theta_0 p + \xi$, the above problem clearly is a special instance of our general model \eqref{prob:sto_minmax}.}
\\
\noindent (B). \emph{Optimization with Misspecification.}
Optimization under misspecification considers a broad class of optimization problem
$$
\min_{x \in \cX} f(x,\theta^*),
$$
where $\theta^*$ is a vector of parameters whose true value is not readily accessible but could be learned from observational data. 
This class of problem finds wide applications in many areas including portfolio management,  power systems, inventory control, 
amongst others. For instance, $\theta^*$ could represent the true mean and variance of stock returns in portfolio management, the supply-demand parameters in power systems \citep{jiang2016solution}, or the holding costs in inventory control problems \citep{ghosh2021new}.
In such scenarios, 
$\theta^*$ is often obtained by solving 
a convex stochastic optimization problem $\min_{\theta} \EE_{\zeta}[g (\theta, \zeta)]$, where $g(\theta,\zeta)$ is constructed via observational data~$\zeta$. 

When the above stochastic optimization problem is strongly convex,
\cite{jiang13solution,jiang2016solution} proposed a simultaneous approach that iteratively updates the solution pair $\{ (x_t,\theta_t)\}$ to the following coupled stochastic optimization problem
\begin{equation*}
\begin{split}
\min_{x \in \cX} &  \,\,\,\, \EE_\omega [f(x,\theta^*,\omega)], \text{ where  } \theta^\ast = \argmin_{\theta \in \Theta} \EE_{\zeta}[g (\theta, \zeta)].
\end{split}
\end{equation*}
Unfortunately, in some practical applications, the function $\EE_{\zeta}[g (\theta, \zeta)]$ might not be strongly convex and preserves multiple optimal solutions. 
\sg{For example, to learn a high-dimensional parameter $\theta^*$ from given data, one appealing approach of modern statistical learning is the penalized quasi-likelihood~\citep{fan2014challenges}. Here, we consider an $\ell_1$-regularized least squares estimator, i.e.,
\[
g(\theta,\zeta) = (\zeta_1^\top \theta - \zeta_2)^2 + \lambda \|\theta\|_1,
\]
where $\zeta_1 \in \mathbb{R}^p, \zeta_2\in \mathbb{R}$ are random data and $\lambda > 0$ is the regularization parameter. In this case, $\EE_{\zeta}[g (\theta, \zeta)]$ is not necessarily strongly convex in $\theta$ and multiple optimal solutions may exist. 
Alternatively, we might consider the pessimistic optimization problem to find the optimal solution $x^*$ against the worst-case optimal parameters that 
\begin{equation*}
\begin{split}
\min_{x \in \cX} \max_{\theta} &  \,\,\,\, \EE_{\omega}[ f(x,\theta, \omega) ], \text{ s.t.  } \EE_{\zeta}[g(\theta, \zeta )] \leq  g^* := \min_{\theta}\ \EE_{\zeta}[g(\theta, \zeta )].
\end{split}
\end{equation*}
This problem can be cast into the class of minimax optimization under expectation constraints if $g^*$ is known as a prior or can be well-estimated by empirical observations.

Moreover, we could always consider the following model 
\begin{equation}\label{prob:misspecified}
\begin{split}
\min_{x \in \cX} \max_{\theta} &  \,\,\,\, \EE_{\omega}[ f(x,\theta, \omega) ], \text{ s.t.  } \EE_{\zeta}[g(\theta, \zeta )] \leq  g^* + \epsilon  := \min_{\theta}\ \EE_{\zeta}[g(\theta, \zeta )] + \epsilon ,
\end{split}
\end{equation}
where $\epsilon > 0 $ is some tolerance parameter. This is in fact a more realistic model as the desired $\theta^*$, even uniquely exists, can not be computed exactly. Here, model \eqref{prob:misspecified} takes this optimization error into consideration by introducing the tolerance parameter $\epsilon$ and the min-max robustness in the objective. }

\noindent \sg{\emph{(C) Online AUC Maximization with Fairness Constraints.}
AUC is a widely used metric for measuring the classification performance for imbalanced data, which concerns the overall performance of a functional family of classifiers and quantifies their ability of correctly ranking any positive instance with 
regards to a randomly chosen negative instance. 
Consider the feature space ${\cal W}\subseteq \mathbb{R}^d$ and response space ${\cal Y} = \{+1, -1\}$. Let the training data set ${\cal D}:=\{(w_i, y_i)\}_{i=1}^n$ be i.i.d. samples drawn from an unknown distribution on ${\cal W}\times {\cal Y}$.  
The  AUC optimization {with the linear classifier} solves the following maximization problem: 
\begin{equation*}
    \max_{x\in \mathbb{R}^d} \, \EE_{(w,y)\in {\cal D}, (w',y')\in {\cal D}} \big[
    \mathbb{I}_{[x^\top w  >  x^\top w'   ]} \mid y=1, y'=-1
    \big].
\end{equation*}
Here $\mathbb{I}$ is the indicator function that takes value $1$ if the argument is true and $0$ otherwise. 

Since the indicator function $\mathbb{I}$ is discontinuous, convex loss functions are used in the literature as surrogates. Specifically, \cite{ying2016stochastic} used the square loss and reformulated the obtained problem into the following stochastic minimax optimization:
\begin{equation*}
    \begin{split}
      \min_{x,a,b} \max_{\alpha} \, \left \{   \EE_{(w,y) \in \cD} \Big [ f \big (x,a,b,\alpha; (w,y) \big )  \Big  ]  \right \}
    \end{split}
\end{equation*}
where 
\begin{equation*}
    \begin{split}
        f(x,a,b,\alpha ;(w,y))  & = (1-p)(x^\top w - a)^2 \mathbb{I}_{[y=1]} + p(x^\top w - b)^2 \mathbb{I}_{[y =-1]} 
        \\
        & \quad + 2(1+\alpha ) \Big (p w_i^\top x \mathbb{I}_{[y=-1]} - (1-p) w^\top x \mathbb{I}_{[y=1]} \Big ) - p(1-p)\alpha^2.
    \end{split}
\end{equation*}
Clearly, this reformulation is a special unconstrained version of our general model \eqref{prob:sto_minmax} and can be solved by various stochastic min-max algorithms. 

Despite the successful development of online AUC maximization, the current framework lacks the incorporation of fairness concerns, which is becoming more and more important recently. Indeed, fairness learning has experienced significant growth and garnered substantial research attention in the past years. \sg{An important stream of research is dedicated to ensuring fairness across different groups (Zafar et al., 2019). In this context, each group represents a sensitive real-world label, denoted as $u=0,1$, which can include attributes such as race or gender. It then aims at building up a classifier that treats different groups similarly. For instance, to ensure the probability of being classified to a certain category, e.g., $y=1$, is not discriminated among sensitive groups $u = 0,1$, we may enforce the following probability constraint in AUC maximization
$$
\text{Prob}( w^\top x \geq 0 \mid  u = 0) = \text{Prob}(w^\top x \geq 0  \mid  u = 1).
$$
However, this type of constraint is generally computationally  intractable. 
}
{Fortunately, \cite{zafar2019fairness} pointed out that for any $x$ satisfying the above constraint, its corresponding empirical covariance is approximately zero, i.e., 
$$
\EE_{(w,u) \in \cD}[ ( u  - \bar u)w^\top x] \approx 0, 
$$
where $\bar u $ represents the average of $u$ over the dataset. As a result, we may consider the fairness-constrained AUC maximization  problem 
\begin{equation}\label{eq:AUC_fairness}
    \begin{split}
      \min_{x,a,b} \max_{\alpha} & \ \left \{    \EE_{(w,y) \in \cD}  \ \Big [ f \big (x,a,b,\alpha; (w,y) \big )  \Big  ] \right \} 
      \\
      \text{s.t.} & \ \  \EE_{(w,u) \in \cD}[ ( u  - \bar u)w^\top x] \leq c,
      \\
      & \ \ \EE_{(w,u) \in \cD}[ ( u  - \bar u)w^\top x]  \geq -c,
    \end{split}
\end{equation}
for some tolerance parameter $c \geq 0$. Clearly, problem \eqref{eq:AUC_fairness} is a special case of \eqref{prob:sto_minmax}.
}}

\sg{We shall point out that all these applications can be regarded as special realizations of the two-player zero-sum game model \citep{myerson1997game,von2007theory}  where each player targets maximizing its own reward against the opponent's strategy subject to certain budget constraints. Specifically, in a zero-sum game, players $A$ and $B$ would propose  strategies $x \in \cX$ and $y \in \cY$ associated with random costs $c_A(x,\xi)$ and $c_B(y,\zeta)$. Employing these strategies would lead to a random reward $f(x,y,\omega)$ to player $A$ and result in the same loss to player $B$. Suppose each player also imposes a budget constraint such that the expected cost incurred by their own strategies must not exceed $b_A$ and $b_B$, respectively. 
Given budgets $b_A$ and $b_B$, the Nash Equilibrium of the game can be found by considering the following:
\begin{equation*}
\min_{x \in \cX} \max_{y \in \cY} \ \EE_{\omega}  [f(x,y,\omega)], \mbox{ s.t. } \EE_{\xi} [ c_A(x,\xi) ] \leq b_A, \,  \EE_{\zeta} [ c_B(x,\zeta) ] \leq b_B.
\end{equation*}
Our algorithms also work for this general model.}

\noindent {\bf Additional examples:} The above examples fit our model well. There are also many interesting examples where complex expectation constraints are needed. For these applications, our analysis can  be applied under some additional appropriate assumptions.

\noindent (D). \emph{Generative Adversarial Network.} The Generative Adversarial Network (GAN) introduced in \citep{goodfellow2014generative} is a new generative framework that aims to generate new data preserving the same characteristics  as the given training set. Given the target distribution $\mu$ of the training set and some prior distribution $\gamma$, GAN solves the following optimization problem
\begin{equation}\label{eq:org_gan}
\min_{\omega} \max_{\theta} \left\{ \EE_{x\sim \mu}[\log D_\theta (x)]  + \EE_{z\sim \gamma}[\log(1 - D_\theta(G_\omega(z)))]\right\},
\end{equation}
where $\theta$ and $\omega$ are unknown parameters and $D_\theta$ and $G_\omega$ are some parameterized discriminator and generator functions. In real applications, additional constraints are imposed on the generator function $G_\omega$ to produce data points satisfying additional structural constraints \citep{heim2019constrained,di2020efficient}, such as ``generating an image more like image A than image B'', ``generating chemically valid drug molecules'', and ``generating playable games'', etc. Hence, the following data-driven constrained extension of \eqref{eq:org_gan} needs to be  investigated carefully:
\begin{equation*}\label{eq:cgan}
\begin{split}
 \min_{\omega} \max_{\theta} \  \EE_{x\sim \mu}[\log D_\theta (x)]  + \EE_{z\sim \gamma}[\log(1 - D_\theta(G_\omega(z)))],  \ \ \ \ 
\mbox{s.t. } \, \EE_{\xi}[\psi(G_\omega(z),\xi)] \le \tau,
\end{split}
\end{equation*}
where certain loss function $\psi$  and parameter $\tau$ are employed to control the possibility of the generator function producing valid structures.

\noindent (E). \emph{Pessimistic Bi-level Optimization.} We consider the following pessimistic bi-level problem in the standard form \citep{loridan1996weak,dempe2002foundations}
\begin{equation}\label{prob:pbbi}
\begin{split}
& \min_{x}\max_{y \in M(x)}  \,\,\,\, f(x,y), \,\,\,\,
\mbox{s.t. }  h(x) \leq 0, \, \mbox{where } M(x) : = \argmin_{y} \tilde g(x,y).
\end{split}
\end{equation}
We are particularly interested in the case where for any given $x$, the nonempty solution set $M(x)$ may not be a singleton. Under mild assumptions, by using the well-known optimal value reformulation \citep{outrata1988note, mitsos2008global, dempe2013bilevel}, Prob.~\eqref{prob:pbbi} can be  written equivalently~as 
\begin{equation}\label{prob:pblo_re}
\begin{split}
 \min_{x} \max_{y}  \,\,\,\, f(x,y),\,\,\,\,
\mbox{s.t. }  h(x) \leq 0, \, \tilde g(x,y) \le \tilde g^*(x), 
\end{split}
\end{equation}
where the optimal
objective value of the lower-level problem $\tilde g^*$  is defined by $\tilde g^*(x) : = \min_y \tilde g(x,y)$ for any fixed $x$.  Consider a special case where the lower-level objective function $\tilde g$ is separable in terms with $x,y$, i.e., 
$\tilde g(x,y) = \hat g(x) + g(y)$
for continuous functions $\hat g$ and $g$. Then, it holds that $\tilde g^*(x) : = \min_y \tilde g(x,y) = \hat g(x) + g^*$ with $g^* = \min_y g(y)$.
Now, we can see that
\begin{equation*}
\left\{
y\mid \tilde g(x,y) \le \tilde g^*(x)
\right\} = \left\{
y\mid \hat g(x) + g(y) \le \hat g(x) + g^*
\right\} = \left\{
y \mid g(y) \le g^*
\right\}.
\end{equation*}
Hence, Prob.~\eqref{prob:pblo_re}, under the separable assumption, can be equivalently recast as
\begin{equation}\label{prob:moti_bi}
\begin{split}
\min_{x} \max_{y}  \,\,   f(x,y) 
\mbox{ s.t. }  h(x) \leq 0, \, g(y) -g^* \le 0.
\end{split}
\end{equation}
The above reformulation has also been discussed in \cite{wiesemann2013pessimistic}.
Clearly, Prob.~\eqref{prob:moti_bi} can be treated as a special case of Prob.~\eqref{prob:sto_minmax} if $h$, $g$ are assumed to be convex and $f$ is convex in $x$ and concave in $y$. 

\subsection{Related Work}
\begin{enumerate}
	\item 	\textbf{Convex-Concave Saddle Problems with Simple Constraints.}
	Many algorithms have been proposed for solving convex-concave saddle problems with simple constraints. We briefly review some related work here. 
	The subgradient method was first analyzed by \cite{Nemirovsky83} for a saddle point problem, and subsequently studied by \cite{nemirovski2009robust} under the stochastic setting. 
	In particular, these classical works established the $\cO(1/{\sqrt{N}})$ convergence rate for both stochastic and deterministic settings. Inspired by Nesterov's smoothing technique \citep{nesterov2005smooth}, the convergence rate under the deterministic setting was further improved to $\cO(1/{N})$ by \cite{Nemirovski2005} using the Mirror-Prox method for solving a special class of saddle point problem with Lipschitz continuous gradient.	The saddle point algorithms have been further extensively studied by \cite{chambolle2011first,chen2014optimal,chen2017accelerated,he2015mirror,hamedani2018primal}. 
	However, all the works mentioned cannot handle complex constrained minimax problems.
	\item 
	\textbf{Expectation-Constrained Optimization.}
    The classic penalty approach, including
    exact penalty, quadratic penalty, and augmented Lagrangian methods \citep{bertsekas99nonlinear}, is an important stream of research to solve constrained optimization problems. However, the validity and efficiency of this approach depend critically on the solvability of the   subproblems involved, for which external algorithms are often required. {Other works} for constrained optimization are cast into two lines, deterministic and stochastic constrained optimization, based on the types of constraints.  
   On the one hand, the nonlinear deterministic constraints problem has been extensively studied 
    under various settings, including  gradient-type methods with functional constraints \citep{nesterov1998introductory}, constrained level-set optimization \citep{lemarechal1995new,lin2020data}, and  Frank-Wolfe \citep{Lan2020ConditionalGM}.
   On the other hand, stochastic gradient methods for stochastic-constrained optimization was first introduced by 
    \cite{lan2016algorithms}, and subsequently studied by \cite{boob2019stochastic} via primal-dual approaches. \sg{Notably, the expectation-constrained problem has also been considered by 
\cite{shapiro2013sample,oliveira2017sample,oliveira2022sample}  using the SAA approach.}
It is also worth mentioning that the online constrained optimization was studied by \cite{yu2017online}. One can refer to \cite{boob2019stochastic} for more detailed reviews.
Quite recently, \cite{zhang2021stochastic} extended the idea in \cite{boob2019stochastic} by proposing a linearized proximal method of multipliers,
which requires external algorithms for  solving the inner subproblems and cannot handle minimax objective functions. 
\end{enumerate}

\subsection{Contributions and Outline}
We summarize our contributions here.

\begin{enumerate}
	\item[(i)] We identify a model of stochastic minimax optimization problems with complex expectation constraints. The new model distinguishes itself from classic minimax optimization problems in at least two aspects: (1) it  deals explicitly with challenging expectation constraints and is a more realistic model; (2) it is flexible and well-suited to data-driven modeling.

	\item[(ii)] We propose a basic primal-dual algorithm that conducts simple stochastic projected gradient descent at each update to handle the proposed expectation-constrained stochastic minimax optimization problem.
	Without \sg{assuming the boundedness of $\cX$ and $\cY$, and bounded second moments of the generated dual iterates,} we show that our algorithm achieves the optimal $\cO(1/{\sqrt{N} })$ rate of convergence for objective optimality gap, duality gap, and feasibility residuals. 
	
	\item[(iii)] We further enhance our basic primal-dual algorithm by allowing adaptive selections of the step-sizes, without fixing the total number of iterations a priori. We provide a rigorous analysis to show that the enhanced adaptive version also enjoys the optimal $\cO(1/{\sqrt{N} })$ rate of convergence for objective optimality gap, duality gap, and feasibility residuals. 
	\item[(iv)] We verify the theoretical rate of convergence results by conducting numerical experiments on quadratic-constrained quadratic saddle point optimization problems and robust optimal pricing problems. Numerical results indicate that our algorithms can efficiently and robustly solve minimax problems with  thousands of expectation constraints.
	\end{enumerate}

\textbf{Notation:} Given a smooth map $g:\RR^n \to \RR^m$,  we denote $g(x)_+ = \max\{ g(x), 0\}$ and write $\nabla g(x) = [\nabla g_1(x), \ldots, \nabla g_m(x)]\in \RR^{n\times m}$ for any $x\in\RR^n$. We use $ \| \cdot \| $ to represent the Euclidean norm, \sg{and use $\EE[\cdot]$ without subscript to indicate that full expectation respected to all the randomness is taken.} 

\textbf{Paper Organization:} The remainder of this paper is as follows. In Section \ref{sec:preliminary}, we reformulate the expectation constrained optimization problem \eqref{prob:sto_minmax} into a saddle point problem and introduce the sampling oracle along with some necessary assumptions.  In Section \ref{sec:basic}, we develop a Basic-CSPD algorithm that 
iteratively updates the primal and dual variables by projected stochastic gradient descent methods. We show that the generated sequence converges to the optimal solution with the optimal $\cO(1/{\sqrt{N}})$ rate of convergence.  In Section \ref{sec:adp_alg}, we propose a modified algorithm, called Adp-CSPD, which employs adaptive step-sizes without fixing the total number of iterations a priori. We further show that the Adp-CSPD algorithm still enjoys the optimal  $\cO(1/{\sqrt{N}})$ rate of convergence.

\section{Primal-Dual Formulation and Sampling Oracle}\label{sec:preliminary}
As  mentioned above, we propose to solve Prob.~\eqref{prob:sto_minmax} using the stochastic primal-dual approach. In this section, we explicitly write the equivalent saddle point reformulation of Prob.~\eqref{prob:sto_minmax} and discuss how the stochastic zeroth- and first-order information of the involved expected functions is acquired. Some basic assumptions imposed on our algorithmic developments are also specified.

Consider the following saddle formulation of Prob.~\eqref{prob:sto_minmax}:
\begin{equation}\label{prob:saddle_point}
\begin{split}
\min_{x\in \cX} \max_{\gamma \in \RR_+^{m_1} } \max_{y \in \cY}  \min_{\lambda \in \RR_+^{m_2} }& \,\,\,\, \Big \{  \cL(x,y,\gamma, \lambda)  : =  F(x,y) +  \gamma^\top H(x)  -   \lambda^\top G(y) \Big \}, 
\end{split}
\end{equation}
where $\gamma \in \RR_+^{m_1}$ and $\lambda  \in \RR_+^{m_2}$ are the multipliers associated with constraints $H(x) \le 0$ and $G(y) \le 0$, respectively. We assume that the set of saddle points corresponding to problem \eqref{prob:saddle_point} is nonempty, which holds under mild Slater's condition. 
Here, we call 
$(x^*,y^*,\gamma^*,\lambda^*) \in \cX \times\cY \times   \RR_+^{m_1} \times  \RR_+^{m_2}$ a \emph{saddle} point to  Prob. \eqref{prob:saddle_point} if
\begin{equation}\label{eq:min_max}
\cL(x, y^*,\gamma^*,\lambda) \geq \cL(x^*,y^*,\gamma^*,\lambda^*) \geq   \cL(x^*, y, \gamma,\lambda^*), \,\,\,\,\forall (x,y,\gamma,\lambda) \in \cX \times\cY \times   \RR_+^{m_1} \times  \RR_+^{m_2}.
\end{equation}
The above relationship \eqref{eq:min_max} directly suggests the complementary slackness conditions, i.e., $   H(x^*)^\top \gamma^*  = 0$ and $ G(y^*)^\top \lambda^* = 0$, and further implies
$$ 
 F(x^*, y) \le F(x^*, y^*) \le F (x,y^*), \quad \forall x\in \widetilde \cX:= \left\{ x\in \cX\mid H(x)\le 0 \right\}, \, y \in  \widetilde \cY := \left\{
y\in \cY \mid G(y) \le 0
\right\}.
$$
That is, $(x^*, y^*)$ is a saddle point to Prob. \eqref{prob:sto_minmax}.

\textbf{Evaluation metrics:}
Let $( x, y) \in \cX\times \cY$ be an approximate solution pair returned by a certain algorithm. We evaluate the quality of this pair $(x,y)$ by considering both the objective optimality gap $F( x,y^*) - F(x^*, y)$, the duality gap $\max_{\tilde y \in \widetilde \cY} F(x,\tilde y) - \min_{\tilde x \in \widetilde \cX}F(\tilde x,y)$,  and feasibility residuals $\| H( x)_+\|_2$ and $\| G(y)_+\|_2$. In fact, we derive upper bounds for  $F(x,\bar y) - F(\bar x,y)$ for any feasible pairs $(\bar x, \bar y)$, i.e., $(\bar x, \bar y)\in \widetilde \cX \times \widetilde \cY$. Note that these evaluation metrics are  used extensively in the literature. See for example \citep{boob2019stochastic}. 
Since $(x,y)$ is not necessarily a feasible solution pair, the objective gap $F(x, y^*) - F(x^*, y)$ may be negative. Fortunately, by using the strong duality, we establish a lower bound for the objective optimality gap in terms of feasibility residuals 
as follows.

\begin{lemma}\label{lemma:obj_lower_bound}
For any $x\in\cX$ and $y\in \cY$, it holds that
\[
F(x,y^*) - F(x^*,y) \ge -\norm{\gamma^*}\norm{H(x)_+} - \norm{\lambda^*}\norm{G(y)_+} \ge -\max\{ \norm{\gamma^*}, \norm{\lambda^*} \} \left\{
\norm{H(x)_+} + \norm{G(y)_+}
\right\}.
\]
\end{lemma}
{We provide the detailed proofs in Appendix Section \ref{sec:proof_of_lemma_obj_lower_bound}.} 

Next, we rigorously specify the sampling environment. {In this paper, we assume the existence of the following black-box sampling oracle $(\cS\cO)$ such that:}
\begin{enumerate}
    \item[(i)] Given $x\in \cX \subset \RR^{d_x}, y \in \cY \subset \RR^{d_y}$, the $\cS\cO$ independently returns a sampled sub-gradient  $\widetilde \nabla_x f(x,y, \omega) \in \RR^{d_x}$ and a sampled sub-gradient $\widetilde \nabla_y f(x,y, \omega) \in \RR^{d_y}$. 
    \item[(ii)] Given $x \in \cX \subset \RR^{d_x}$, the $\cS\cO$ independently returns a sampled noisy vector $h(x,\xi) \in \RR^{m_1} $ and a sampled gradient $\widetilde \nabla_x h(x,\xi) \in \RR^{d_x \times m_1} $.
    \item[(iii)] Given $y \in \cY \subset \RR^{d_y}$, the $\cS\cO$ independently returns a sampled noisy vector $g(y,\zeta) \in \RR^{m_2} $ and a sampled sub-gradient $\widetilde \nabla_y g(y,\zeta) \in \RR^{d_y \times m_2}  $. 
\end{enumerate}

Note that since the smoothness of any $f, g, h$ is not assumed, {when any of these functions is \emph{non-smooth}, upon each query,}
we assume that the $\cS\cO$ returns a noisy sub-gradient to serve as the sampled first-order information.

Throughout this paper, we impose the following unbiasedness and bounded second moments assumptions on the stochastic objective function $f$.
\begin{AS}\label{assumption:01}
	Let $C_f$  be  a positive scalar. The function $f$ satisfies
	\begin{enumerate}
\item {For every $x \in \cX$, the function $F(x,\cdot)$ is Lipschitz continuous, i.e., $$
		\| F(x,y_1) - F(x,y_2) \| \leq C_f \| y_1 - y_2\|, \,\, \forall y_1,y_2 \in \cY.
		$$
		For every $y \in \cY$, the function $F(\cdot,y)$ is Lipschitz continuous, i.e.,} $$
		{\| F(x_1,y) - F(x_2,y) \| \leq C_f \| x_1 - x_2\|, \,\, \forall x_1,x_2 \in \cX.}
		$$
		\item For every $x \in \cX$ and $y \in \cY$, the sampled sub-gradients $\widetilde \nabla_x f(x,y,\omega)$ and $\widetilde \nabla_y f(x,y,\omega)$ are unbiased such that 
		$$
		{\EE[ \widetilde \nabla_x f(x,y,\omega) ] = \widetilde \nabla_x F(x,y) \in \partial_x F(x,y), \mbox{ and }\EE[ \widetilde \nabla_y f(x,y,\omega) ] = \widetilde \nabla_y F(x,y) \in \partial_y F(x,y)}
		$$
		and preserve bounded second moments such that 
		$
		\EE[ \| \widetilde \nabla_x f(x,y,\omega)  \|^2 ] \leq  C_f^2$, and $\EE[ \| \widetilde \nabla_y f(x,y,\omega) \|^2 ] \leq C_f^2.
		$
	\end{enumerate}
\end{AS}
We also impose similar assumptions on the constraints $h$ and $g$:
\begin{AS}\label{assumption:02}
Let $C_h,\sigma_h, C_g, \sigma_g$ be positive scalars. Then functions $g$ and $h$ satisfy
	\begin{enumerate}
		\item The function  $H$ is  Lipschitz continuous such that 
		$$
		\| H(x_1) - H(x_2)\| \leq C_h \|x_1 - x_2 \|, \,\,\,\forall x_1,x_2 \in \cX.
		$$
		For any $x\in\cX$, the sampled sub-gradient $\widetilde \nabla h(x,\xi)$ is unbiased and preserves bounded second moments such that 
		$
		\EE[\widetilde \nabla h(x,\xi)] = \widetilde \nabla H(x) \in \partial H(x)$, and $\EE[\| \widetilde \nabla h(x,\xi) \|^2] \leq C_h^2.
		$
		Moreover, the sampled value $h(x,\xi)$ is unbiased and preserves bounded variance such that
		$$
		\EE[ h(x,\xi)] = H(x), \mbox{ and }  \EE[ \| h(x,\xi)  - H(x)  \|^2] \leq \sigma_h^2.
		$$
		\item The function  $G$ is  Lipschitz continuous such that 
		$$
		\| G(y_1) - G(y_2)\| \leq C_g \| y_1 - y_2\|,\,\,\,\forall y_1,y_2 \in \cY.
		$$
		For any $y\in\cY$, the sampled sub-gradient $\widetilde \nabla g (y,\zeta)$ is unbiased and preserves bounded second moments such that 
		$\EE[\widetilde \nabla g(y,\zeta)] = \widetilde \nabla G(y) \in \partial G(y)$,  and $\EE[ \|\widetilde \nabla g(y,\zeta) \|^2] \leq C_g^2.
		$
		The sampled value $g(y,\zeta)$ is unbiased and preserves bounded variance such that
		$$
		\EE[ g(y,\zeta)] = G(y), \mbox{ and }  \EE[ \| g(y,\zeta) - G(y) \|^2] \leq \sigma_g^2.
		$$
	\end{enumerate}
\end{AS}
We remark that these assumptions are \emph{standard} in stochastic optimization literature. 

\section{Basic Primal-Dual Algorithm.} \label{sec:basic}
In Section \ref{sec:preliminary}, we reformulate the expectation-constrained minimax optimization problem \eqref{prob:sto_minmax} as a saddle point problem \eqref{prob:saddle_point}. 
To deal with this saddle point reformulation, we propose a primal-dual algorithm that alternately updates the primal sequences $\{x_t,y_t\}$ and the associated dual sequences $\{\gamma_t,\lambda_t\}$ by conducting projected stochastic gradient descent steps. We also conduct a comprehensive convergence analysis of the algorithm. The corresponding convergence rates of the objective optimality gap, duality gap, and feasibility residuals are derived.

We start with a detailed description of our stochastic primal-dual scheme to solve the expectation-constrained minimax optimization problem \eqref{prob:sto_minmax}. 
In iteration $t$, we query the $\cS\cO$ at $(x_t,y_t)$ twice to obtain both the stochastic zeroth-order information $h(x_{t},\xi_t^1), g(y_{t},\zeta_t^1)$, and first-order information $ \widetilde \nabla_x f(x_t, y_t ,\omega_t^1)$, $\widetilde \nabla_y f(x_t, y_t ,\omega_t^2)$, $\widetilde  \nabla h(x_t,\xi_t^2)$, and $\widetilde \nabla g (y_t,\zeta_t^2) $. Here, $\{ \omega_t^j,  \xi_t^j, \zeta_t^j \}_{j=1,2}$ represents two independent realizations of random variables $\omega,\xi,\zeta$ in iteration $t$. Consequently, the returned stochastic function values and sub-gradients are \emph{independent} of each other. 
With this information available, we first update  $\gamma_t$ and $\lambda_t$ using projected stochastic gradient steps, i.e., 
\begin{equation}\label{update:basic_gamma_lambda}
\begin{aligned}
\gamma_{t+1} = {}& \argmax_{ \gamma \in \RR_{+}^{m_1} } \Big \{h(x_{t},\xi_t^1)^\top \gamma - \frac{\beta_{t}}{2} \| \gamma_t - \gamma \|^2 \Big \}, \\
\lambda_{t+1} = {}&\argmin_{ \lambda \in \RR_{+}^{m_2} } \Big \{ - g(y_{t},\zeta_t^1)^\top \lambda  + \frac{\alpha_{t}}{2} \| \lambda_t - \lambda \|^2 \Big \} .
\end{aligned}
\end{equation}
Here we denote by $\gamma_{t+1,i}$ and $\lambda_{t+1,j}$ the $i$-th and $j$-th component of $\gamma_{t+1}$ and $\lambda_{t+1}$, respectively.
Using these updated dual variables, we update our primal variables $x_t$ and $y_t$ by:
\begin{equation}\label{update:basic_x}
\begin{aligned} 
x_{t+1} ={}& \argmin_{x\in \cX}  \Big \{  \big( \widetilde \nabla_x f(x_t, y_t ,\omega_t^1)  + \sum_{i=1}^{m_1} \gamma_{t+1,i} \widetilde  \nabla h_i(x_t,\xi_t^2)\big) ^\top   x  + \frac{\eta_t}{2} \| x - x_t\|^2 \Big \},
\\
y_{t+1}  ={}& \argmax_{y\in \cY}  \Big \{  \big (  \widetilde \nabla_y f(x_t, y_t ,\omega_t^2)  - \sum_{j=1}^{m_2} \lambda_{t+1,j} \widetilde \nabla g_{j} (y_t,\zeta_t^2)  \big ) ^\top  y  - \frac{\kappa_t}{2} \| y - y_t\|^2 \Big \}. 
\end{aligned}
\end{equation}
In these updates, $\beta_t$, $\alpha_t$, $\eta_t$, and $\kappa_t$ are positive step-sizes.
We summarize the details of the above process in Algorithm \ref{alg:1}, which we refer to as the Basic Constrained Stochastic Primal-Dual (Basic-CSPD) algorithm.

\begin{algorithm}[t]  \small 
	\caption{Basic Constrained Stochastic Primal-Dual (Basic-CSPD) for Constrained Minimax Optimization} \label{alg:1}
	\begin{algorithmic}
		\STATE{\bfseries Input : }  Positive step-sizes $\{ \alpha_t\}$,  $\{ \beta_t\}$,  $\{ \kappa_t\}$, $\{ \eta_t\}$, and initial points $(x_0, y_0, \gamma_0, \lambda_0)\in \cX\times\cY\times\RR_+^{m_1}\times\RR_+^{m_2}$ \\
		\FOR{$t = 0,1, 2, ..., N-1$}
		\STATE  Query the $\cS\cO$ at $x_t$ to obtain $h(x_{t},\xi_t^1) $, update the dual variable $\gamma_{t+1}$ by 
		$$
		\gamma_{t+1} =  \argmax_{ \gamma \in \RR_{+}^{m_1} } \Big \{h(x_{t},\xi_t^1)^\top \gamma - \frac{\beta_{t}}{2} \| \gamma_t - \gamma \|^2 \Big \} .
		$$
		Query the $\cS\cO$ at $y_t$ to obtain $g(y_{t},\zeta_t^1) $, 
		update the dual variable $\lambda_{t+1}$ by 
		$$
		\lambda_{t+1} =  \argmin_{ \lambda \in \RR_{+}^{m_2} } \Big \{ - g(y_{t},\zeta_t^1)^\top \lambda  + \frac{\alpha_{t}}{2} \| \lambda_t - \lambda \|^2 \Big \} .
		$$
		Query the $\cS\cO$ at $(x_t,y_t)$ to obtain $\widetilde \nabla_x f(x_t, y_t ,\omega_t^1) $ and $\widetilde \nabla h(x_t,\xi_t^2)$,
		update $x_{t+1}$ by 
		$$
		x_{t+1} = \argmin_{x\in \cX}  \Big \{  \big( \widetilde \nabla_x f(x_t, y_t ,\omega_t^1)  + \sum_{i=1}^{m_1} \gamma_{t+1,i} \widetilde  \nabla h_i(x_t,\xi_t^2)\big) ^\top   x  + \frac{\eta_t}{2} \| x - x_t\|^2 \Big \}. 
		$$
		Query the $\cS\cO$ at $(x_t,y_t)$ to obtain $\widetilde \nabla_y f(x_t, y_t ,\omega_t^2) $ and $\widetilde \nabla g(y_t,\zeta_t^2)$, update $y_{t+1}$ by 
		$$
		y_{t+1} = \argmax_{y\in \cY}  \Big \{  \big (  \widetilde \nabla_y f(x_t, y_t ,\omega_t^2)  - \sum_{j=1}^{m_2}  \lambda_{t+1,j} \widetilde \nabla g_j (y_t,\zeta_t^2)  \big ) ^\top  y  - \frac{\kappa_t}{2} \| y - y_t\|^2 \Big \}. 
		$$
		\ENDFOR
		\STATE{\bfseries Output :} $\bar x_N = \frac{1}{N} \sum_{t=1}^N x_t$ and $\bar y_N = \frac{1}{N} \sum_{t=1}^N y_t$.
	\end{algorithmic}
\end{algorithm}

\subsection{Convergence Analysis}

Having presented Algorithm \ref{alg:1}, both its theoretical and practical  performances must be investigated. The key question is whether and how fast the generated primal-dual iterates converge to a saddle point to Prob.~\eqref{prob:sto_minmax}, in the presence of expectation constraints. 
Suppose that \sg{$\cX$ and $\cY$ are compact convex sets} and the generated dual sequences $\{\lambda_t\}, \{\gamma_t\}$ are uniformly bounded or have uniformly bounded second moments, then similar convergence analysis for convex-concave saddle problems \citep{nemirovski2009robust,lan2020first} can be used to analyze the convergence behavior of Algorithm~\ref{alg:1}. However, such an assumption is not guaranteed to be true and is usually considered to be rather restrictive in the literature.
 \sg{We note that very recently, \citep{boob2019stochastic} 
 proposed an algorithm for solving the vanilla convex stochastic optimization with compact convex sets $\cX, \cY$ and convex expectation constraints, and showed the uniform boundedness of {second moment of the} corresponding dual iterates. Unlike \citep{boob2019stochastic}, this work considers the min-max stochastic optimization with expectation constraints, which is more complicated than \citep{boob2019stochastic} where two pairs of primal-dual sequences $\{x_t,\gamma_t\}$ and  $\{y_t,\lambda_t\}$ inevitably interact with each other and induce compounded randomness.} This makes our analysis much more challenging.

For any given $(x,y,\gamma, \lambda)\in \RR^{d_x}\times \RR^{d_y} \times \RR^{m_1} \times \RR^{m_2}$, from 
\eqref{prob:saddle_point}, we obtain
subgradients of $\cL$ with respect to $x$ and $y$:
 \begin{equation}\label{eq:H3}
	{\widetilde \nabla_x \cL} (x,y,\gamma) = \widetilde \nabla_x F(x, y)   + \sum_{i=1}^{m_1}  \gamma_i \widetilde \nabla H_i(x), \quad {\widetilde \nabla_y \cL} (x,y,\lambda) = \widetilde \nabla_y F(x, y)  - \sum_{j=1}^{m_2} \lambda_{j} \widetilde \nabla G_j(y).  
\end{equation}
Given any random variables $\omega$, $\xi$, and $\zeta$, the sampled version of $\widetilde\nabla_x \cL$ and $\widetilde\nabla_y \cL$
 \begin{equation}\label{eq:H5}
 \left\{
 \begin{aligned}
{\widetilde \nabla_x L}(x,y,\gamma, \omega,\xi) := {}& \widetilde \nabla_x f(x, y,\omega)   +  \sum_{i=1}^{m_1} \gamma_{i} \widetilde \nabla h_i(x,\xi), \\[2pt]  
{\widetilde \nabla_y L} (x,y,\lambda, \omega,\zeta) :={}&\widetilde \nabla_y f(x, y,\omega)  - \sum_{j=1}^{m_2} \lambda_{j} \widetilde \nabla g_j(y,\zeta).
 \end{aligned}
 \right. 
 \end{equation}
For any $t \ge 0$, let $z_t = (x_t,y_t,\gamma_t,\lambda_t)$ and choose any feasible, {possibly random}, reference point $z = (x,y,\gamma,\lambda)$. Then, we define the \emph{gap function} at the pair $(z_t, z)$ as
\begin{equation}
Q(z_t, z) = \cL(x_{t}, y,\gamma,\lambda_{t}) -  \cL(x, y_{t}, \gamma_{t},\lambda).
\end{equation}
It is also noteworthy that, 
with specific choices of $\lambda$ and $\gamma$, this gap function can be used to investigate the convergence rates of the objective optimality gap, duality gap, and feasibility residuals, which is discussed in Sections \ref{sec:optimality_basic} and \ref{sec:optimality_adaptive}. 

 {We assume that Assumptions \ref{assumption:01} and \ref{assumption:02} hold throughout this section.} To start our analysis, in the following lemma, we provide an upper bound for the gap function $Q(z_{t+1},  z) $. 
\begin{lemma} \label{lemma:1}
Let $\{ z_t = (x_t,y_t,\gamma_t,\lambda_t) \}$ be the sequence generated by Algorithm \ref{alg:1}. Then, it holds for all {$(x,y,\gamma, \lambda)\in \cX \times \cY \times \RR^{m_1} \times \RR^{m_2}$} that
	\begin{equation}\label{eq:lemma_1}
	\begin{split}
	&Q(z_{t+1}, z) =  \cL(x_{t+1}, y,\gamma,\lambda_{t+1}) -  \cL(x, y_{t+1}, \gamma_{t+1},\lambda)   \\
	& \quad \leq  \Delta_x^{t+1} + \Delta_\gamma^{t+1} + \Delta_y^{t+1} + \Delta_\lambda^{t+1}   + \frac{3C_f^2 }{2\eta_t} + \frac{\eta_t}{6}\| x_{t+1} - x_t\|^2 + 	 \frac{3C_f^2 }{2\kappa_t}  +\frac{\kappa_t}{6} \| y_{t+1} - y_t\|^2,
	\end{split}
	\end{equation}
	where 
	\begin{equation}\label{def:Deltas}
\begin{split}
\Delta_x^{t+1} & := \big(\widetilde \nabla_x F(x_t, y_t) +   \sum_{i=1}^{m_1}   \gamma_{t+1, i } \widetilde \nabla H_i(x_t) \big) ^\top (x_t - x), \,\,\,\,
\Delta_\gamma^{t+1} : = \gamma^\top  H(x_{t+1})  -  \gamma_{t+1}^\top H(x_t), \\ 
\Delta_y^{t+1} & := \big( \widetilde \nabla_y F(x_t,y_t) -  \sum_{j=1}^{m_2} \lambda_{t+1,j} \widetilde \nabla G_j(y_t) \big)^\top(y - y_t) 
, 
\,\,\,\,
\Delta_\lambda^{t+1} : = \lambda^\top  G(y_{t+1})  -  \lambda_{t+1}^\top G(y_t) .
\end{split}
\end{equation}
\end{lemma}

As can  be observed from our proof in Appendix Section \ref{sec:diffL}, the obtained bound \eqref{eq:lemma_1} is rather general, in the sense that it depends only on our assumptions about the associated functions $f,g,h$, but not on the specific updating rules of the involved sequence $\{(\gamma_t, \lambda_t, x_t, y_t)\}$. We will use \eqref{eq:lemma_1}  to conduct convergence analysis for the algorithm discussed in Section \ref{sec:adp_alg}. 

Here, we can see that $\Delta_{x}^{t+1}$ and $\Delta_y^{t+1}$ serve as the upper bounds of $F(x_t,y_t)  - F(x,y_t)  +    \gamma_{t+1}^\top \big(H(x_t) -  H(x)\big) $ and $F(x_t,y) - F(x_t,y_t) - \lambda_{t+1}^\top\big(G(y) - G(y_t)\big)$, respectively. Further, we emphasize that here $(x,y)$ is allowed to be chosen as a random pair depending on the trajectory $\{(x_t,y_t,\gamma_t,\lambda_t)\}$, in contrast to existing studies where $(x,y) = (x^*,y^*)$ is deterministic \citep{boob2019stochastic}. This {dependence} makes our analysis more challenging especially when taking expectations on both sides of $\Delta_x^{t+1}$ and $\Delta_y^{t+1}$.


Suppose Algorithm \ref{alg:1} runs for $N$ iterations in total. 
We wish to evaluate the performance of the algorithm in the first $K$ $(K\le N)$ iterations. 
In the next result, we start by providing a bound for the cumulative value of $\Delta_{x}^{t+1} + \Delta_{\gamma}^{t+1}$ in the first $K$ rounds, i.e., $\sum_{t=0}^{K-1}  \big( \Delta_{x}^{t+1} + \Delta_{\gamma}^{t+1} \big)$. {The detailed proofs are provided in Appendix Section \ref{sec:proof_of_lemma_delta_x}.}
\sg{
\begin{lemma}
    \label{lem:deltaxgamma}
   Let $\{ (x_t,y_t,\gamma_t,\lambda_t) \}_{t=1}^N$ be the sequence generated by Algorithm \ref{alg:1} with $\eta_t = \eta_0 > 0, \, \beta_t = \beta_0 >0$ for all $1\leq t \leq N$. It holds for any $(x,\gamma)\in \cX \times \RR_+^{m_1}$ that
\begin{align*}
&\sum_{t=0}^{K-1} \big( 
\Delta_{x}^{t+1} + \Delta_{\gamma}^{t+1} + \frac{\eta_t}{6}\norm{x_{t+1} - x_t}^2
\big) + \frac{\eta_0}{2} \norm{x_K - x}^2 + \frac{\beta_0}{2}\norm{\gamma_K - \gamma}^2 \\
{} \le  & \  \frac{\eta_0}{2}\norm{x_0 - x}^2 + \frac{\beta_0}{2}\norm{\gamma_0 - \gamma}^2 + \frac{3K\norm{\gamma}^2 C_h^2}{2\eta_0} + \sum_{t=0}^{K-1} U_t(x,\gamma)
\end{align*}
where for any $(x,\gamma)\in \cX \times \RR_+^{m_1}$, 
\begin{align}\label{eq:Uxgamma}
    U_t(x,\gamma) :={}& \big( \widetilde \nabla_x L(x_{t},y_{t},\gamma_{t+1},\omega_t^1,\xi_t^2) - \widetilde \nabla_x \cL(x_{t},y_{t},\gamma_{t+1}) \big)^\top ( x -x_{t})  + \frac{3 \|\widetilde \nabla_x L(x_{t},y_{t},\gamma_{t+1},\omega_t^1,\xi_t^2)\|^2}{2\eta_t} \nonumber\\
    & +\big (  H(x_{t}) -  h(x_{t},\xi_t^1 ) \big )^\top (\gamma - \gamma_t)   + \frac{ \|H(x_{t}) -  h(x_{t},\xi_t^1 )\|^2 }{2\beta_{t} }.
\end{align}
\end{lemma}

We can obtain a similar upper bound for $\sum_{t=0}^{K-1}  \big( \Delta_{y}^{t+1} + \Delta_{\lambda}^{t+1} \big)$. Then, it is not difficult to combine these bounds to derive an upper bound for $\sum_{t=0}^{K-1} Q(z_{t+1},z)$ in the following lemma. 

\begin{lemma}\label{lem:qzz}
    Let $\{ z_t:=(x_t,y_t,\gamma_t,\lambda_t) \}_{t=1}^N $ be the sequence generated by Algorithm \ref{alg:1} with positive step-sizes $\eta_t = \eta_0 $,  $\kappa_t = \kappa_0 $, $\alpha_t = \alpha_0$, and $\beta_t = \beta_0$ for all $t = 1,2,\cdots, N$. Then it holds for any  $ z = (x,y,\gamma,\lambda) \in \cX \times \cY \times \RR_+^{m_1} \times \RR_+^{m_2}$ and $K = 1,\ldots,N$ that
    \begin{equation}
        \label{eq:qzz}
        \begin{aligned}
&\sum_{t=0}^{K-1} Q(z_{t+1},z) + \frac{\eta_0}{2} \norm{x_K - x}^2 + \frac{\beta_0}{2}\norm{\gamma_K - \gamma}^2 + \frac{\alpha_0}{2}\norm{\lambda_K - \lambda}^2 + \frac{\kappa_0}{2}\norm{y_K - y}^2\\
{}\le{}& \frac{3KC_f^2}{2\eta_0} + \frac{\eta_0}{2}\norm{x_0 - x}^2 + \frac{\beta_0}{2}\norm{\gamma_0 - \gamma}^2 + \frac{3K\norm{\gamma}^2 C_h^2}{2\eta_0} + \sum_{t=0}^{K-1} U_t(x,\gamma) \\
&+ \frac{3K C_f^2}{2\kappa_0} + \frac{\kappa_0}{2}\norm{y_0 - y}^2 + \frac{\alpha_0}{2}\norm{\lambda_0 - \lambda}^2 + \frac{3K\norm{\lambda}^2 C_g^2}{2\alpha_0} + \sum_{t=0}^{K-1} V_t(y,\lambda), 
\end{aligned}
    \end{equation}
where $U_t$ is defined in \eqref{eq:Uxgamma} and for any $(y,\lambda) \in \cY \times \RR_+^{m_2}$,
\begin{align}\label{eq:Vylambda}
    V_t(y,\lambda) := {}& \big (\widetilde \nabla_y L(x_{t},y_{t},\lambda_{t+1},\omega_t^2,\zeta_t^2) - \widetilde \nabla_y \cL(x_{t},y_{t},\lambda_{t+1}) \big )^\top ( y - y_{t})  + \frac{3 \|\widetilde \nabla_y L(x_{t},y_{t},\lambda_{t+1},\omega_t^2,\zeta_t^2)\|^2}{2\kappa_t} \nonumber\\
    {}&+\big (  G(y_{t}) -  g(y_{t},\zeta_t^1 ) \big )^\top (\lambda - \lambda_t)   + \frac{ \|G(y_{t}) -  g(y_{t},\zeta_t^1 )\|^2 }{2\alpha_{t} }.
\end{align}
\end{lemma}

In the next lemma, we establish bounds for $\sum_{t=0}^{K-1} U_t(x,\gamma)$ and $\sum_{t=0}^{K-1}  V_t(y,\lambda)$ appeared in \eqref{eq:qzz} for any  $(x,y,\gamma,\lambda)\in \cX \times \cY \times \RR_+^{m_1} \times \RR_+^{m_2}$ with bounded second moments. Appendix Section \ref{sec:apen_UtVt} provides detailed proof.
\begin{lemma}
    \label{lem:EUV}
    Under the same settings as in Lemma \ref{lem:qzz}, for any  {$(x,y,\gamma, \lambda)\in \cX \times \cY \times \RR_+^{m_1} \times \RR_+^{m_2}$ satisfying $\EE[\| x\|^2] < + \infty$, $\EE[\| y\|^2] < + \infty$, $\EE[\| \gamma\|^2] < + \infty$, and $\EE[\| \lambda\|^2] < + \infty$}, it holds that
    \begin{equation}
        \label{eq:EUVrandom}
        \begin{aligned}
            &\EE\Big[ \sum_{t=0}^{K-1} U_t(x,\gamma) \Big] \le \frac{\eta_0}{2}\EE[\|x\|^2] + \sum_{t=0}^{K-1} \frac{4}{\eta_t}\big(C_f^2 + C_h^2 \EE[\|\gamma_{t+1}\|^2]\big) + \sqrt{K}\EE[\|\gamma\|]\sigma_h + \frac{K\sigma_h^2}{2\beta_0},  \\
        &\EE\Big[ \sum_{t=0}^{K-1}  V_t(y,\lambda) \Big] \le \frac{\kappa_0}{2}\EE[\|y\|^2] + \sum_{t=0}^{K-1} \frac{4}{\kappa_t}\big(C_f^2 + C_g^2 \EE[\|\lambda_{t+1}\|^2]\big) + \sqrt{K}\EE[\|\lambda\|]\sigma_g + \frac{K\sigma_g^2}{2\alpha_0} .
        \end{aligned}
    \end{equation}
\end{lemma}

Now we are ready to investigate the convergence behavior of the gap function $Q(z_t,  z)$. }

\begin{theorem} \label{thm:1}
 Let $\{ z_t = (x_t,y_t,\gamma_t,\lambda_t) \}_{t=1}^N $ be the sequence generated by Algorithm \ref{alg:1} with positive step-sizes $\eta_t = \eta_0 $,  $\kappa_t = \kappa_0 $, $\alpha_t = \alpha_0$, and $\beta_t = \beta_0$ for all $t = 1,2,\cdots, N$.
 For any  {$z = (x,y,\gamma, \lambda)\in \cX \times \cY \times \RR_+^{m_1} \times \RR_+^{m_2}$ satisfying $\EE[\| x\|^2] < + \infty$, $\EE[\| y\|^2] < + \infty$, $\EE[\| \gamma\|^2] < + \infty$, and $\EE[\| \lambda\|^2] < + \infty$},
and all $1 \leq K \leq N$, it holds that
 \begin{equation}\label{eq:EQztz}
\begin{split}
&\EE  \Big [ \sum_{t=0}^{K-1} Q(z_{t+1}, z)   +  \frac{\eta_0}{2} \norm{x_K - x}^2 + \frac{\beta_0}{2}\norm{\gamma_K - \gamma}^2  + \frac{\alpha_0}{2} \norm{\lambda_K - \lambda}^2 + \frac{\kappa_0}{2}\norm{y_K - y}^2  \Big ] \\
\leq {}& \frac{3K C_f^2 }{2\eta_0}   + \frac{\beta_0}{2} \EE [ \| \gamma_0 - \gamma\|^2  ]    + \sqrt{K} \EE[\| \gamma\|] \sigma_h
 + \frac{ K \sigma_h^2 }{2\beta_0 } + \frac{3K \EE[\| \gamma\|^2 ] C_h^2 }{2\eta_0}  +  \frac{\eta_0}{2} \EE [  \| x_0 -x\|^2 ] 
 \\
 &  + \frac{3K C_f^2 }{2\kappa_0}  + \frac{\alpha_0}{2} \EE[ \| \lambda_0 - \lambda\|^2  ] + \sqrt{K} \EE[\| \lambda \|] \sigma_g
 + \frac{ K \sigma_g^2 }{2\alpha_0 } + \frac{3K \EE[\| \lambda\|^2 ] C_g^2 }{2\kappa_0}  + \frac{\kappa_0}{2} \EE [ \| y_0 -y\|^2 ] 
 \\
 &  +  \sum_{t=0}^{K-1} \frac{4}{\eta_t} (C_f^2 + C_h^2\EE[ \| \gamma_{t+1}\|^2] )      + \sum_{t=0}^{K-1} \frac{4}{\kappa_t} (C_f^2 + C_g^2\EE[ \| \lambda_{t+1}\|^2] )   +  \frac{\eta_0}{2} \EE [ \| x\|^2]  + \frac{\kappa_0}{2} \EE [ \| y\|^2]    .
\end{split}
\end{equation}
\end{theorem}
{We note that the reference point $(x,y,\gamma,\lambda) \in \cX\times\cY\times\RR_+^{m_1}\times\RR_{+}^{m_2}$ in \eqref{eq:EQztz} can be dependent on the solution trajectory $\{ (x_t,y_t,\gamma_t,\lambda_t) \}_{t=1}^N$ as long as it possesses bounded second moments. This flexibility allows us 
to employ different choices of $(x,y,\gamma,\lambda)$ to derive the convergence rates of objective optimality gap, duality gap, and feasibility residual in Section \ref{sec:optimality_basic}.}
Note that Theorem \ref{thm:1} can be easily proved by taking expecation of \eqref{eq:qzz} and using \eqref{eq:EUVrandom}.

We shall also point out that due to the presence of dual iterates $\lambda_{t}$ and $\gamma_{t}$ on the right-hand side of \eqref{eq:EQztz}, it is unclear whether the second moments of the primal and dual iterates, i.e., $
\EE[\|x_t\|^2]$, $
\EE[\|y_t\|^2]$, $
\EE[\|\lambda_t\|^2]$ and $\EE[\|\gamma_t\|^2]$, generated by our algorithm are \emph{uniformly bounded}. This necessitates us to investigate the trajectory of $\{(x_t,y_t,\gamma_t, \lambda_t)\}$ further.

\subsubsection{Uniform Boundedness of the Second Moments of Primal and Dual Iterates}
We observe that Theorem \ref{thm:1} provides a recursive relationship between the dual iterates in the $K$-th iteration, $(\lambda_K,\gamma_K)$, and those generated in previous iterations $\{(\lambda_t,\gamma_t)\}_{t=1}^{K-1}$. By using this recursion, in the next result we show that the dual iterates $\{(\lambda_t, \gamma_t )\}$, and consequently the primal iterates $\{(x_t,y_t)\}$, generated by our Basic-CSPD algorithm with specially chosen step-sizes have uniformly bounded second moments.  
\sg{
 \begin{proposition}[Boundedness of the second moments] \label{prop:boundedness}
 Let $\{ z_t:=(x_t,y_t,\gamma_t,\lambda_t) \}_{t=1}^N $ be the sequence generated by Algorithm \ref{alg:1} with positive step-sizes  $\eta_t =4\sqrt{N} C_h^2$, $ \kappa_t = 4\sqrt{N} C_g^2$, and $\alpha_t  =  \beta_t =4\sqrt{N} $ for all $t=0,1,\cdots, N$. Then for any saddle point $z^*= (x^*,y^*,\gamma^*,\lambda^*)$ and $1\leq K \leq N$, we have  
 \begin{equation*}
 \left\{
 \begin{aligned}
 & \EE [ \| \lambda_K \|^2 ] +  \EE [ \| \gamma_K \|^2 ]     \leq 2R e^{2}, \\
 & \EE [ \| \lambda_K \|^2 ] +  \EE [ \| \gamma_K \|^2 ] + C_h^2 \EE[\|x_K\|^2] + C_g^2 \EE[\|y_K\|^2] \le (2e^2 + 1) R,
\end{aligned}
\right.
 \end{equation*}
where $0<R<+\infty$ is a constant given by
 \begin{equation}\label{def:R}
\begin{split}
R & = \frac{11C_f^2 }{8 C_h^2  }   +2 \| \gamma_0 - \gamma^*\|^2     + \|\gamma^* \| \sigma_h
 + \frac{  \sigma_h^2 }{8 } + \frac{19 \|\gamma^*\|^2  }{8}  +  2 C_h^2 \| x_0 -x^*\|^2  +  4 C_h^2  \| x^*\|^2    
 \\
 & \quad + \frac{11 C_f^2 }{8 C_g^2 }  + 2  \| \lambda_0 - \lambda^*\|^2  + \| \lambda^*\| \sigma_g
 + \frac{  \sigma_g^2 }{8 } + \frac{19 \|\lambda^*\|^2  }{8}  + 2  C_g^2 \| y_0 -y^*\|^2  + 4 C_g^2  \| y^*\|^2 .    
\end{split}
\end{equation}
  \end{proposition}
}
The detailed proof is deferred to Appendix Section \ref{sec:app_A2}. 
By combining this key result with Theorem \ref{thm:1}, we can characterize the convergence behavior of the gap function $Q(z_t,z)$ in the following theorem.

\begin{theorem}\label{thm:boundsQ}
   Let $\{ z_t:=(x_t,y_t,\gamma_t,\lambda_t) \}_{t=1}^N $ be the sequence generated by Algorithm \ref{alg:1} with positive step-sizes $\eta_t =4\sqrt{N} C_h^2$, $ \kappa_t = 4\sqrt{N} C_g^2$, and $\alpha_t  =  \beta_t =4\sqrt{N} $ for all $t=0,1,2,\cdots, N$. Let  $R$ be the constant defined in \eqref{def:R}.  For any $ z = (x,y,\gamma,\lambda) 
   \in \cX \times\cY\times\RR_+^{m_1}\times\RR_+^{m_2}$ {satisfying $\EE[\| x\|^2] < + \infty$, $\EE[\| y\|^2] < + \infty$, $\EE[\| \gamma\|^2] < + \infty$, and $\EE[\| \lambda\|^2] < + \infty$}, it holds that
  \begin{equation*}
\begin{split}
&\EE  \Big [ \sum_{t=0}^{N-1} Q(z_{t+1}, z)  \Big ]  \\
 \leq {}& \sqrt{N} \Big ( 2Re^2 +  \frac{11C_f^2 }{8 C_h^2  }   +2 \EE [  \| \gamma_0 - \gamma \|^2 ]    + \EE [ \|\gamma  \|] \sigma_h
 + \frac{  \sigma_h^2 }{8 } + \frac{3 \EE[\| \gamma\|^2 ]  }{8}  +  2 C_h^2 \| x_0 -x \|^2  +  2 C_h^2 \EE [ \| x \|^2 ]       \Big ) 
 \\
 &  + \sqrt{N} \Big ( \frac{11 C_f^2 }{8 C_g^2 }  + 2 \EE [ \|  \lambda_0 - \lambda \|^2  ] + \EE [  \| \lambda \| ]\sigma_g
 + \frac{  \sigma_g^2 }{8 } + \frac{3 \EE[\| \lambda\|^2 ]  }{8}  + 2  C_g^2 \EE [ \| y_0 -y \|^2]  + 2 C_g^2 \EE [  \| y \|^2     ]     \Big ).
\end{split}
\end{equation*}
\end{theorem}

\subsubsection{Convergence Rates of Objective Optimality Gap, Duality Gap, and Feasibility Residuals} \label{sec:optimality_basic}
We denote the moving average of the iterates $\{x_t\}_{t=1}^N$ and $\{y_t\}_{t=1}^N$ as 
$$\bar x_N = \frac{1}{N}\sum_{t=1}^N x_t, \text{ and } \bar y_N = \frac{1}{N} \sum_{t=1}^N y_t,$$ respectively. Based on Theorem \ref{thm:boundsQ}, in the next theorem, we derive the convergence rate of the objective optimality gap $F(\bar x_N, y^*) - F(x^*,\bar y_N)$, and  feasibility residuals $\| H(\bar x_N)_+\|$ and $\| G(\bar y_N)_+ \|$.
\begin{theorem}\label{thm:rates_basic}
Suppose Assumption \ref{assumption:01} holds. 
Let $\{ (x_{t},y_{t},\gamma_{t},\lambda_{t}) \}$ be the sequence generated by Algorithm \ref{alg:1} with $\eta_t =4\sqrt{N} C_h^2$, $ \kappa_t = 4\sqrt{N} C_g^2$, and $\alpha_t  =  \beta_t =4\sqrt{N}$ for all $t=1,2,\cdots,N$. Let $R$ be the constant defined in \eqref{def:R}. For any  $(x,y) \in \{ x\in \cX \mid H(x) \le 0\}  \times \{y \in \cY\mid  G(y) \le 0\}$ satisfying $\EE[\|x\|^2] < +\infty$ and $\EE[\|y\|^2]<+\infty$, it holds that
\begin{equation*}\label{eq:objieq}
\begin{split}
\EE \Big [ F(\bar x_N,y )  - F(x,\bar y_N) \Big ] 
 \leq {}& \frac{1}{\sqrt{N}}  \Big ( 2Re^2 +  \frac{11C_f^2 }{8 C_h^2  }   +2   \| \gamma_0 \|^2    
 + \frac{  \sigma_h^2 }{8 }  +  2 C_h^2 \EE [\| x_0 -x \|^2]  +  2 C_h^2 \EE [ \| x \|^2 ]       \Big ) 
 \\
 & + \frac{1}{\sqrt{N} } \Big ( \frac{11 C_f^2 }{8 C_g^2 }  + 2  \|  \lambda_0  \|^2    + \frac{  \sigma_g^2 }{8 }  + 2  C_g^2 \EE [ \| y_0 -y \|^2]  + 2 C_g^2 \EE [  \| y \|^2     ]     \Big )
\end{split}
\end{equation*}
and 
\begin{equation*}
\begin{split}
& \EE [ \| H(\bar x_N)_+\|_2 ]  +     \EE [ \| G(\bar y_N)_+ \|_2 ]  \\
\leq {}&  \frac{1}{\sqrt{N}} \Big ( 2Re^2 +  \frac{11C_f^2 }{8 C_h^2  }   + 4  \| \gamma_0 \|^2 + \frac{35  }{8}  (\| \lambda^* \| +1)^2 +(\| \lambda^* \| +1) \sigma_h
 + \frac{  \sigma_h^2 }{8 }   +  2 C_h^2  \| x_0 -x^* \|^2  +  2 C_h^2 \| x^* \|^2        \Big ) 
 \\
 &  + \frac{1}{\sqrt{N}} \Big ( \frac{11 C_f^2 }{8 C_g^2 }  + 4  \| \lambda_0\|^2 + \frac{35}{8}  (\| \lambda^* \| +1)^2  + (\| \lambda^* \| +1) \sigma_g
 + \frac{  \sigma_g^2 }{8 }  + 2  C_g^2  \| y_0 -y^* \|^2  + 2 C_g^2  \| y^* \|^2       \Big )
 .
\end{split}
\end{equation*}
Moreover, there exist constants $C_1, C_2>0$ such that 
$$-\frac{C_1}{\sqrt{N}} \leq \EE \Big [ F(\bar x_N,y^* )  - F(x^*,\bar y_N) \Big ] \leq \frac{C_2}{\sqrt{N}}.$$
\end{theorem}


In the above result, we establish an $\cO(1/\sqrt{N})$ rate of convergence for both the objective optimality gap $F(\bar x_N,y^*) - F(x^*, \bar y_N)$ and feasibility residual $ \| H(\bar x_N)_+\|_2  +  \| G(\bar y_N)_+ \|_2$. \sg{This result matches the optimal convergent rate for standard convex stochastic optimization \cite[Proposition 14.1.1]{nemirovski1995information}, where projections are assumed to be computable through simple calculations.} 
We point out that our Basic-CSPD algorithm \ref{alg:1} achieves a comparable rate of convergence with the ConEx algorithm of \cite{boob2019stochastic} employing momentum techniques for smooth functions, despite  Algorithm \ref{alg:1} being able to accommodate the \emph{non-smoothness} in both objective and constraints, and the involvement of the ``maximization'' counterpart in our minimax setting would not  affect the rate of convergence adversely.
\sg{We shall also mention that it is unlikely to remove the dependencies of $\|\lambda^*\|$ and $\|\gamma^*\|$ in the bounds since they also appeared in lower bounds for related problems. For example, see \cite[Theorem 1.1]{ouyang2021lower} and  \cite[Theorem 3]{zhang2022solving}.}
Further, Theorem \ref{thm:rates_basic} implies that $F(\bar x_N, y) - F(x,\bar y_N)$ diminishes to zero at the rate of $\cO(1/\sqrt{N})$ for any feasible reference point $(x,y)$ {with bounded second moments}, \sg{demonstrating the flexibility of our approach.}
Here, under additional boundedness assumptions of the feasible regions $\widetilde \cX$ and $\widetilde \cY$, we provide one example that  characterizes the convergence behavior of the \emph{duality gap} as follows.
\begin{corollary}\label{cor:strong_gap_basic}
Suppose the conditions in Theorem \ref{thm:rates_basic} hold, and the feasible sets $\widetilde \cX:= \{ x \in \cX  \mid  H(x) \leq 0\}$ and $\widetilde \cY:=  \{ y \in  \cY \mid G(y) \leq 0 \}$ are bounded. Let $y^{\star}(\bar x_N) = \argmax_{ y \in \widetilde \cY}  F(\bar x_N,y)$
and $x^{\star}(\bar y_N) = \argmin_{x \in \tilde \cX } F(x,\bar y_N)$ be the best responses to $\bar x_N$ and $\bar y_N$, respectively. 
Then there exist constants $C_1, C_2>0$ such that 
$-\frac{C_1}{\sqrt{N}} \leq \EE \Big [ F(\bar x_N, y^{\star}(\bar x_N) )  - F(x^{\star}(\bar y_N),\bar y_N) \Big ] \leq \frac{C_2}{\sqrt{N}}.$
\end{corollary}


\begin{remark}\label{rmk:comparison}
    \sg{Here, we compare our Algorithm \ref{alg:1} with classic primal-dual methods \citep{nemirovski2009robust,lan2020first} for solving \eqref{prob:saddle_point}.
Indeed, in the convergence analysis of classic primal-dual methods for solving the convex-concave saddle problem \eqref{prob:saddle_point}, i.e.,
\[
\min_{x\in{\cal X}, \, \lambda \in \mathbb{R}_+^{m_2}} \max_{y\in {\cal Y}, \, \gamma\in \mathbb{R}_+^{m_1}} \left\{ \cL(x,y,\gamma, \lambda)  : =  F(x,y) + \gamma^\top  H(x)  -    \lambda^\top G(y) \right\},
\]
one typically requires $\cX, \cY$ to be compact, ${\cal L}$ to be Lipschitz continuous in $(x,y,\gamma,\lambda)$ and assumes bounded second moments of the sampled version of $\widetilde\nabla_{x,\lambda} \cL$ and $\widetilde\nabla_{y,\gamma} \cL$.
However, in our settings, due to the unboundedness of the constraint sets corresponding to $x, y, \gamma$, and $\lambda$, it is not difficult to observe that the aforementioned assumptions are rather strong and often fail to hold.

In contrast, for our analysis of Algorithm \ref{alg:1}, we only assume Lipschitz continuous and bounded second moments on the corresponding components of ${\cal L}$, i.e., the corresponding parts of $F$, $H$, and $G$ (see Assumption \ref{assumption:01} for more details). Under these much weaker and more realistic assumptions,
by choosing wisely on the step-sizes in Algorithm \ref{alg:1} and conducting a much more refined analysis, we derive strong results such as the uniform boundedness of $\{\EE[\|x_t\|^2], \EE[\|y_t\|^2], \EE[\|\lambda_t\|^2], \EE[\|\gamma_t\|^2]\}$ and the optimal convergence rates in terms of the objective value gap, feasibility residual, and duality gap.}
\end{remark}

\section{Primal-Dual Algorithm with Adaptive Step-sizes}\label{sec:adp_alg}
In the previous section, we propose a Basic-CSPD algorithm to tackle the minimax stochastic optimization problem and establish an $\cO(1/{\sqrt{N}})$ complexity result for the objective optimality  gap, duality gap, and feasibility residuals. One potential drawback of Basic-CSPD is that it requires prior knowledge of the total iterations $N$ to determine the step-sizes $\{ \eta_{t}, \kappa_t, \beta_t, \alpha_t \}$. Unfortunately, in many real-world scenarios, such as online streaming, where data are received sequentially,  the DM must interact with the stochastic environment constantly, and optimize their decisions after each interaction, without knowing the total number of iterations a priori.
In this case, the number of iterations $N$ is not prefixed so that our Algorihtm Basic-CSPD might not be directly applicable.
Therefore, the necessity of an algorithm adopting adaptive step-sizes to accommodate this practical concern emerges.

 To overcome this issue, in this section, we propose a modified algorithm, called the Adaptive Constrained Stochastic Primal-Dual (Adp-CSPD), which employs adaptive strategies to adjust step-sizes. In particular, in each iteration $t$, the step-sizes adopted by Adp-CSPD depend only on $t$ rather than the total number of iterations $N$. In this way, our refined algorithm has the flexibility to adjust the step-sizes over iterations. It thus  enables us to handle  practical concerns in a broad range of real-world scenarios using online streaming data. 

In the remainder of this section, we first present our Adp-CSPD algorithm.
We then analyze its convergence behavior. Without imposing \emph{any} additional assumptions, we prove that the primal and dual iterates $\{x_t, y_t, \gamma_t,\lambda_t\}$ generated by our algorithm are also uniformly bounded, and derive the rate of convergence for  the objective optimality gap, duality gap, and feasibility residuals.

\subsection{Algorithm Adp-CSPD}
One of the main difficulties when we try to incorporate adaptive step-sizes into our Basic-CSPD is to maintain the boundedness of the generated dual iterates. 
For this purpose, in iteration $t$ of Adp-CSPD, when updating the dual variable $\gamma_t$, we impose an additional majorization term $ \frac{\tau_t}{2} \| \gamma_0 - \gamma \|^2$, and conduct the following projected stochastic gradient step:
  \begin{equation}\label{eq:gamma_diminishing}
 \gamma_{t+1} =  \argmax_{ \gamma \in \RR_+^{m_1} } \Big \{h(x_{t},\xi_t^1)^\top \gamma - \frac{\beta_{t}}{2} \| \gamma_t - \gamma \|^2 - \frac{\tau_t}{2} \| \gamma_0 - \gamma \|^2 \Big \} .
 \end{equation}
By simple calculations, we can equivalently express $\gamma_{t+1}$ as
$$
\gamma_{t+1} = \Pi_{\RR_+^{m_1}} \Big \{  \frac{1}{\beta_t + \tau_t} \big (\beta_t \gamma_t + \tau_t \gamma_0 + h(x_{t},\xi_t^1) \big )\Big \}.
$$
Intuitively, the generated $\gamma_{t+1}$ is drawn towards a mid-point between $\gamma_0$ and $\gamma_t$ in each update. Since $\gamma_0$ is prefixed as the initial point, the dual sequence $\{ \gamma_t \}$ is less likely to be unbounded than the basic primal-dual algorithm employed in Section \ref{sec:basic}. By adopting a sophisticated choice of stepsizes $\beta_t$ and $\tau_t$, the involvement of this majorization term may facilitate the technical challenges in showing the boundedness of dual sequences, which would  provide further potentialities in designing more efficient algorithms with the desired properties, in particular, adaptive step-sizes. 

After updating $\gamma_t$, we impose an additional regularization term $\frac{\rho_t}{2} \| x - x_0\|^2$, and update $x_{t+1}$ by 
\begin{equation} \label{eq:x_diminishing} 
 x_{t+1} = \argmin_{x\in \cX}  \Big \{  \big( \widetilde \nabla_x f(x_t, y_t ,\omega_t^1)  + \sum_{i=1}^{m_1} \gamma_{t+1,i} \widetilde  \nabla h_i(x_t,\xi_t^2)\big) ^\top   x  + \frac{\eta_t}{2} \| x - x_t\|^2  + \frac{\rho_t}{2} \| x - x_0\|^2 \Big \}. 
 \end{equation}
Here, we note that if  $\rho_t$ decreases to zero sufficiently fast, the long-term effect of this majorization is minuscule and the convergence behavior of $\{ x_t\}$ would be affected negligibly. 

For the ``maximization'' counterpart, similar as the update steps of $x_t$ and $\gamma_t$, we query the $\cS\cO$ at $y_t$ to obtain $g(y_{t},\zeta_t^1) $, 
 and update the dual variable $\lambda_{t+1}$ by 
 $$
 \lambda_{t+1} =  \argmin_{ \lambda \in \RR_+^{m_2} } \Big \{ - g(y_{t},\zeta_t^1 )^\top \lambda  + \frac{\alpha_{t}}{2} \| \lambda_t - \lambda \|^2  +\frac{\nu_{t}}{2} \| \lambda_0 - \lambda \|^2   \Big \} .
 $$
Then,  we query the $\cS\cO$ at $(x_t,y_t)$ to obtain $\widetilde \nabla_y f(x_t, y_t ,\omega_t^2) $, query the $\cS\cO$ at $y_t$ to obtain $\widetilde \nabla g(x_t,\zeta_t^2)$, and update $y_{t+1}$ by 
 $$
y_{t+1} = \argmax_{y\in \cY}  \Big \{  \big (  \widetilde \nabla_y f(x_t, y_t ,\omega_t^2)  - \sum_{j=1}^{m_2}\lambda_{t+1,j} \widetilde \nabla g_j (y_t,\zeta_t^2 )  \big ) ^\top  y  - \frac{\kappa_t}{2} \| y - y_t \|^2 - \frac{\phi_t}{2} \| y - y_0\|^2 \Big \}. 
 $$
 In summary, the key idea of our Algorithm Adp-CSPD is to introduce additional majorization terms so that our primal and dual iterates converge in a better manner, which may allow us to employ a more flexible choice of step-sizes to accommodate our practical concern. We summarize the details of the Adp-CSPD algorithm in Algorithm \ref{alg:2}.

 \begin{algorithm}[t] \small
 \caption{Adaptive Constrained Stochastic Primal-Dual  (Adp-CSPD) for Constrained Minimax Optimization} \label{alg:2}
\begin{algorithmic}
 \STATE{\bfseries Input : }   Positive step-sizes  $\{ \alpha_t\}$,  $\{ \beta_t\}$,  $\{ \kappa_t\}$, $\{ \eta_t\}$, $\{ \tau_t\}$, $\{ \nu_t \}$, $\{ \rho_t \}$, $\{ \phi_t\}$, and initial points $(x_0, y_0, \gamma_0, \lambda_0)$\\
 \FOR{$t = 0, 1, 2, ..., N-1$}
 \STATE  Query the $\cS\cO$ at $x_t$ to obtain $h(x_{t},\xi_t^1) $, update the dual variable $\gamma_{t+1}$ by 
   $$
 \gamma_{t+1} =  \argmax_{ \gamma \in \RR_+^{m_1} } \Big \{h(x_{t},\xi_t^1)^\top \gamma - \frac{\beta_{t}}{2} \| \gamma_t - \gamma \|^2  - \frac{\tau_t}{2} \| \gamma_0 - \gamma \|^2  \Big \} .
 $$
 Query the $\cS\cO$ at $y_t$ to obtain $g(y_{t},\zeta_t^1) $, 
 update the dual variable $\lambda_{t+1}$ by 
 $$
 \lambda_{t+1} =  \argmin_{ \lambda \in \RR_+^{m_2}} \Big \{ - g(y_{t},\zeta_t^1)^\top \lambda  + \frac{\alpha_{t}}{2} \| \lambda_t - \lambda \|^2 +\frac{\nu_{t}}{2} \| \lambda_0 - \lambda \|^2   \Big \} .
 $$
 Query the $\cS\cO$ at $(x_t,y_t)$ to obtain $\widetilde \nabla_x f(x_t, y_t ,\omega_t^1) $ and $\widetilde \nabla h(x_t,\xi_t^2)$,
 update $x_{t+1}$ by 
 $$
 x_{t+1} = \argmin_{x\in \cX}  \Big \{  \big( \widetilde \nabla_x f(x_t, y_t ,\omega_t^1)  + \sum_{i=1}^{m_1}\ \gamma_{t+1,i} \widetilde  \nabla h_i(x_t,\xi_t^2)\big) ^\top   x  + \frac{\eta_t}{2} \| x - x_t\|^2   + \frac{\rho_t}{2} \| x - x_0\|^2 \Big \}. 
 $$
 Query the $\cS\cO$ at $(x_t,y_t)$ to obtain $\widetilde \nabla_y f(x_t, y_t ,\omega_t^2) $ and $\widetilde \nabla g(y_t,\zeta_t^2)$, update $y_{t+1}$ by 
 $$
y_{t+1} = \argmax_{y\in \cY}  \Big \{  \big (  \widetilde \nabla_y f(x_t, y_t ,\omega_t^2)  - \sum_{j=1}^{m_2}\ \lambda_{t+1,j} \widetilde \nabla g_j (y_t,\zeta_t^2)  \big ) ^\top  y  - \frac{\kappa_t}{2} \| y - y_t\|^2  - \frac{\phi_t}{2} \| y - y_0\|^2  \Big \}. 
 $$
\ENDFOR
\STATE{\bfseries Output: } $\bar x_N = \frac{1}{N} \sum_{t=1}^N x_t$ and $\bar y_N = \frac{1}{N} \sum_{t=1}^N y_t$.
  \end{algorithmic}
\end{algorithm}

 \subsection{Convergence Analysis} In the remainder of this section, we characterize the iterative performance of Adp-CSPD by following similar steps to Section \ref{sec:basic}. We assume Assumptions \ref{assumption:01} and \ref{assumption:02} hold throughout this section and start our analysis by bounding the error term $\sum_{t=0}^{K-1}Q(z_{t+1},z)$ for all $1\le K \le N$. 
\sg{
  \begin{lemma}\label{lem:delta_gamma_diminish}
  Let $\{ z_t = (x_t,y_t,\gamma_t,\lambda_t)\}_{t=1}^N$ be the sequence generated by Algorithm \ref{alg:2} with positive step-sizes $\alpha_t, \beta_t, \nu_t, \tau_t, \eta_t, \rho_t, \kappa_t, \phi_t$ satisfying $\beta_t + \tau_t \geq \beta_{t+1}$, $\eta_t + \rho_t \ge \eta_{t+1}$, $\alpha_t + \nu_t \ge \alpha_{t+1}$ and $\kappa_t + \phi_t \ge \kappa_{t+1}$.  Then for any $z=(x,y,\gamma, \lambda) \in \cX \times\cY \times \Re_+^{m_1}\times \Re_+^{m_2}$ and $1 \le K \le N$, it holds that
\begin{equation*}
\begin{aligned}
  &\sum_{t=0}^{K-1} Q(z_{t+1},z) + \frac{\eta_{K}}{2} \norm{x_K - x}^2 + \frac{\beta_{K}}{2}\norm{\gamma_K - \gamma}^2 + \frac{\kappa_{K}}{2}\norm{y_K - y}^2 + 
  \frac{\alpha_{K}}{2} \norm{\lambda_K - \lambda}^2 
  \\  
\le & (\frac{\eta_0}{2} + \sum_{t=0}^{K-1} \frac{\rho_t}{2}) \norm{x_0 - x}^2 + (\frac{\beta_0}{2} + \sum_{t=0}^{K-1} \frac{\tau_t}{2}) \norm{\gamma_0 - \gamma}^2 + \sum_{t=0}^{K-1} \big(\frac{3 C_h^2\norm{\gamma}^2}{2\eta_t}- \frac{\tau_t}{2}\|\gamma_{t+1} - \gamma_0\|^2 \big) + \sum_{t=0}^{K-1} 
U_t(x,\gamma) \\
& + (\frac{\kappa_0}{2} + \sum_{t=0}^{K-1} \frac{\phi_t}{2}) \norm{y_0 - y}^2 + (\frac{\alpha_0}{2} + \sum_{t=0}^{K-1} \frac{\nu_t}{2}) \norm{\lambda_0 - \lambda}^2 + \sum_{t=0}^{K-1} \big( \frac{3 C_g^2\norm{\lambda}^2}{2\kappa_t} - \frac{\nu_t}{2}\|\lambda_{t+1} - \lambda_0\|^2 \big) + \sum_{t=0}^{K-1} V_t(y,\lambda),
\end{aligned}
\end{equation*}
where $U_t$ and $V_t$ are defined in \eqref{eq:Uxgamma} and \eqref{eq:Vylambda}, respectively. 
\end{lemma}

Similar to Lemma \ref{lem:EUV}, we provide the following upper bounds for $\sum_{t=0}^{K-1} U_t(x,\gamma)$ and $\sum_{t=0}^{K-1}  V_t(y,\lambda)$  for any $(x,y,\gamma,\lambda) \in \cX \times \cY \times \RR_+^{m_1} \times \RR_+^{m_2}$ with bounded second moments.  
\begin{lemma}
    \label{lem:EUVadp}
    Under the same settings as in Lemma \ref{lem:delta_gamma_diminish}, for any  {$(x,y,\gamma, \lambda)\in \cX \times \cY \times \RR_+^{m_1} \times \RR_+^{m_2}$ satisfying $\EE[\| x\|^2] < + \infty$, $\EE[\| y\|^2] < + \infty$, $\EE[\| \gamma\|^2] < + \infty$, and $\EE[\| \lambda\|^2] < + \infty$}, it holds that
    \begin{equation*}
        \begin{aligned}
            &\EE\Big[ \sum_{t=0}^{K-1} U_t(x,\gamma) \Big] \le (\frac{\eta_0}{2} + \sum_{t=0}^{K-1} \frac{\rho_t}{2}) \EE[\|x\|^2] + \sum_{t=0}^{K-1} \frac{4}{\eta_t}\big(C_f^2 + C_h^2 \EE[\|\gamma_{t+1}\|^2]\big) + \sqrt{K}\EE[\|\gamma\|]\sigma_h + \sum_{t=0}^{K-1}\frac{\sigma_h^2}{2\beta_t},  \\
        &\EE\Big[ \sum_{t=0}^{K-1}  V_t(y,\lambda) \Big] \le(\frac{\kappa_0}{2} + \sum_{t=0}^{K-1} \frac{\phi_t}{2})\EE[\|y\|^2] + \sum_{t=0}^{K-1} \frac{4}{\kappa_t}\big(C_f^2 + C_g^2 \EE[\|\lambda_{t+1}\|^2]\big) + \sqrt{K}\EE[\|\lambda\|]\sigma_g + \sum_{t=0}^{K-1}\frac{\sigma_g^2}{2\alpha_t} .
        \end{aligned}
    \end{equation*}
\end{lemma}
}

To further investigate the convergence behavior of Adp-CSPD, we employ a sequence of step-sizes of the following form:
\begin{equation}\label{eq:adaptive_stepsize}
\begin{split}
\text{update $\gamma_t$:}  \, \, \, \,   \, \, \, \, \, \, \, \,  \, \, \, \,  \, \, \, \,   \beta_t = C_h^2 \sqrt{t+1}, & \ \ \tau_t = C_h^2( \sqrt{t+2} - \sqrt{t+1}), \\
\text{update $\lambda_t$:}  \, \, \, \,   \, \, \, \,  \, \, \, \, \, \, \, \,  \, \, \, \,  \alpha_t = C_g^2 \sqrt{t+1}, & \ \ \nu_t = C_g^2( \sqrt{t+2} - \sqrt{t+1}), \\ 
\text{update $x_t$:} \, \, \, \,   \, \, \, \, \, \, \, \,  \, \, \, \,  \, \, \, \,     \eta_t =  16\sqrt{t+2}, & \ \  \rho_t = 16(\sqrt{t+3} - \sqrt{t+2}), \\
\text{update $y_t$:}  \, \, \, \,   \, \, \, \,  \, \, \, \, \, \, \, \,  \, \, \, \,    \kappa_t = 16\sqrt{t+2}, & \ \  \phi_t = 16(\sqrt{t+3} - \sqrt{t+2}) .
\end{split}
\end{equation}
We note that the above setting of the step-sizes satisfies the assumption made in Lemma  \ref{lem:delta_gamma_diminish}. With this and Lemmas \ref{lem:delta_gamma_diminish} and \ref{lem:EUVadp}, we obtain in the following theorem a useful bound of the gap function $Q(z_t, z)$ for any $z$ with bounded second moments. 
\begin{theorem}\label{thm:diminishing_1}
Choose $\gamma_0 = \bf{0}$ and $\lambda_0 = \bf{0}$ and let $\{ z_t = (x_t,y_t,\gamma_t,\lambda_t)\}_{t=1}^N$ be the sequence generated by Algorithm \ref{alg:2} with step-sizes given in \eqref{eq:adaptive_stepsize}. For any  $z = (x,y,\gamma,\lambda) \in \cX \times \cY \times \RR_+^{m_1} \times \RR_+^{m_2}$ satisfying $\EE[\| x\|^2] < + \infty$, $\EE[\| y\|^2] < + \infty$, $\EE[\| \gamma\|^2] < + \infty$, and $\EE[\| \lambda\|^2] < + \infty$, and $1\le K\le N$ it holds that
\begin{equation*}
\begin{split}
&\EE\Big[\sum_{t=0}^{K-1}  Q(z_{t+1}, z) \Big]  +  \frac{
 C_h^2 \sqrt{K+1} }{2} \EE[  \| \gamma_{K} - \gamma\|^2 ]  
  + \frac{C_g^2 \sqrt{K+1}}{2} \EE [\| \lambda_K - \lambda\|^2] \\
  & + 8\sqrt{K+2}\big(\EE[\|x_K - x\|^2] + \EE[\|y_K - y\|^2]\big) \\
\leq & \sqrt{K+2} \Big (  \frac{11C_h^2}{16}  \EE[\| \gamma\|^2]  
+ \frac{11C_g^2}{16}  \EE [ \| \lambda\|^2 ] \Big )  + \sqrt{K} \Big ( \EE[\|\gamma\|] \sigma_h + 
\EE[\|\lambda \|] \sigma_g \Big ) + \sqrt{K+1}     \Big ( \frac{ \sigma_h^2 }{ C_h^2 }  +  \frac{ \sigma_g^2 }{ C_g^2 } \Big ) 
\\
&  +  \sqrt{K+2} \Big ( 8\EE [ \|x - x_0 \|^2]  + 8\EE [ \| x\|^2 ]  +  \frac{11   C_f^2     }{16}  +  8 \EE [ \|y - y_0 \|^2]  + 8\EE [ \| y \|^2 ]+  \frac{11   C_f^2     }{16}  \Big ) .
\end{split}
\end{equation*} 
\end{theorem}
{We defer the detailed proof to Appendix Section \ref{sec:proof_of_thm_diminishing_1}.}
\sg{
\begin{remark}
It is not difficult to obtain the uniform boundedness of the second moments of the  primal and dual iterates. Indeed, let $z$ be chosen as any saddle point $(x^*,y^*,\gamma^*,\lambda^*)$ and set 
\begin{align*}
\tilde R:= {}&  \frac{15C_h^2}{16}  \EE[\| \gamma^*\|^2]  
+ \frac{15C_g^2}{16}  \EE [ \| \lambda^*\|^2 ]  + \sqrt{K} \Big ( \EE[\|\gamma^*\|] \sigma_h + 
\EE[\|\lambda^*\|] \sigma_g \Big ) +     \frac{ \sigma_h^2 }{ C_h^2 }  +  \frac{ \sigma_g^2 }{ C_g^2 } 
\\
&  +   8\EE [ \|x^* - x_0 \|^2]  + 16\EE [ \| x^*\|^2 ]  +  \frac{11   C_f^2     }{16}  +  8 \EE [ \|y^* - y_0 \|^2]  + 16\EE [ \| y^* \|^2 ]+  \frac{11   C_f^2     }{16}.
\end{align*}
Then, it is not difficult to observe from Theorem \ref{thm:diminishing_1} that 
\begin{align*}
    \frac{
 C_h^2 }{8} \EE[  \| \gamma_{K}\|^2 ]  
  + \frac{C_g^2}{8} \EE [\| \lambda_K \|^2] + 4 \big(\EE[\|x_K \|^2] + \EE[\|y_K\|^2]\big) \le \tilde R, \quad \forall\, 1\le K\le N.
\end{align*}
\end{remark}
}

\subsection{Convergence Rates of Objective Optimality Gap, Duality Gap, and Feasibility Residuals} \label{sec:optimality_adaptive}
By using Theorem \ref{thm:diminishing_1}, we derive the convergence rates of both objective optimality gap and feasibility residuals in the next result. 
\begin{theorem} \label{thm:rates_adp} 
Choose $\gamma_0 = \bf{0}$ and $\lambda_0 = \bf{0}$ and let $\{ (x_t,y_t,\gamma_t,\lambda_t)\}_{t=1}^N$ be the sequence generated by Algorithm \ref{alg:2} with step-sizes given in \eqref{eq:adaptive_stepsize}.  Let $\bar x_N = \frac{1}{N}\sum_{t=1}^N x_t$, $\bar y_N = \frac{1}{N}\sum_{t=1}^N y_t$. Then, it holds that 
\begin{equation*}
\begin{split}
& \EE [F(\bar x_N,y)  - F(x,\bar y_N)]  \\
\leq {}&  \frac{ \sqrt{N+1} }{N}    \Big ( \frac{ \sigma_h^2 }{ C_h^2 }  +  \frac{ \sigma_g^2 }{ C_g^2 } \Big ) 
 +  \frac{ \sqrt{N+2}  }{N}\Big ( 8\EE [ \|x - x_0 \|^2+ \| x\|^2] +  \frac{11   C_f^2     }{16}  +  8 \EE [ \|y - y_0 \|^2  + \| y \|^2 ]+  \frac{11   C_f^2     }{16}  \Big )
 \end{split}
\end{equation*}
and
\begin{equation*} 
\begin{split}
& \EE [ \| H(\bar x_N)_+\|_2 ]  +     \EE [ \| G(\bar y_N)_+ \|_2 ] \\
 \leq {}& \frac{ \sqrt{N+2} }{N}\Big (  \frac{11C_h^2}{16}  (\|  \gamma^*\| + 1 )^2  
+ \frac{11C_g^2}{16}  ( \| \lambda^*\| + 1)  \Big )  + \frac{1}{ \sqrt{N} } \Big ( (\|  \gamma^*\| + 1 )  \sigma_h + 
 ( \| \lambda^*\| + 1)  \sigma_g \Big ) 
\\
& 
\quad + \frac{\sqrt{N+1}}{N}     \Big ( \frac{ \sigma_h^2 }{ C_h^2 }  +  \frac{ \sigma_g^2 }{ C_g^2 } \Big )  +  \frac{ \sqrt{N+2} }{N} \Big ( 8 \|x^* - x_0 \|^2  +  8\| x^* \|^2 +  \frac{11   C_f^2     }{16}  +  8 \|y^* - y_0 \|^2  + 8 \| y^* \|^2 +  \frac{11   C_f^2     }{16}  \Big ).
\end{split}
\end{equation*} 
Moreover, there exist constants $C_1, C_2>0$ such that 
$$-\frac{C_1}{\sqrt{N}} \leq \EE \Big [ F(\bar x_N,y^* )  - F(x^*,\bar y_N) \Big ] \leq \frac{C_2}{\sqrt{N}}.$$
\end{theorem}

\begin{remark}
The above result implies that our  Adp-CSPD algorithm can still achieve an optimal $\cO(1/{\sqrt{N}})$ rate of convergence for both objective optimality gap and feasibility residuals, with an adaptive choice of step-sizes that does not require the total number of iterations to be fixed in advance. This establishes a benchmark for expectation constrained stochastic minimax optimization. 
\end{remark}
Similar to Corollary \ref{cor:strong_gap_basic}, we 
establish the convergence rate of the duality gap for Adp-CSPD algorithm in the following corollary.
\begin{corollary}
Suppose the conditions in Theorem \ref{thm:rates_adp} hold, and the feasible regions $\widetilde \cX:= \{ x \in \cX  \mid  H(x) \leq 0\}$ and $\widetilde \cY:=  \{ y \in \widetilde \cY \mid G(y) \leq 0 \}$ are bounded. Let $y^{\star}(\bar x_N) = \argmax_{ y \in \widetilde \cY}  F(\bar x_N,y)$
and $x^{\star}(\bar y_N) = \argmin_{x \in \tilde \cX } F(x,\bar y_N)$ be the best responses to $\bar x_N$ and $\bar y_N$, respectively. 
Then there exist constants $C_1, C_2>0$ such that 
$-\frac{C_1}{\sqrt{N}} \leq \EE \Big [ F \big (\bar x_N, y^{\star}(\bar x_N) \big )  - F \big (x^{\star}(\bar y_N),\bar y_N \big ) \Big ] \leq \frac{C_2}{\sqrt{N}}.$
\end{corollary}
\section{Numerical Experiments}
In this section, we conduct empirical studies of our proposed algorithms. Specifically, we evaluate our algorithms to solve {three} problems:
 a  quadratic-constrained quadratic saddle point optimization, a robust pricing problem with a large number of expectation constraints, {and AUC maximazation with fairness constraints}. All these problems are formulated as expectation constrained minimax optimization problems. Detailed settings of our experiments are also discussed.
\subsection{Quadratic-Constrained Quadratic Saddle Point Optimization}
The first numerical example we consider is a \emph{quadratic-constrained quadratic saddle point optimization problem} taking the following form
\begin{equation}\label{numeric:qp}
\begin{split}
 \min_{x \in \cX} & \max_{y \in \cY}   \ \EE_{\omega}[ f_1(x,\omega)] + x^\top y\\
  \mbox{ s.t. }  &  \EE_{\xi_i}[ h_i(x,\xi_i)] \leq 0, \, i =1,2,\cdots,m,
\end{split}
\end{equation}
where $f_1$ and $h_i$, $i=1,\ldots, m$, are  quadratic functions.
In the experiments, we consider the quadratic objective function  $f_1(x,\omega) = (x - \tilde x_0 )^\top Q  (x - \tilde x_0 ) + x^\top \omega $ where $\tilde x_0, Q$ are fixed  components and $\omega \sim \text{Unif}[0,1]^d$ is a random vector and we consider the following least-square constraint: 
$$
h_j(x,\xi_j) = ((x - \tilde x_j)^\top s_j  + \xi_j )^2 - \theta_j,
$$
where $ \xi_i$ is a standard Gaussian noise and $\tilde x_j, s_j, \theta_j$ are given data. The convex set $\cX$ is chosen to be the full space $\RR^d$ and the convex set $\cY$ is set to be an ellipsoid:
\begin{equation*}
    \cY = \Big \{ y : (y - y_0)^\top  M  (y - y_0) \leq  c_{y}^2 \Big \},
\end{equation*}
where $M \succ 0$ is a given positive definite matrix. 
Note that by maximizing $y\in \cY$, problem \eqref{numeric:qp} can be recast in the following form
\begin{equation}\label{numeric:qp_1}
\begin{split}
\min_{x \in \cX} & \ \EE[ f_1(x,\omega)] + x^\top y_0 + c_y \sqrt{x^\top M^{-1} x}\\
\mbox{ s.t. } & \EE_{\xi_i}[ h_i(x,\xi_i)] \leq 0, \, i =1,2,\cdots,m.
\end{split}
\end{equation}
The above problem can be rewritten as a second-order-cone program if the randomness in $f_1(x,\omega)$ and each $h_i(x,\xi)$ is absent. 
Hence, to obtain a highly accurate optimal solution {$(x^*,y^*)$} to \eqref{numeric:qp}, we simulate $10^5$ independent samples of $\omega$ and $\xi$, and solve the batched version of problem \eqref{numeric:qp_1} using CVX solver \citep{cvx}.

\textbf{Implementation Details.}
In our experiments, we generated synthetic test instances using the following steps:
(1) set $d = 50$ and $m = 15$; (2) generate $x_0 \in \RR^d$, $L\in \RR^{d\times d}$, where each entry of $x_0$ is i.i.d. sampled from $\cN(0,0.3)$ and each entry of $L$ is i.i.d. sampled from $\cN(0,1)$; set $Q = LL^T + I_d$; (3) for $j=1,\ldots,m$, sample $\tilde{x}_j \in \RR^d$ from $\cN(0, I_d)$ and generate $s_j\in \RR^d$ where each entry of $s_j$ is i.i.d. sampled from  $\text{Unif}[0,1]$; (4) set $M = I_d$ and $c_y = 1$.

We evaluate the performance of our algorithms over two settings where (i) the optimal solution falls in the interior of the feasible set and (ii) at the boundary. To do so, we first solve the unconstrained version of Prob.\eqref{numeric:qp} and obtain its optimal solution $\widehat x^*$. Then, we evaluate each constraint at $\widehat x^*$ and set $\widehat \theta_j = \EE_{\xi_j}[ ((\widehat x^* - \tilde x_j)^\top w_j  + \xi_j )^2  ]$. We set $\theta_j = 1.2 \times \widehat \theta_j $ to study the behavior when the optimum falls in the interior and set $\theta_j = 0.9 \times \widehat \theta_j $ to investigate the scenario where the optimum falls at the boundary.

In each setting, we conduct 10 independent simulations.  In  each simulation, we run basic-CSPD with different prefixed total numbers of iterations  
$N \in \{10^4,2\times 10^4,6\times 10^4,15\times 10^4,45\times 10^4,5\times 10^5\}$,
 with  step-sizes  $ \alpha_t = 500\sqrt{N}$ and $  \eta_t = \kappa_t = 30\sqrt{N}$  for $t =0,1,\ldots, N$; we run Adp-CSPD for $N_{\text{total}} = 5\times 10^5$ iterations with step-sizes  $\alpha_t = \beta_t = 500 \sqrt{t+1}$, $ \nu_t = 500 (\sqrt{t+2} - \sqrt{t+1})$, $\eta_t = \kappa_t = 30\sqrt{t+2}$, and $\rho_t = \phi_t = 30(\sqrt{t+3} - \sqrt{t+2})$ for $t = 0,1,2,\cdots, N_{\text{total}}$.

\begin{figure}[t]
\vskip 0.5cm
\begin{minipage}{0.31\textwidth}
    \centering
        \includegraphics[width=1\linewidth]{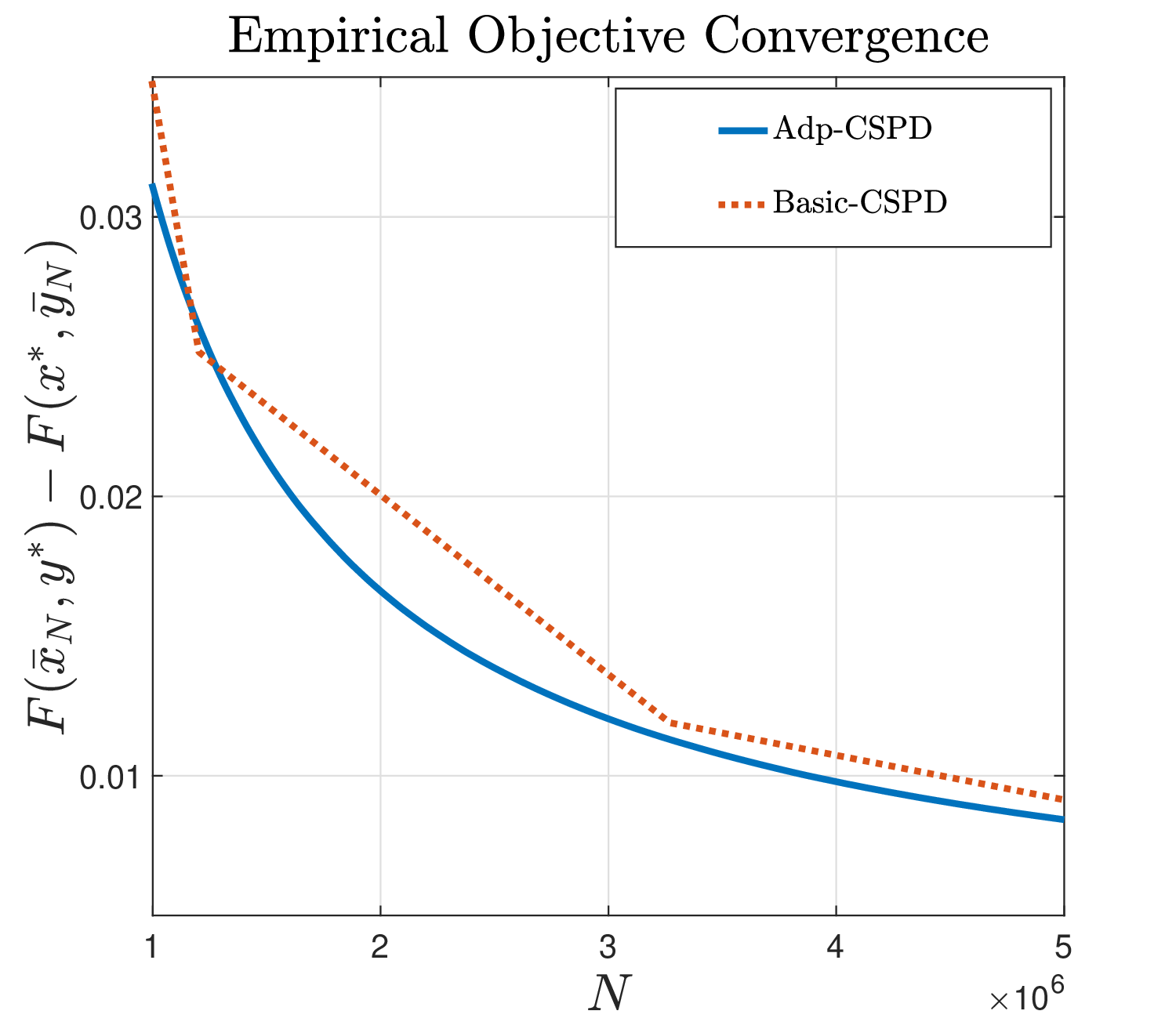}
        (a)
        \end{minipage}
        \begin{minipage}{0.31\textwidth}
        \centering
    \includegraphics[width=1\linewidth]{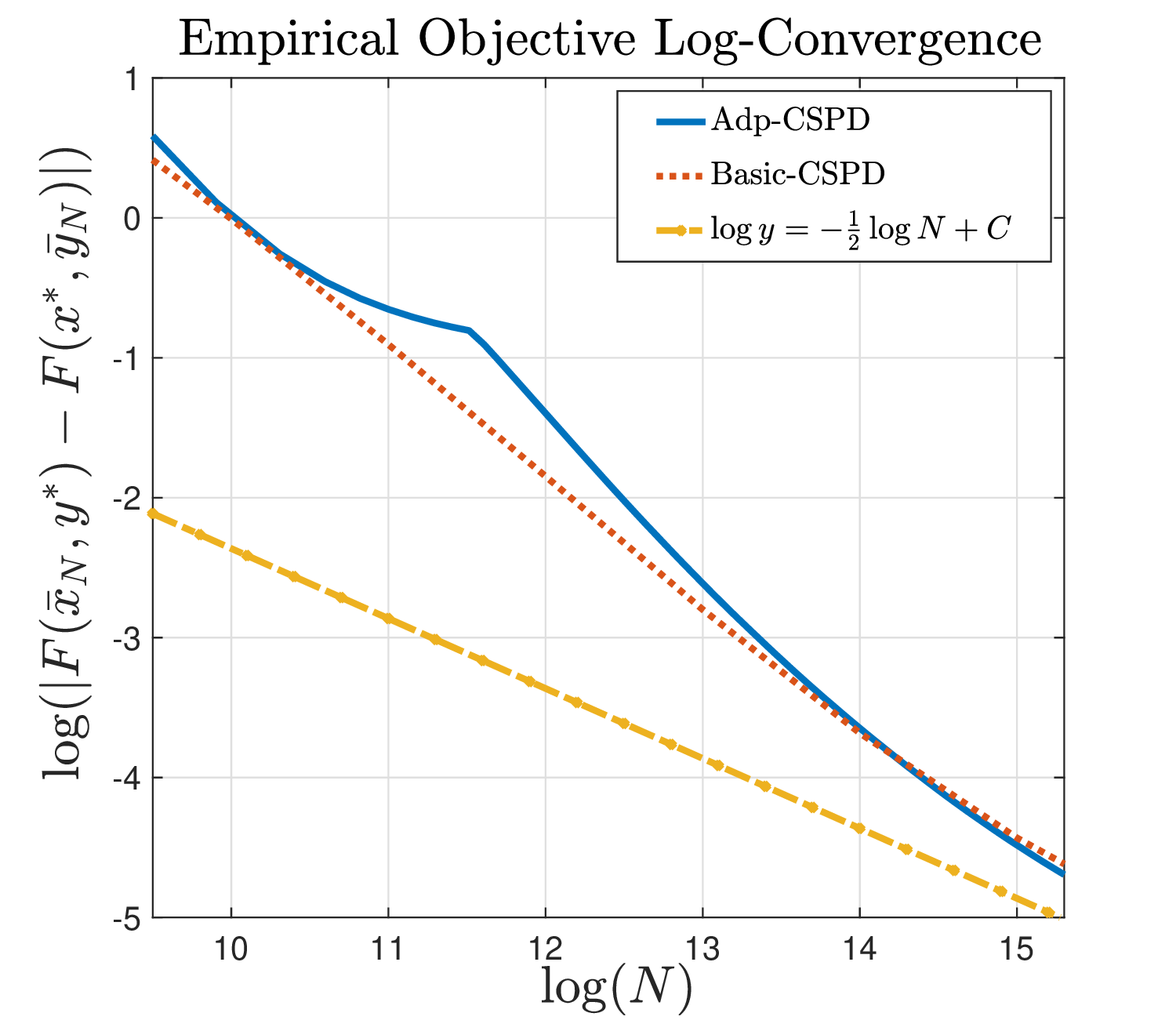}
    (b)
    \end{minipage}
  \begin{minipage}{0.31\textwidth}
  \centering
    \includegraphics[width=1\linewidth]{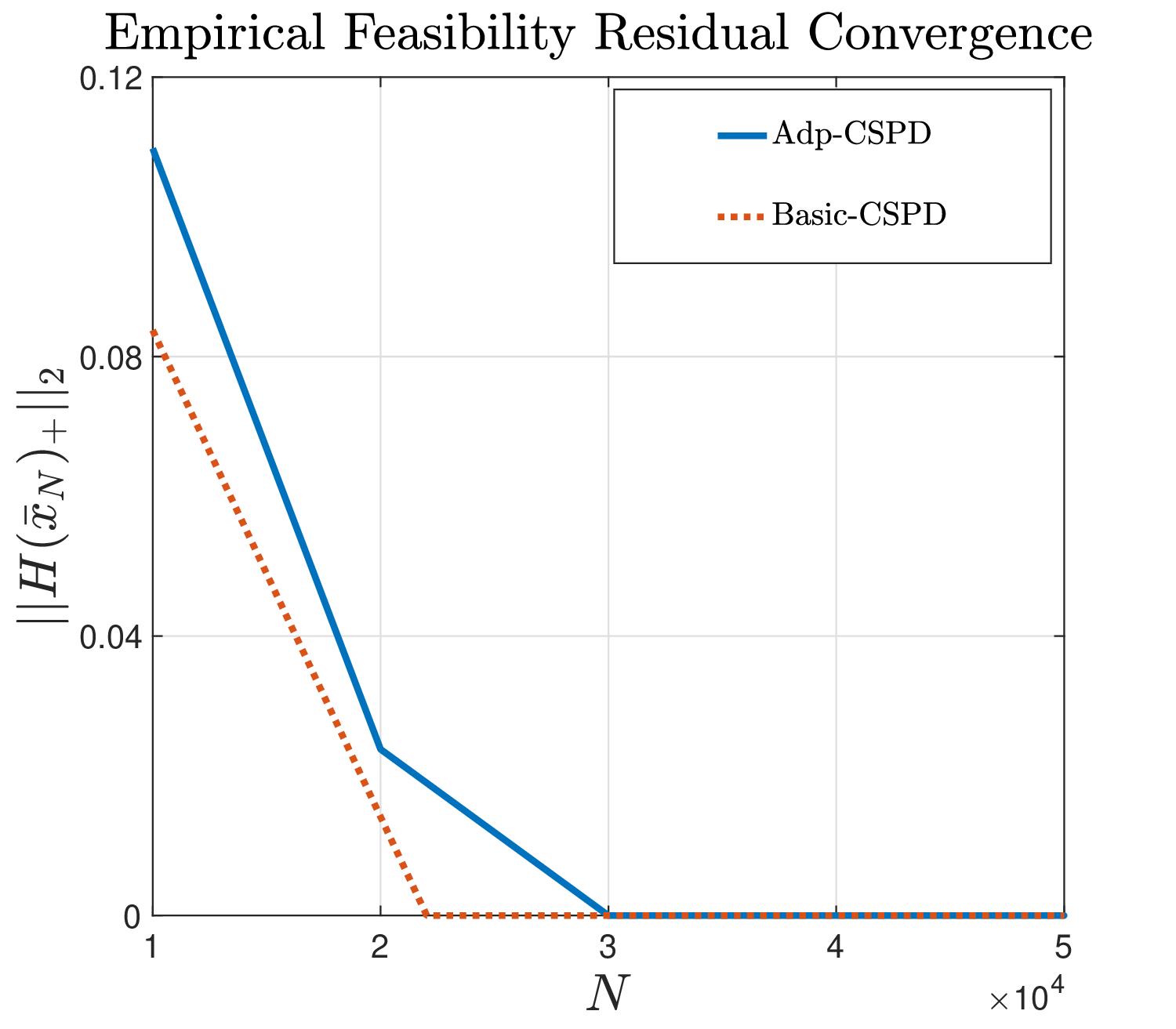}
    (c)
    \end{minipage}
    \caption{ Empirical convergence rate of the objective gap $F(\bar x_N, y^*) - F(x^*,\bar y_N)$ and feasibility residual $\| H(\bar x_N)_+ \|_2$ when the optimal solution falls in the interior of the feasible region. }
    \label{fig:obj_interior}
\end{figure}

\begin{figure}[t]
\vskip 0.5cm
    \centering
        \includegraphics[width=0.4\linewidth]{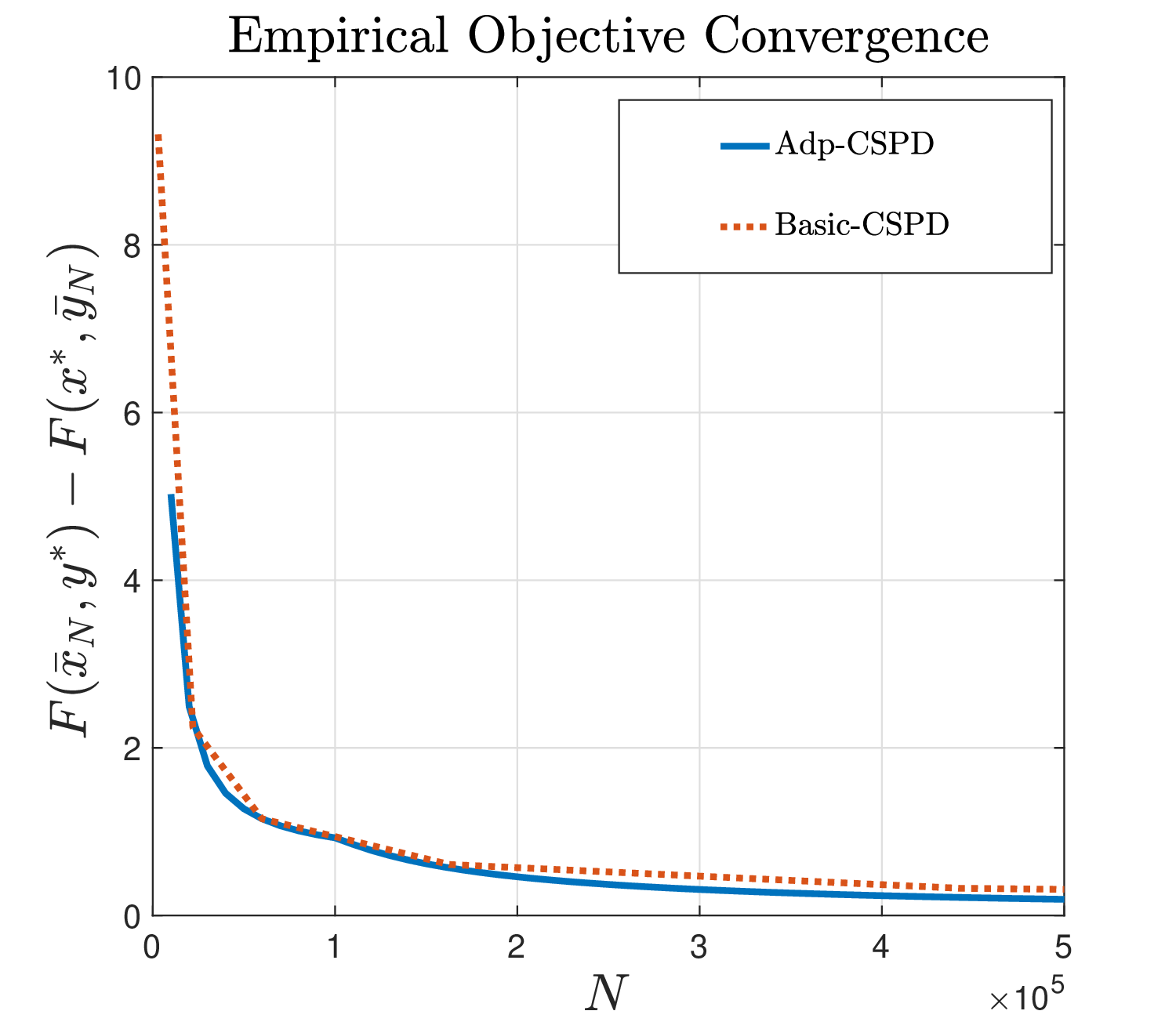}
    \includegraphics[width=0.4\linewidth]{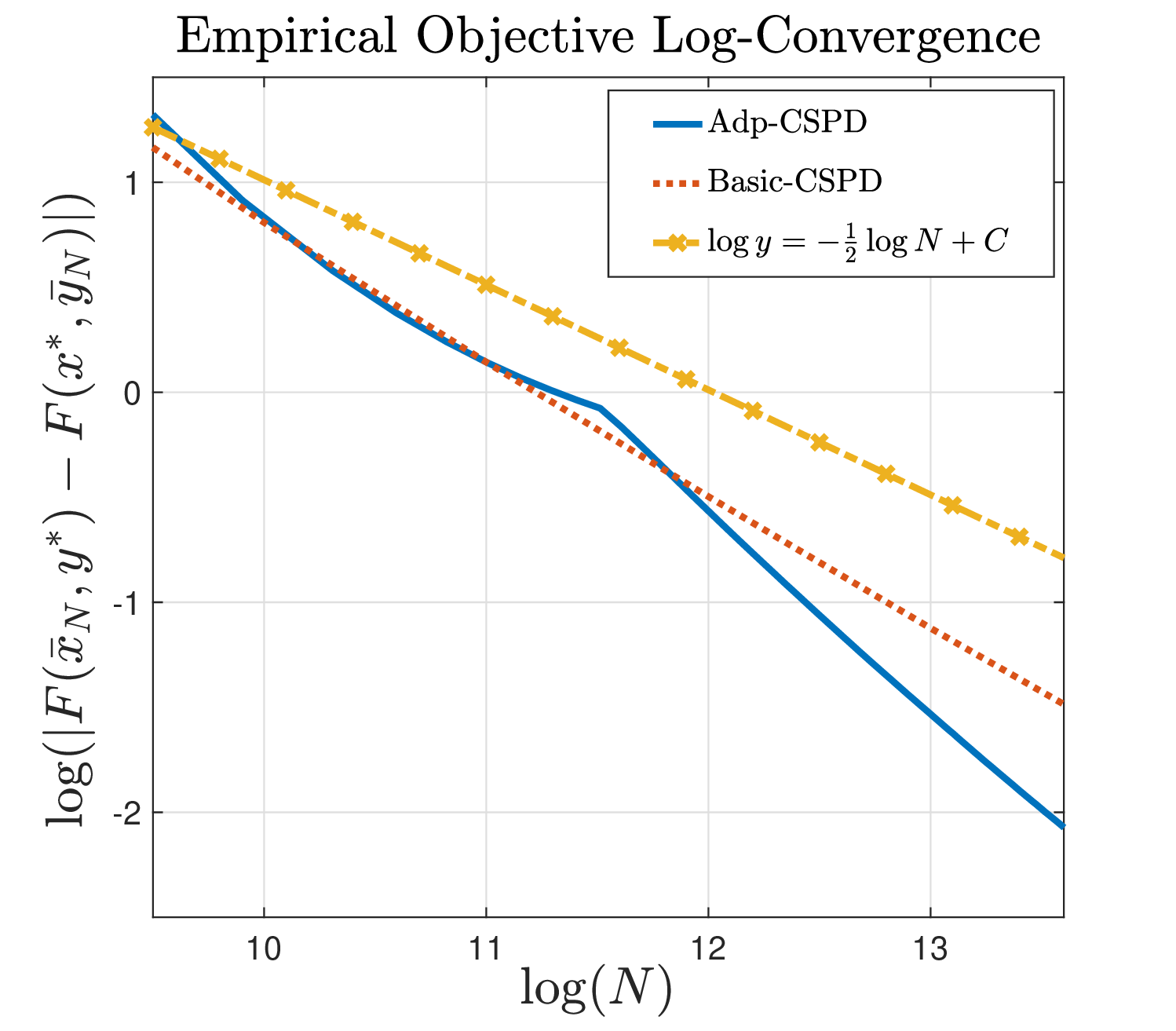}
    \caption{Empirical convergence rate of the objective gap $F(\bar x_N, y^*) - F(x^*,\bar y_N)$ when the optimal solution falls at the boundary of the feasible region. }
    \label{fig:obj_boundary}
\end{figure}

\begin{figure}[t]
\vskip 0.5cm
    \centering
    \includegraphics[width=0.4\linewidth]{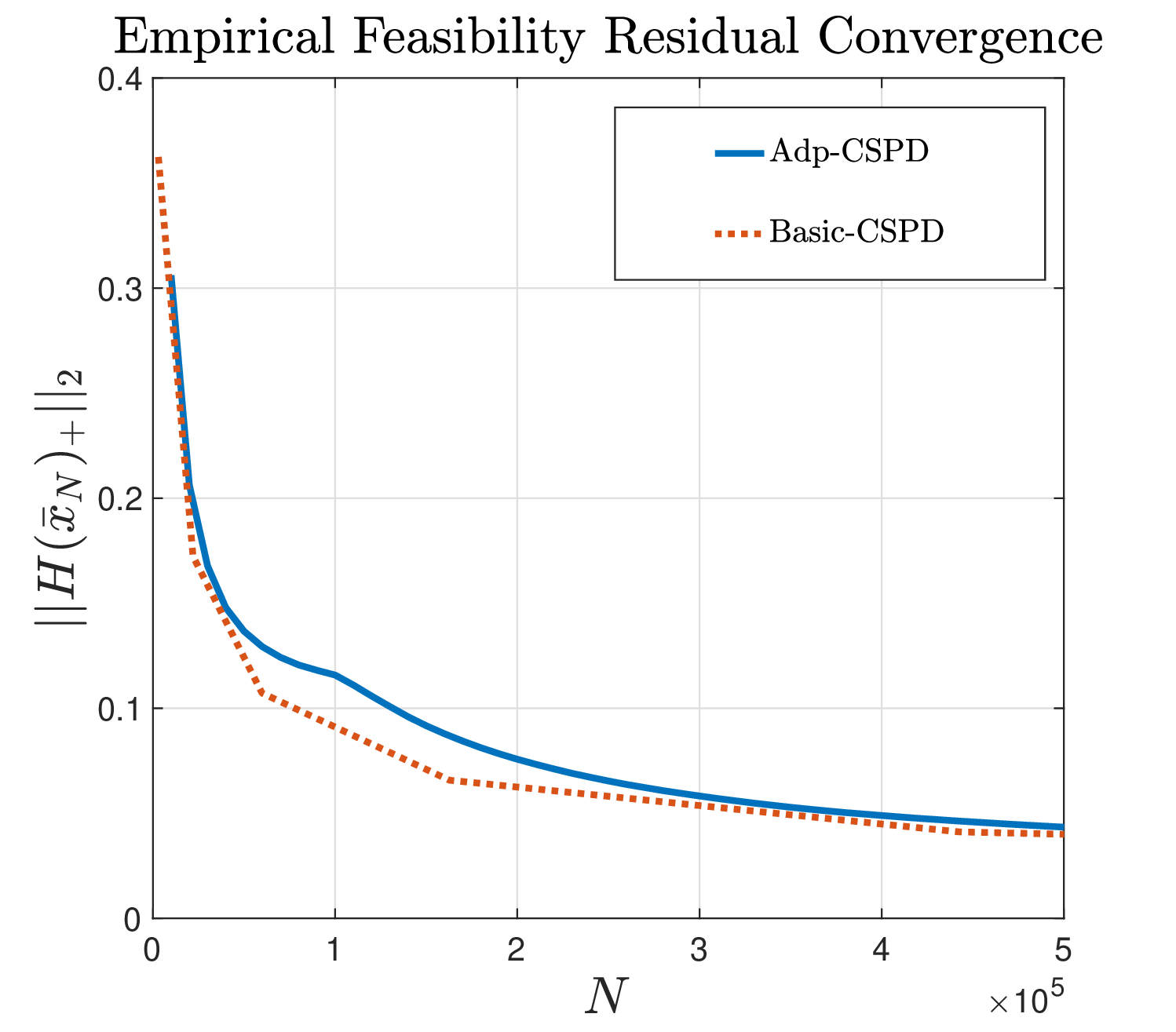}
    \includegraphics[width=0.4\linewidth]{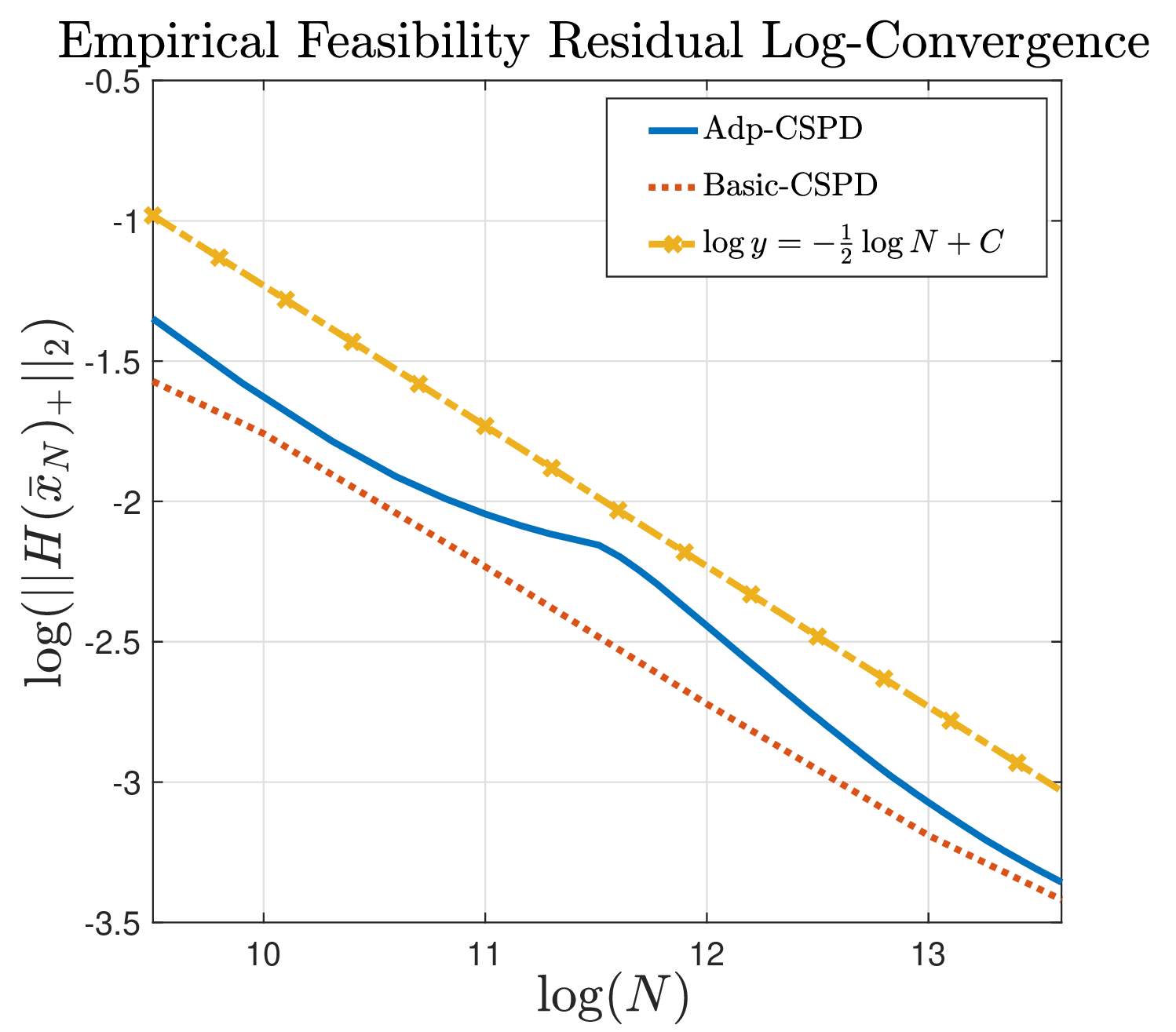}
    \caption{Empirical convergence rate of the feasibility residual $\| H(\bar x_N)_+ \|_2$ when the optimal solution falls at the boundary of feasible region. }
    \label{fig:infeasibility_boundary}
\end{figure}

{Let $(\bar x_N,\bar y_N)$ be the solution returned by the tested algorithms over $N$ iterations.} We report below the detailed numerical results for each setting. 
\\
\noindent(i) \emph{Optimal Solution in the Interior. } 
 We plot the averaged objective  gap $F(\bar x_N, y^*) - F(x^*,\bar y_N)$, and plot $\log N$ against $\log \big ( |F(\bar x_N, y^*) - F(x^*,\bar y_N)| \big )$. We summarize the results in Figure \ref{fig:obj_interior}(a)-(b), and provide one line of slope $-1/2$ for benchmark comparison when reporting the log-convergence.  Meanwhile, to study the performance of the tested algorithm in terms of the infeasibility, we plot the averaged feasibility residual $\|H(\bar x_N)_+ \|$ and summarize the results in Figure \ref{fig:obj_interior}(c). 
In Figure \ref{fig:obj_interior}(b), we observe that the slopes of $\log N$ against \lxd{$\log \big ( | F(\bar x_N, y^*) - F(x^*,\bar y_N) | \big )$} approximately equals $-1/2$ for both Basic-CSPD and Adp-CSPD. This matches our theoretical claims in Theorems \ref{thm:rates_basic} and \ref{thm:rates_adp} that the objective optimality gaps converge to zero at the rate of $\cO(\frac{1}{\sqrt{N}})$. Meanwhile, \sg{Figure~\ref{fig:obj_interior}(c)} suggests that the feasibility residual $\|H(\bar x_N)_+ \|$ decreases to zero in around \sg{$3\times 10^4$} iterations for both  Basic-CSPD and Adp-CSPD. This implies that both algorithms can  find strictly feasible solutions efficiently if the optimum falls in the interior. 
\\
\noindent(ii) \emph{Optimal Solution at the Boundary.} 
In this case, we report the empirical averaged objective gap $F(\bar x_N, y^*) - F(x^*,\bar y_N)$ and $\log N $ against  \lxd{$\log \big ( | F(\bar x_N, y^*) - F(x^*,\bar y_N) | \big )$} in Figure \ref{fig:obj_boundary}, and summarize the averaged feasibility residual $ \| H(\bar x_N)_+\|$ and $\log N$ against the log-residual  $ \log \big ( \| H(\bar x_N)_+\| \big )$ in Figure \ref{fig:infeasibility_boundary}, with additional lines of slope $-1/2$ provided as theoretical benchmarks for log-convergence. 

In Figures \ref{fig:obj_boundary} and \ref{fig:infeasibility_boundary}, we observe that the slopes of 
$\log N$ against   \lxd{$\log \big ( | F(\bar x_N, y^*) - F(x^*,\bar y_N) | \big )$} and $\log N $ against $ \log \big ( \| H(\bar x_N)_+\|_2 \big ) $ are close to $-1/2$ for both algorithms. Again, this matches our theoretical claims in Theorems \ref{thm:rates_basic} and \ref{thm:rates_adp} that both objective gap and feasibility residuals converge to zero at the rate of $\cO(1/{\sqrt{N}})$ for both  algorithms. These empirical studies suggest that our algorithms demonstrate efficient performances in solving quadratic saddle point problems with quadratic constraints, and these empirical convergence behaviors further support our theoretical rate claims.

\subsection{Robust Optimal Pricing}
\label{subsec:rop}
In this subsection, we consider the robust optimal pricing problem \eqref{prob:robust_pricing} discussed in Section \ref{sec:applications}.
We consider a linear demand function of form  $D(s,p,\xi ;\theta) = s^\top \theta_{1:d}  + \theta_0 p + \xi$ where $\xi$ is a random noise term (see Section 8.6.1 of \cite{gallego2019revenue}), and consider the scenario where the uncertainty set $\Theta$ is given by 
$$
\Theta = \Big \{  \theta \in \RR^{d+1}: \ l_i \leq \theta_i \leq u_i, \  i = 0,1,2,\cdots, d \Big \}.
$$
In this scenario, by further assuming $\theta_0 <0$, the objective function $F(\theta,p):=  \EE_{\xi} [ p D(s,p ,\xi ;\theta)]$ in \eqref{prob:robust_pricing} is convex in the parameter $\theta$ and concave in the price $p$.
In our experiments, we set the number of expectation constraints $m = 5000$. All features $\tilde s_i$ and prices $\tilde p_i$, $i=1,\ldots,m$ are i.i.d. sampled from the uniform distributions $\text{Unif}[0,3]^d$ and $\text{Unif}[10,20]$, respectively. For  test purposes, we construct the  uncertainty set $\Theta$ by the following steps: (1) set $l_0 = -5$ and sample each $l_i$ from the uniform distribution $\text{Unif}(0,2)$; (2) sample each $\omega_i$ from again the uniform distribution $\text{Unif}[0,3]$ and set $u_i = l_i + \omega_i$; (3) for $i=1,\ldots,m$, the historical lower bound $d_i$ is set to be $d_i = \tilde s_i^\top \tilde \theta_{1:d} + \tilde \theta_0 \tilde p_i + u_i$ where $u_i \sim \text{Unif}[0,5]$ and $\tilde \theta$ is sampled from $\Theta$.

\begin{figure}[t]
\vskip 0.5cm
    \centering
    \includegraphics[width=0.4\linewidth]{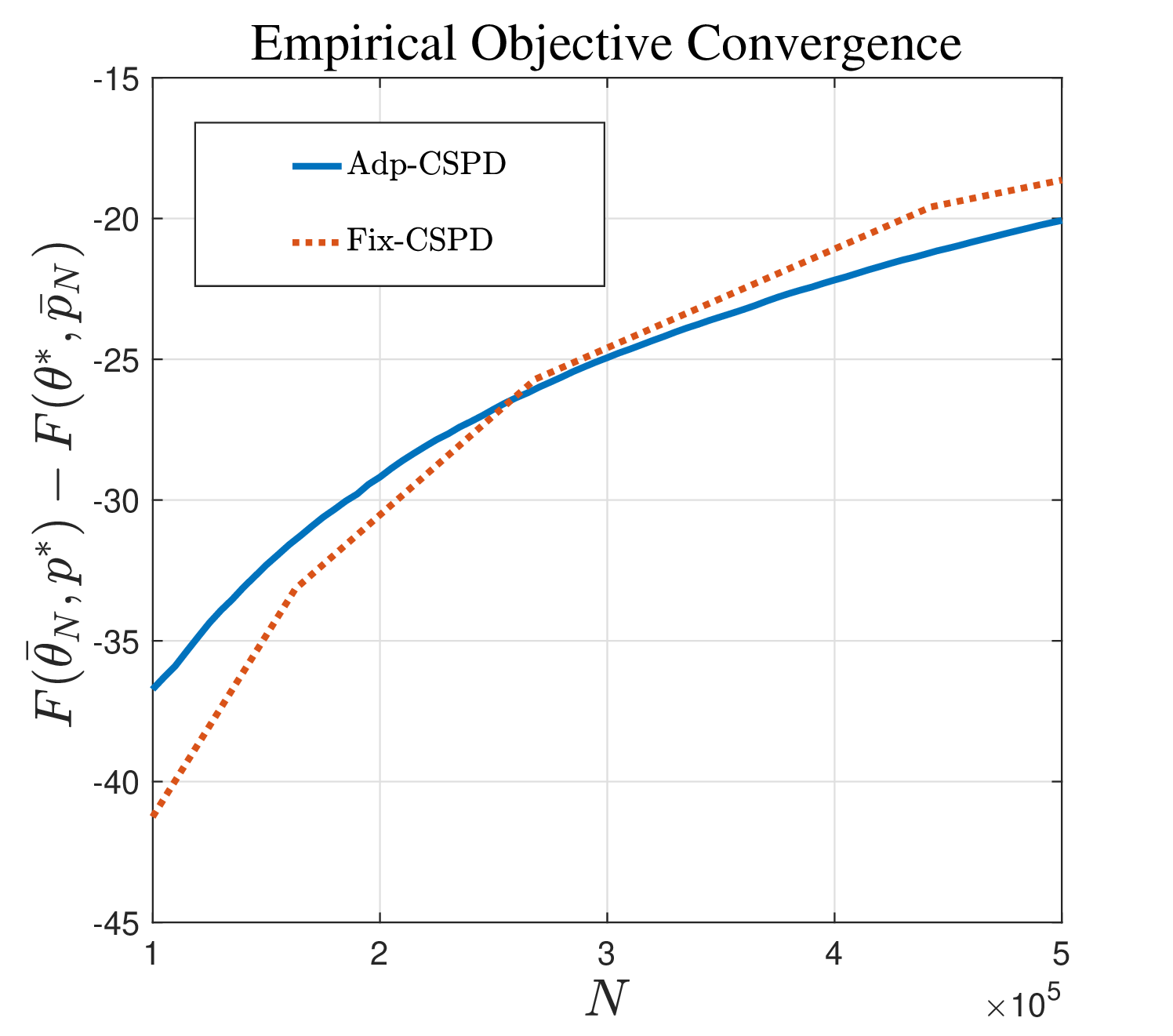}
    \includegraphics[width=0.4\linewidth]{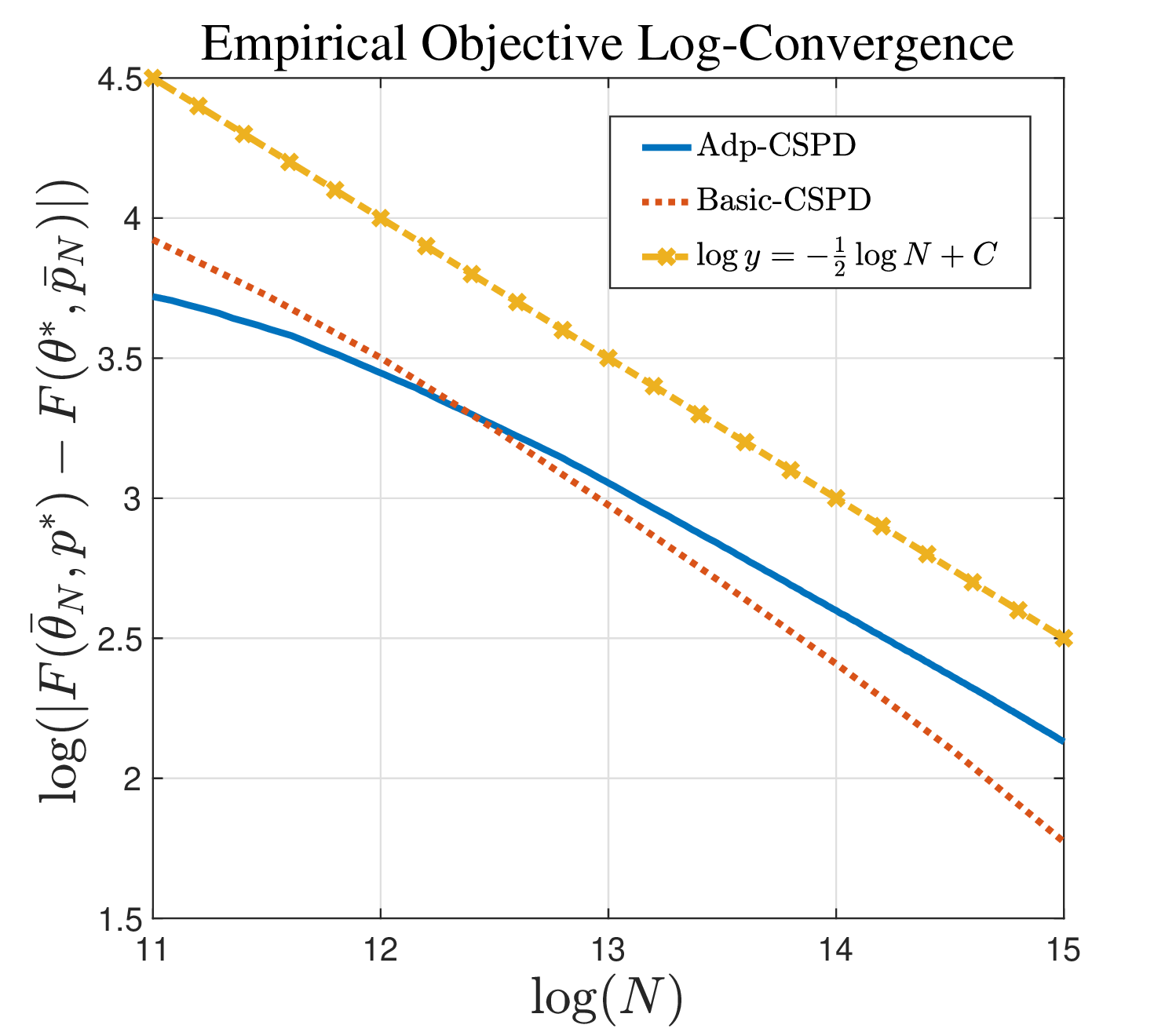}
    \caption{Empirical convergence rate of the objective gap  $ F(\bar \theta_N, p^*) - F(\theta^*,\bar p_N) $ for robust optimal pricing. }
    \label{fig:pricing_obj}
\end{figure}

\begin{figure}[t]
    \centering
    \includegraphics[width=0.4\linewidth]{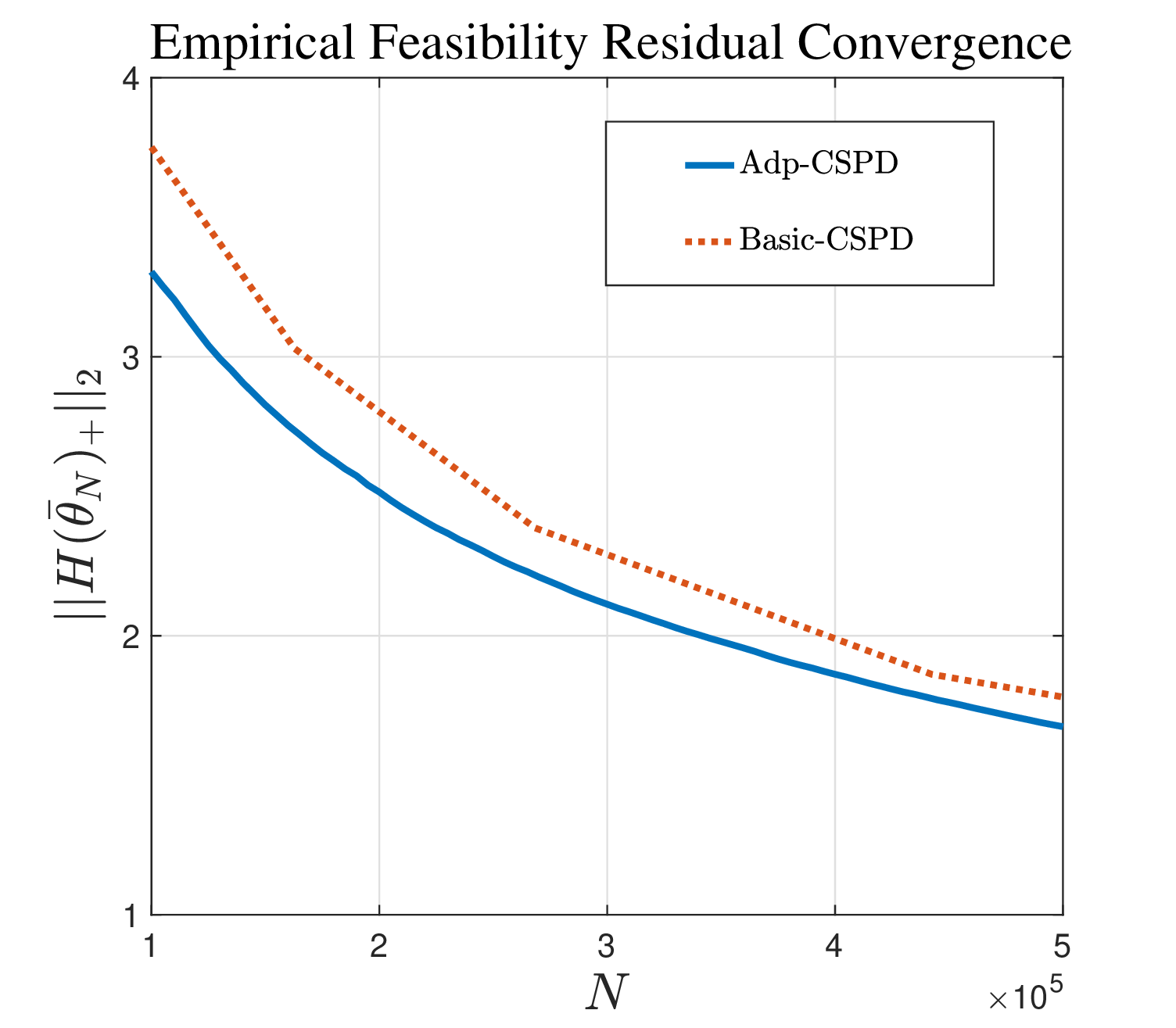}
    \includegraphics[width=0.4\linewidth]{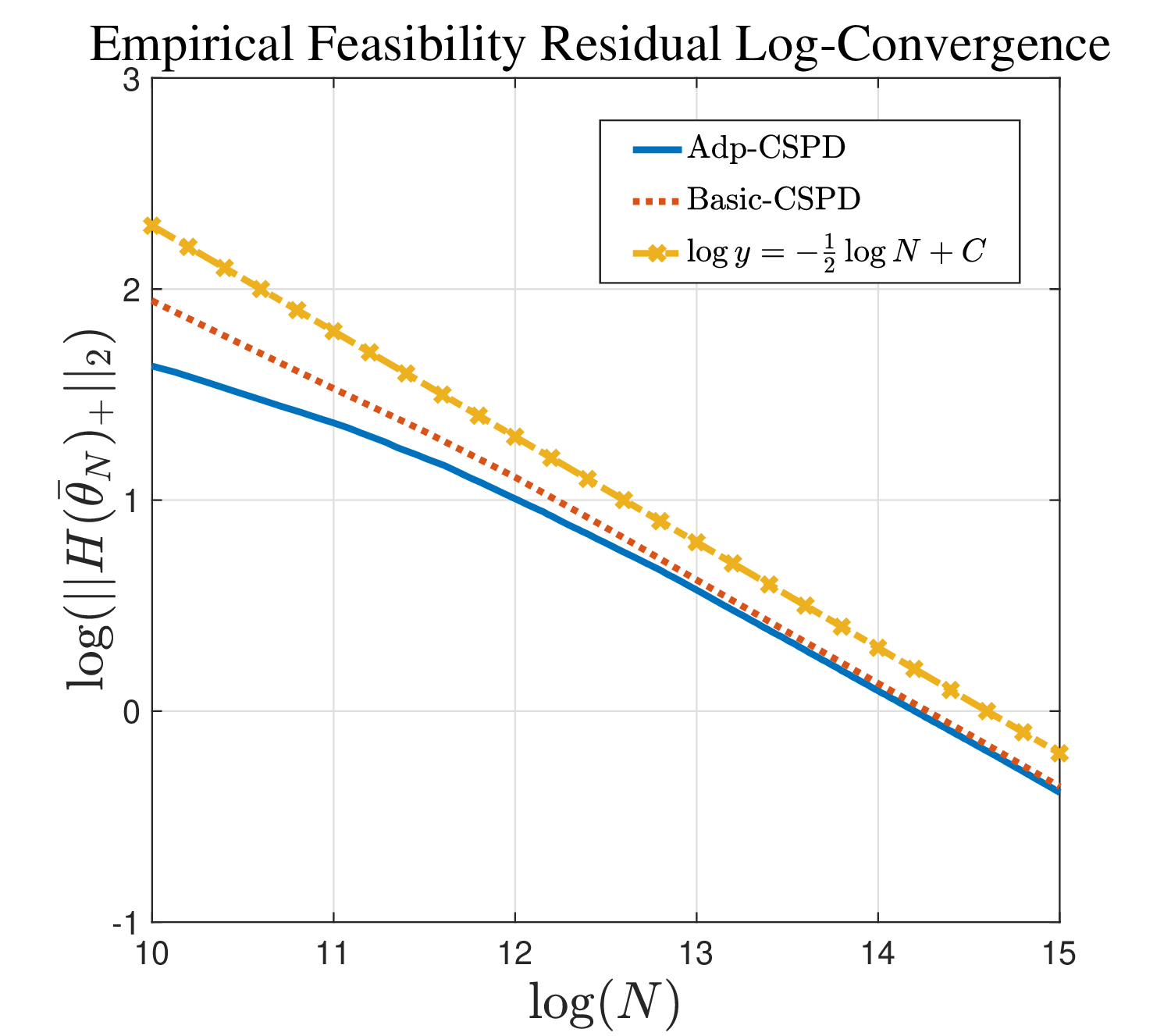}
    \caption{Empirical convergence rate of the feasibility residual $\| H(\bar \theta_N)_+ \|_2$ for robust optimal pricing. }
    \label{fig:pricing_feasibility}
\end{figure}

We test Basic-CSPD and Adp-CSPD algorithms for 
dimension $d = 100$ by 10 simulations.
In  each simulation, we run Basic-CSPD with different prefixed total numbers of iterations
$N \in \{ 2\times 10^4,5\times 10^4, 1.5\times 10^5, 5\times 10^5, 1\times 10^6, 5\times 10^6 \}$
with  step-sizes  $\beta_t = 100\sqrt{N}$ and $\eta_t = \kappa_t = 10 \sqrt{N}$ for $t =0,1,\ldots, N$; we run Adp-CSPD for $N_{\text{total}} = 5\times 10^6$ iterations with step-sizes  $\beta_t = 100\sqrt{t+1}$, $\tau_t = 100 (\sqrt{t+2} - \sqrt{t+1})$, $\eta_t = \kappa_t = 10 \sqrt{t+2}$, and $\rho_t = \psi_t = 10(\sqrt{t+3} - \sqrt{t+2})$ for $t =0,1,\ldots, N_{\text{total}}$. 

{Let $(\bar \theta_N,\bar p_N)$ be the returned pair by running our algorithms over $N$ iterations.}
We report the empirical averaged objective optimality gap $F(\bar \theta_N, p^*) - F(\theta^*,\bar p_N)$ and $\log N $ against $\log \big ( | F(\bar \theta_N, p^*) - F(\theta^*,\bar p_N) | \big )$ in Figure \ref{fig:pricing_obj}, and summarize the averaged feasibility residual $ \| H(\bar \theta_N)_+\|$ and $\log N$ against the log-residual  $ \log \big ( \| H(\bar \theta_N)_+\| \big )$ in Figure \ref{fig:pricing_feasibility}, with additional lines of slope $-1/2$ provided as theoretical benchmarks for log-convergence.  As can be seen in Figures \ref{fig:pricing_obj} and \ref{fig:pricing_feasibility}, the slopes of 
$\log N$ against  $\log \big (  | F(\bar \theta_N, p^*) - F(\theta^*,\bar p_N) | \big )$ and $\log N $ against $ \log \big ( \| H(\bar \theta_N)_+\|_2 \big ) $ are close to $-1/2$ for both algorithms. This again verifies our convergence results obtained in Theorems \ref{thm:rates_basic} and \ref{thm:rates_adp}. Moreover, 
our numerical results here indicate that our algorithms can handle minimax problems with a large number of constraints ($m=5000$), which are computationally intractable if the relaxed problem \eqref{prob:minmaxpenaltied} is solved instead with hyper-parameter optimization conducted. \sg{Finally, to further test the performance of our algorithm, we conduct additional numerical experiments where the features $\tilde s_i$ are generated under normal and student distributions, and provide the detailed numerical results in Appendix Section~\ref{app:numerics}.}

\subsection{AUC Maximization with Fairness Constraints} \label{sec:AUC_fairness}
\begin{figure}[t]
\begin{minipage}{0.32\textwidth}
    \centering
        \includegraphics[width=1\linewidth]{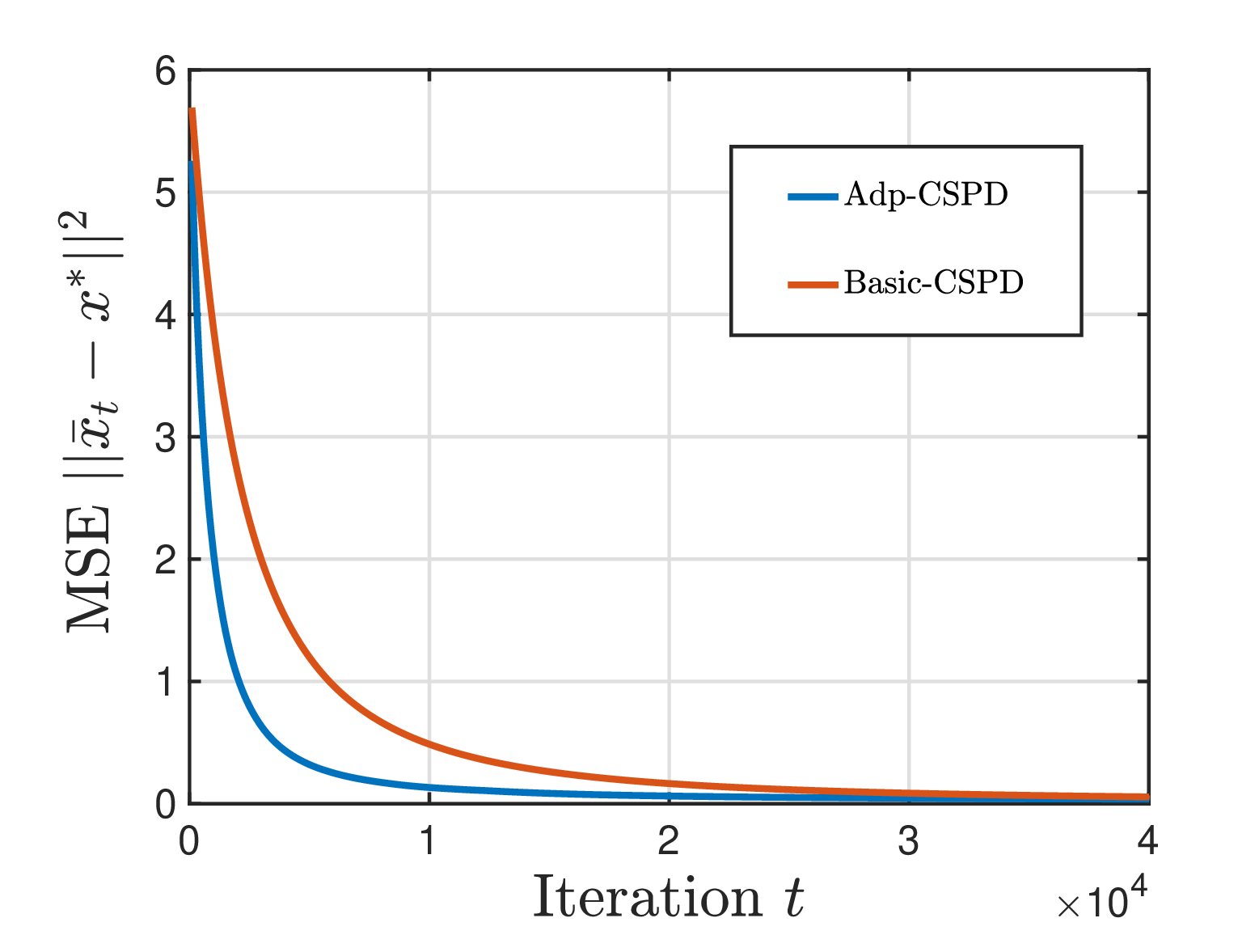}
        (a)
        \end{minipage}  
\begin{minipage}{0.32\textwidth}
    \centering
        \includegraphics[width=1\linewidth]{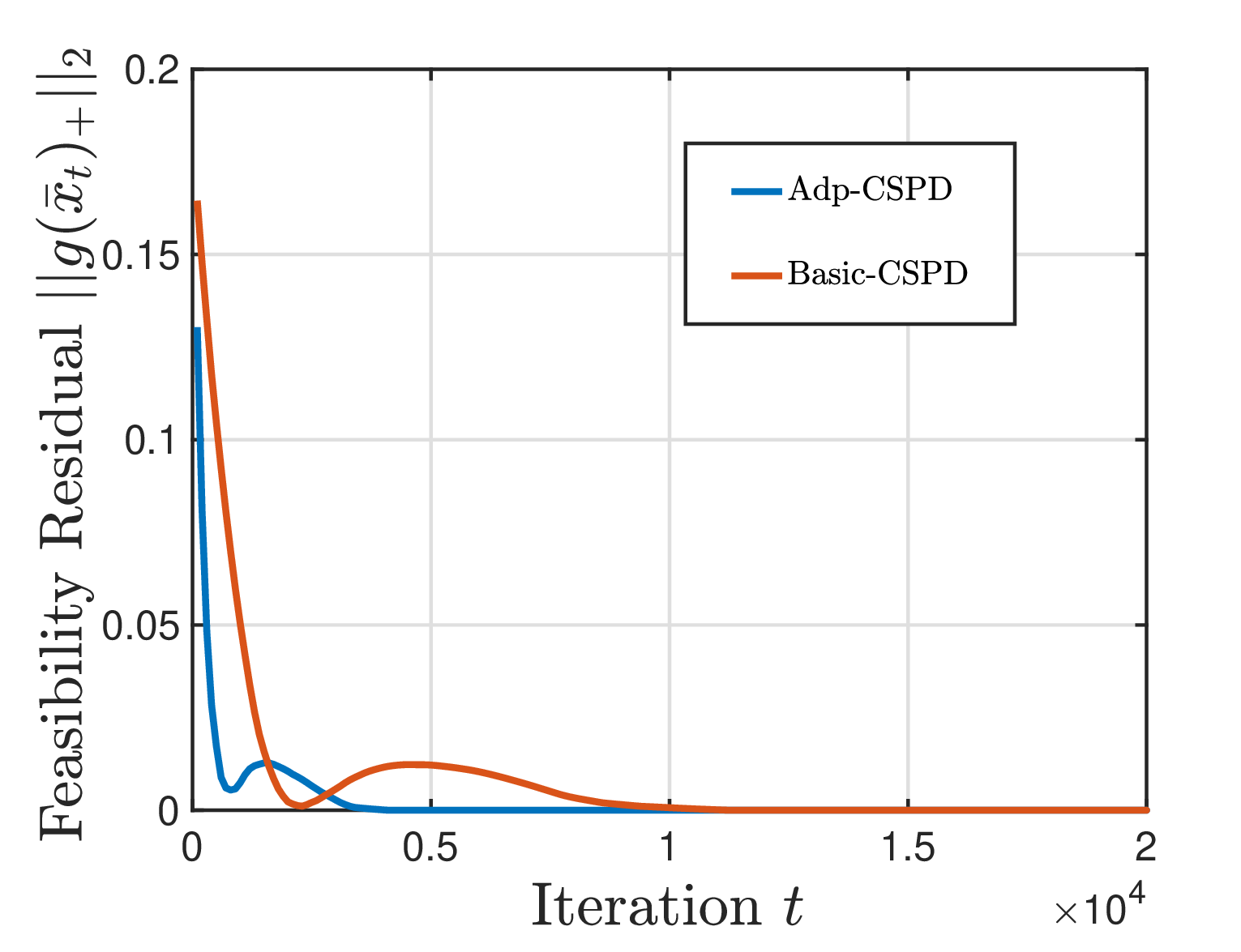}
        (b)
        \end{minipage}  
        \begin{minipage}{0.32\textwidth}
    \centering
        \includegraphics[width=1\linewidth]{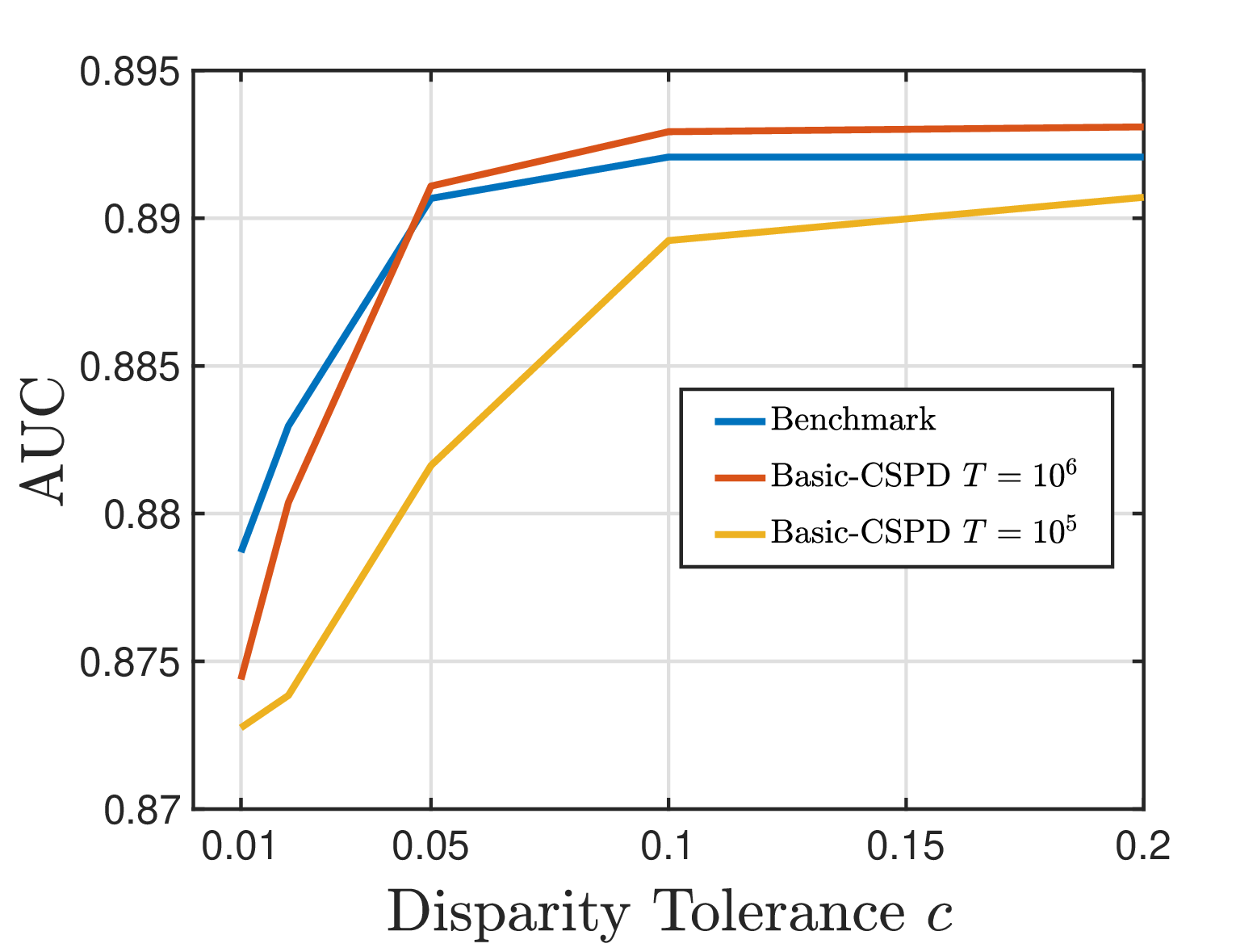}
        (c)
        \end{minipage}  
    \caption{ (a) and (b) Convergence of $\| \bar x_t - x^*\|$ and feasibility residual $\| g(\bar x_t)_+\|_2$ against iteration $t$ with $c = 0.02$. 
    (c) AUC against the disparity tolerance $c \in \{ 0.01,0.02,0.05,0.1,0.2\}$ with total iterations $T \in \{ 10^5, 10^6\}$
    }
    \label{fig:fairness}
\end{figure}

\sg{We consider an AUC maximization problem with the linear classifier over the Adult income dataset~\citep{adultincome1996}. The dataset consists  of $N = 48,842$ data points $\cD= \{ (w_i,y_i) \}_{i=1}^N $, where each $w_i \in \RR^{87}$ is the feature (after encoding categorical variables) containing the descriptive information of an adult, such as ``gender'', ``age'', and $y_i \in \{-1,1\}$ is a label indicates whether the household income is above 50K or not. The dataset $\cD$ can be further splitted as $\cD^+$ and $\cD^-$ based on whether $y_i = -1$ or $y_i = 1$, with $|\cD^+| = 37,155$ and $|\cD^-| = 11,687$. 
In our experiment, we treat ``gender'' as the sensitive variable $u\in \{0,1\}$, and consider problem~\eqref{eq:AUC_fairness} for  maximizing the convex surrogate of the AUC score while reducing the disparate impact. {In the model, the empirical average $\bar u = 0.3315$ is obtained via screening the dataset.}
We conduct the following two sets of experiments:
\begin{enumerate}
\item[(i)]
Consider problem~\eqref{eq:AUC_fairness} with a fixed disparate tolerance level $c = 0.02$. The optimal solution $x^*$ is obtained by solving the offline problem using the full batch data. We conduct $10$ independent simulations to test our Basic-CSPD and Adp-CSPD algorithms, where step-sizes are set to be $\eta_t = \beta_t = \kappa_t= 10\sqrt{T}$ for all $t\leq T$ for Basic-CSPD, and $\eta_t = \beta_t = \kappa_t= 10\sqrt{t}$ and $\rho_t= \tau_t = \phi_t = 10(\sqrt{t+1} - \sqrt{t})$ for Adp-CSPD. We report the averaged MSE $\| \bar x_t - x^*\|^2$ against iterations $t$ in Figure~\ref{fig:fairness} (a) and the averaged feasibility residual $\| g(\bar x_t)_+\|_2$ in Figure~\ref{fig:fairness} (b). 
\item[(ii)] Consider problem~\eqref{eq:AUC_fairness} with various disparate tolerance levels $c \in \{ 0.01,0.02,0.05,0.1, 0.2\}$. For each choice of $c$, we solve the corresponding online optimization problem by using the Basic-CSPD algorithm and compute the AUC score evaluated at $\bar x_T$ for $T =  10^5$ and $T = 10^6$. We report the AUC scores obtained for each $c$ against the disparate tolerance level $c$ in Figure~\ref{fig:fairness} (c). For the benchmark comparison, for each tolerance level $c$, we also solve the offline batch problem  
and compute the corresponding AUC score, and report the results in Figure~\ref{fig:fairness} (c). 
\end{enumerate}
From Figures~\ref{fig:fairness} (a) and (b), we can see that both our Basic-CSPD and Adp-CSPD algorithms converge in terms of the averaged MSE and the feasibility residual in $4\times 10^4$ iterations.  Further, Figure~\ref{fig:fairness} (c) illustrates the trade-off between AUC score and disparate tolerance level. Specifically, the benchmark plot suggests that with more restrictive disparate tolerance $c$, the corresponding optimal AUC score would decrease. There is a small gap between the plot generated by Basic-CSPD with $T = 10^5$, while if we increase the total number of iterations to $T = 10^6$, the corresponding plot is quite close to that of the benchmark. 
These results demonstrate the encouraging potential of our algorithms for solving real-world large-scale complicated optimization problems.
}

\section{Conclusion}
To cope with the emerging growth of data-driven applications, in this paper, we propose an expectation constrained minimax optimization model. Efficient primal-dual type  methods are developed to handle non-smooth convex-concave stochastic problems. Comprehensive convergence analysis reveals that the proposed algorithms achieve an optimal iteration complexity of $\cO(1/\sqrt{N})$. We also conduct some preliminary numerical experiments to verify the proven theoretical results as well as the efficiency and robustness of the proposed algorithms. 

\section*{Acknowledgement}
This research was partially supported by the National
Key R\&D Program of China, 2020YFA0711900, 2020YFA0711901; and the National Science Foundation, DMS-1953199.

\newpage
\appendix
\section*{Appendix}

\section{Proof of Result in Section \ref{sec:preliminary}} \label{app:lemmas}

\subsection{Proof of Lemma \ref{lemma:obj_lower_bound}}\label{sec:proof_of_lemma_obj_lower_bound}
{We first provide the following technical result to help derive Lemma \ref{lemma:obj_lower_bound}.}
\begin{lemma}\label{lemma:low_bounds}
Let $(x^*,\gamma^*) \in \RR^n\times\RR^m_+$ be a saddle point to the following convex optimization problem
\begin{equation}
\label{prob:fg}
\min_{x\in\cX} \left\{
f(x) \mid g(x) \le 0
\right\},
\end{equation}
where $f$ and $g_i$, $i=1,\ldots,m$ are closed proper convex functions and $\cX \subseteq \RR^n$ is a nonempty closed convex set.
Then, it holds for all $x\in \cX$ that
\[
f(x) - f(x^*) \ge -\norm{\gamma^*} \norm{g(x)_+}.
\]
\end{lemma}
\textit{Proof:}
The arguments here essentially follow from the proof to 
\cite[Corollary 2]{lan2013iteration}. 
Consider the following Lagrangian function associated with \eqref{prob:fg}:
\[
\cL(x;\gamma) = f(x) + \gamma^\top g(x), \quad \forall \, (x,\gamma)\in \cX \times \RR^m_+.
\]
Since $(x^*,\gamma^*)$ is a saddle point, we have that
\[
\cL(x^*;\gamma) \le \cL(x^*; \gamma^*)  = f(x^*) \le \cL(x; \gamma^*), \quad \forall \, (x,\gamma) \in \cX\times \RR^m_+.
\]
For any given $x\in\cX$, the above inequality implies that 
\[
f(x) = \cL(x;\gamma^*) - g(x)^\top {\gamma^*} \ge \cL(x^*;\gamma^*) -  g(x)^\top {\gamma^*} = f(x^*) -  g(x)^\top {\gamma^*}.
\]
Since $ (g(x)_+ - g(x))^\top \gamma^* \ge 0$, we further have that
\[
f(x) \ge f(x^*) - (g(x)_+)^\top \gamma^* \ge f(x^*) - \|\gamma^*\| \|g(x)_+\|. 
\]
This completes the proof.
\QED

{Now we are ready to prove Lemma \ref{lemma:obj_lower_bound}.}

\textit{Proof of Lemma \ref{lemma:obj_lower_bound}:}
Recall that $(x^*,y^*,\gamma^*,\lambda^*)$ is a saddle point to problem \eqref{prob:saddle_point}. Then, by using \eqref{eq:min_max}, we observe that $(x^*, \gamma^*)$ is saddle point to the following convex optimization problem
\[
\min_{x\in \cX} \left\{ 
F(x,y^*) \mid H(x) \le 0
\right\}.
\]
Hence, by using Lemma \ref{lemma:low_bounds}, we have 
\[
F(x,y^*) - F(x^*,y^*) \ge -\norm{\gamma^*}\norm{H(x)_+}, \quad \forall x\in\cX.
\]
Similarly, it holds that
\[
-F(x^*,y) + F(x^*,y^*) \ge -\norm{\lambda^*}\norm{G(y)_+}, \quad \forall y\in\cY.
\]
 The desired result is obtained by combining the  two inequalities above.
\QED

\section{Proof of Results in Section \ref{sec:basic}} \label{sec:appA}
We first present a three-point lemma, which is frequently used throughout our analysis.
\begin{lemma}[Lemma 3.8 of \cite{lan2020first}]\label{lemma:three_point}
Let \lxd{$Y$ be a given closed convex set} and $V$ be some Bregman distance, and assume function $\phi$ is $\mu$-strongly convex such that 
$\phi(y) - \phi(\bar y ) -\phi'(\bar y)^\top (y - \bar y ) \geq \mu V(\bar y,y)$ for all $y,y' \in Y$. For given $\pi$, if $\hat y \in \argmin_{y \in Y} \{ \pi^\top y + \phi(y) + \tau V(\bar y, y)\}$, then 
\begin{equation*}
(\hat y - y)^\top \pi + \phi( \hat y) -\phi(y) \leq \tau V(\bar y, y) -(\tau + \mu) V(\hat y,y) -\tau V(\bar y,\hat y), \quad \forall y\in Y.
\end{equation*}
\end{lemma}

\subsection{Lemma \ref{lemma:sum_martingale} and Its Proof} \label{sec:proof_of_lemma_sum_martingale}
We provide the following result to resolve the complex structure and facilitate our analysis.
	\begin{lemma} \label{lemma:sum_martingale}
\sg{Let $\{ \delta_t \}_{t=0}^{N-1}$ be a sequence of random variables. Suppose $\delta_t$ is conditionally mean-zero such that $\EE[\delta_t \mid  \delta_{0},\delta_1,\cdots, \delta_{t-1} ] = 0$ for all $t\geq 1$.} 
For any $\pi$ such that $\EE[\|\pi\|^2] < + \infty$, it holds that: 
		\begin{itemize}
			\item [(a)] Suppose 
			$\EE\Big[ \| \delta_t \|^2 \mid  \delta_{0},\delta_1,\cdots, \delta_{t-1} \Big] \leq \sigma^2$, then 
			$$
			\EE[\sum_{t=0}^{N-1}    \pi^{\top} \delta_t   ] \leq \sqrt{N}\EE[\| \pi\|]\sigma.
			$$
			\item[(b)] Let $\tau >0$ be a positive number, then 
			$$
			\EE \Big [ \sum_{t=0}^{N-1}   \pi^{\top} \delta_t  \Big ] \leq \frac{\tau}{2} \EE[ \| \pi \|^2 ] + \sum_{t=0}^{N-1} \frac{1}{2\tau} \EE [ \|  \delta_t \|^2] .
			$$
			\item[(c)] Let $\{ \tau_t\}, \{\rho_t \}$ be two positive sequences such that $\tau_t + \rho_t \geq \tau_{t+1}$, then 
			\begin{equation*}
			\begin{split}
			\EE \Big[ \sum_{t=0}^{N-1}  \pi^{\top} \delta_t \Big ]  
			\leq \Big ( \frac{\tau_0}{2}+ \sum_{t=0}^{N-1} \frac{\rho_t}{2} \Big ) \EE[ \| \pi \|^2 ] +  \sum_{t=0}^{N-1} \frac{1}{2\tau_t} \EE [ \|  \delta_t \|^2].
			\end{split}
			\end{equation*}
		\end{itemize}
	\end{lemma}
\textit{Proof:}
	Part (a) comes from Lemma 2 of \cite{zhang2020optimal}, part (b) can be derived by similar ideas with  $\tau >0$ been kept as a generic value, and part (c) is a generalization of part (b). Here we present the proof of part (c), and part b) can be derived from part c) by setting $\tau_t = \tau$ and $\rho_t = 0$.
	\\
	\noindent Part (c): We define an  auxiliary sequence $\{ \pi_t\}$ by 
	\begin{equation*}
        \sg{
	\pi_t = 
	\begin{cases}
	0 & \text{ if } t = 0 , \\
	\argmin_{\tilde \pi } -  \delta_{t-1}^\top \tilde \pi    + \frac{\tau_{t-1}}{2} \|  \pi_{t-1} - \tilde \pi  \|^2 + \frac{\rho_{t-1}}{2} \| \pi_0 - \tilde \pi  \|^2 & \text{ if }t \geq 1.
	\end{cases}
        }
	\end{equation*}
	We can see that $ \pi_{t}$ is conditionally independent of $\delta_t$. By using the three-point lemma \ref{lemma:three_point}, we have for all $t \ge 0$ that \sg{for any $\pi$,}
	$$
	- (\pi_{t+1} - \pi)^\top \delta_t  \leq \frac{\tau_t}{2} \| \pi_t -  \pi \|^2 - \frac{\tau_t}{2} \| \pi_{t+1} - \pi_t\|^2 - \frac{\tau_t + \rho_t}{ 2} \|  \pi_{t+1} - \pi\|^2 - \frac{\rho_t}{2}  \| \pi_{t+1} -  \pi_0 \|^2 + \frac{\rho_t}{2} \|  \pi_0 - \pi \|^2,
	$$
	which further implies that 
	\begin{equation*}
	\begin{split}
	- (\pi_{t} - \pi)^\top \delta_t  \leq {}& \frac{\tau_t}{2} \| \pi_t -  \pi \|^2 - \frac{\tau_t}{2} \|  \pi_{t+1} 
	-  \pi_t\|^2 - \frac{\tau_t + \rho_t}{ 2} \|  \pi_{t+1} - \pi\|^2 \\
	& 
	- \frac{\rho_t}{2}  \| \pi_{t+1} -  \pi_0 \|^2 + \frac{\rho_t}{2} \|  \pi_0 - \pi \|^2 + (\pi_{t+1} - \pi_{t})^\top \delta_t  
	\\
	\leq {}& \frac{\tau_t}{2} \| \pi_t -  \pi \|^2 - \frac{\tau_t + \rho_t}{ 2} \|  \pi_{t+1} - \pi\|^2 
	- \frac{\rho_t}{2}  \| \pi_{t+1} -  \pi_0 \|^2 + \frac{\rho_t}{2} \|  \pi_0 - \pi \|^2  +  \frac{1}{2\tau_t}\|  \delta_t \|^2,
	\end{split}
	\end{equation*}
	where the last inequality holds by the fact that $ (\pi_{t+1} - \pi_{t})^\top \delta_t  \leq \frac{1}{2\tau_t} \|  \delta_t \|^2 + \frac{\tau_t}{2} \| \pi_{t+1} - \pi_{t}\|^2$. 
By summing the above inequality over $t=0,1,\cdots, N-1$ and noting that $\pi_0 = 0$, $\tau_t + \rho_t \ge \tau_{t+1}$, we arrive at 
\begin{equation*}
	\begin{split}
\sum_{t=0}^{N-1}  \delta_t^\top \pi  & \le \sum_{t=0}^N \delta_t^\top \pi_t  + \Big ( \frac{\tau_0}{2}+ \sum_{t=0}^{N-1} \frac{\rho_t}{2} \Big )  \|\pi\|^2 
	+  \sum_{t=0}^{N-1} \frac{1}{2\tau_t} \|  \delta_t \|^2 + \sum_{t=0}^{N-1} \frac{\tau_{t+1} - \tau_t - \rho_t}{ 2} \|  \pi_{t+1} - \pi\|^2  \\
& \le \sum_{t=0}^N \delta_t^\top \pi_t  + \Big ( \frac{\tau_0}{2}+ \sum_{t=0}^{N-1} \frac{\rho_t}{2} \Big )  \|\pi\|^2 
	+  \sum_{t=0}^{N-1} \frac{1}{2\tau_t} \|  \delta_t \|^2
	\end{split}
	\end{equation*}
Since $\EE[  \pi_t^\top \delta_t  \mid  \delta_{0},\delta_1,\cdots, \delta_{t-1} ] = 0$, we take expectations on both sides and conclude that 
	\begin{equation*}
	\begin{split}
	\EE \Big [\sum_{t=0}^{N-1}   \pi^\top \delta_t  \Big ] & 
	\leq \Big ( \frac{\tau_0}{2}+ \sum_{t=0}^{N-1} \frac{\rho_t}{2} \Big )\EE[  \| \pi \|^2]+  \sum_{t=0}^{N-1} \frac{1}{2\tau_t} \EE [ \|  \delta_t \|^2].
	\end{split}
	\end{equation*}
	This completes the proof.  
	\QED

\subsection{Proof of Lemma~\ref{lemma:1}}\label{sec:diffL}
\textit{Proof:}
For any $(x,y,\lambda,\gamma)\in \cX \times \cY \times\RR_{m_1}\times\RR_{m_2}$, we split the gap function $Q(z_{t+1}, z)$ in the following manner: 
\begin{equation}\label{eq:thm1_split}
\begin{split}
&\cL(x_{t+1}, y,\gamma,\lambda_{t+1}) -  \cL(x, y_{t+1}, \gamma_{t+1},\lambda)   \\
 = {}&  F(x_{t+1},y) + \gamma^\top H(x_{t+1}) - \lambda_{t+1}^\top G(y)  -  F(x, y_{t+1}) - \gamma_{t+1}^\top H(x) + \lambda^\top G(y_{t+1})  \\
 =  {}& F(x_{t+1},y) - F(x_t, y) + F(x_t,y) - F(x_t,y_t) + F(x_t,y_t)  - F(x,y_t) + F(x,y_t) -  F(x, y_{t+1}) \\
& + \gamma^\top H(x_{t+1}) - \lambda_{t+1}^\top G(y) - \gamma_{t+1}^\top H(x) + \lambda^\top G(y^{t+1})  
\\
\leq {}&   C_f \| x_{t+1} - x_t\| + C_f \| y_{t+1} - y_t\|  +  Q_x^{t+1} + Q_y^{t+1} \\
 \leq {}&  \frac{3C_f^2 }{2\eta_t} + \frac{\eta_t}{6}\| x_{t+1} - x_t\|^2 + \frac{3C_f^2 }{2\kappa_t} + \frac{\kappa_t}{6} \| y_{t+1} - y_t\|^2 +  Q_x^{t+1} + Q_y^{t+1},
\end{split}
\end{equation}
where 
\begin{equation*}
\begin{split}
Q_x^{t+1} & : =   F(x_t,y_t)  - F(x,y_t)  + \gamma^\top  H(x_{t+1})  - \gamma_{t+1}^\top  H(x),  \\
Q_y^{t+1} & : =  F(x_t,y) - F(x_t,y_t) - \lambda_{t+1}^\top  G(y)  + \lambda^\top  G(y_{t+1}) , 
\end{split}
\end{equation*}
and the first inequality holds since $F$ is Lipschitz continuous under  Assumption \ref{assumption:01}. Since $ F(\bullet,y_t) + \gamma_{t+1}^\top  H(\bullet)$ is a convex function, we have 
\begin{equation*}\label{eq:def_qx}
\begin{split}
Q_x^{t+1}&  =  F(x_t,y_t)  - F(x,y_t)  +    \gamma_{t+1}^\top \big(H(x_t) -  H(x)\big) -  \gamma_{t+1}^\top  H(x_t) + \gamma^\top  H(x_{t+1})  \\
& \leq  \big( \widetilde \nabla_x F(x_t, y_t) +    \sum_{i=1}^{m_1}   \gamma_{t+1,i} \widetilde \nabla H_i(x_t) \big)^\top (x_t - x) + \gamma^\top  H(x_{t+1})  -  \gamma_{t+1}^\top  H(x_t).
\end{split}
\end{equation*}
Similarly, by using the concavity of $F(x_t,\bullet) - \lambda_{t+1}^\top G(\bullet)$, we have 
\begin{align*}
    Q_y^{t+1} & = F(x_t,y) - F(x_t,y_t) - \lambda_{t+1}^\top \big(G(y) - G(y_t)\big)
    + \lambda^\top  G(y_{t+1})  -  \lambda_{t+1}^\top  G(y_t)   
    \\
    & \leq \big( \widetilde \nabla_y F(x_t,y_t) -  \sum_{j=1}^{m_2} \lambda_{t+1,j} \widetilde \nabla G_j(y_t) \big)^\top(y - y_t)   + \lambda^\top  G(y_{t+1})  -  \lambda_{t+1}^\top  G(y_t)   .
\end{align*}
By substituting the above inequalities into \eqref{eq:thm1_split}, and conducting a similar analysis for $Q_y^{t+1}$, we conclude 
\begin{equation*}
\begin{split}
&\cL(x_{t+1}, y,\gamma,\lambda_{t+1}) -  \cL(x, y_{t+1}, \gamma_{t+1},\lambda)   \\
\leq {}& \big(\widetilde \nabla_x F(x_t, y_t) +  \sum_{i=1}^{m_1} \gamma_{t+1,i} \widetilde \nabla H_i(x_t) \big)^\top (x_t - x)  + \gamma^\top  H(x_{t+1})  -  \gamma_{t+1}^\top  H(x_t)   \\
& + \big( \widetilde \nabla_y F(x_t,y_t) -  \sum_{j=1}^{m_2} \lambda_{t+1,j} \widetilde \nabla G_j(y_t) \big)^\top(y - y_t)  + \lambda^\top  G(y_{t+1})  -  \lambda_{t+1}^\top  G(y_t) 
\\
& +  \frac{3C_f^2 }{2\eta_t} + \frac{\eta_t}{6}\| x_{t+1} - x_t\|^2 + \frac{3C_f^2 }{2\kappa_t} + \frac{\kappa_t}{6} \| y_{t+1} - y_t\|^2.
\end{split}
\end{equation*}
This completes the proof. 
\QED

\subsection{Proof of Lemma \ref{lem:deltaxgamma}}\label{sec:proof_of_lemma_delta_x}
\textit{Proof:} Recall the definitions of $\cH$ and $\Delta_x^{t+1}$ in \eqref{eq:H5}, \eqref{eq:H3}, and \eqref{def:Deltas}.
It holds that 
\begin{equation}\label{eq:delta_x_split}
\begin{split}
\Delta_x^{t+1} & =   \big(\widetilde \nabla F(x_{t}, y_{t}) +   \sum_{i=1}^{m_1} \gamma_{t+1, i} \widetilde \nabla H_i(x_{t})\big)^\top (x_{t} - x)   = 
\widetilde \nabla_x \cL(x_{t},y_{t},\gamma_{t+1})^\top (x_{t} - x) \\
& =    \widetilde \nabla_x L(x_{t},y_{t},\gamma_{t+1},\omega_t^1,\xi_t^2)^\top(x_{t+1}  -x)   + \big(\widetilde \nabla_x L(x_{t},y_{t},\gamma_{t+1},\omega_t^1,\xi_t^2) - \widetilde \nabla_x
\cL(x_{t},y_{t},\gamma_{t+1}) \big)^\top ( x -x_{t}) \\
& \quad  +  \widetilde \nabla_x L(x_{t},y_{t},\gamma_{t+1},\omega_t^1,\xi_t^2)^\top (x_{t} - x_{t+1} ). \\
\end{split}
\end{equation}
By recalling the update rule \eqref{update:basic_x} for $x_{t+1}$ and using the three-point Lemma \ref{lemma:three_point} in Appendix Section \ref{sec:appA}, for all $x \in \cX$, we have 
\begin{equation}\label{eq:delta_x2}
\begin{split}
\widetilde \nabla_x L(x_{t},y_{t},\gamma_{t+1},\omega_t^1,\xi_t^2)^\top(x_{t+1}  -x) \leq \frac{\eta_{t}}{2} \| x_{t} -x\|^2 - \frac{\eta_{t}}{2} \| x_{t} - x_{t+1}\|^2 - \frac{\eta_{t}}{2} \| x_{t+1} - x\|^2. 
\end{split}
\end{equation}
Meanwhile, it holds by simple calculations that 
\begin{equation*}
\begin{split}
\widetilde \nabla_x L(x_{t},y_{t},\gamma_{t+1},\omega_t^1,\xi_t^2)^\top (x_{t} - x_{t+1}) 
& \leq    \frac{3 \|\widetilde \nabla_x L(x_{t},y_{t},\gamma_{t+1},\omega_t^1,\xi_t^2)  \|^2}{2\eta_{t}} + \frac{ \eta_{t} \| x_{t} - x_{t+1}\|^2}{6}
\end{split}
\end{equation*}
Therefore, we have from \eqref{eq:delta_x_split} that
\begin{align}\label{eq:deltaxpxttp1}
\Delta_x^{t+1} + \frac{\eta_t}{3} \norm{x_t  - x_{t+1}}^2 \le {}& \frac{\eta_{t}}{2} \| x_{t} -x\|^2  - \frac{\eta_{t}}{2} \| x_{t+1} - x\|^2 + \big(\widetilde \nabla_x L(x_{t},y_{t},\gamma_{t+1},\omega_t^1,\xi_t^2) - \widetilde \nabla_x \cL(x_{t},y_{t},\gamma_{t+1}) \big)^\top ( x -x_{t}) \nonumber \\
{}& + \frac{3 \|\widetilde \nabla_x L(x_{t},y_{t},\gamma_{t+1},\omega_t^1,\xi_t^2)  \|^2}{2\eta_{t}}.
\end{align}
By summing \eqref{eq:deltaxpxttp1} over $t=0,1,\cdots, K-1$, we see that for any $x\in \cX$, 
\begin{equation}
    \label{eq:sumdeltax}
    \begin{aligned}
    &\sum_{t=0}^{K-1} \big( \Delta_x^{t+1} + \frac{\eta_t}{3} \norm{x_t  - x_{t+1}}^2) + \frac{\eta_{0}}{2} \| x_{K} - x\|^2\\ \le {}&\frac{\eta_{0}}{2} \| x_{0} -x\|^2 + \sum_{t=0}^{K-1} \frac{3 \|\widetilde \nabla_x L(x_{t},y_{t},\gamma_{t+1},\omega_t^1,\xi_t^2)  \|^2}{2\eta_{t}}
    \\
    {}&+ \sum_{t=0}^{K-1} \big(\widetilde \nabla_x L(x_{t},y_{t},\gamma_{t+1},\omega_t^1,\xi_t^2) - \widetilde \nabla_x \cL(x_{t},y_{t},\gamma_{t+1}) \big)^\top ( x -x_{t}).
    \end{aligned}
\end{equation}

Next, we focus on the term $\Delta_\gamma^{t+1}$ and split it as 
\begin{equation}\label{eq:delta_gamma}
\begin{split}
\Delta_\gamma^{t+1} & =  \gamma^\top H(x_{t+1})  -  \gamma_{t+1}^\top H(x_{t})   \\
& =   (  \gamma - \gamma_{t+1} )^\top H(x_{t})  + \gamma^\top [ H(x_{t+1}) - H(x_{t}) ]   \\
& =    \underbrace{ - h(x_{t},\xi_t^1  )^\top (  \gamma_{t+1} - \gamma  ) }_{ \Delta_{\gamma,1}^{t+1} }  +  \underbrace{ (  \gamma - \gamma_{t+1} )^\top [H(x_{t}) -  h(x_{t},\xi_t^1 ) ] }_{ \Delta_{\gamma,2}^{t+1} } + \underbrace{ \gamma^\top \big( H(x_{t+1}) - H(x_{t}) \big)}_{ \Delta_{\gamma,3}^{t+1} }. 
\end{split}
\end{equation}
First, consider $\Delta_{\gamma,1}^{t+1} $, recall the update rule that 
$$
\gamma_{t+1} =  \argmin_{ \gamma \in \RR_+^{m_1} } \Big \{ - h(x_{t},\xi_t^1 )^\top \gamma + \frac{\beta_{t}}{2} \| \gamma_{t} - \gamma \|^2 \Big \} .
$$
By using the three-point Lemma \ref{lemma:three_point} in Appendix Section \ref{sec:appA}, 
we have for any $\gamma \in \RR_+^{m_1}$ that 
\begin{equation}\label{eq:delta_gamma_1oE}
\begin{split}
\Delta_{\gamma,1}^{t+1}=  - h(x_{t},\xi_t)^\top (  \gamma_{t+1} - \gamma  )  \leq \frac{\beta_{t}}{2} \Big (  \| \gamma_{t} - \gamma\|^2 - \| \gamma_{t} - \gamma_{t+1}\|^2 - \| \gamma_{t+1} - \gamma\|^2 \Big ). 
\end{split}
\end{equation}
Meanwhile, it holds that
\begin{equation}\label{eq:delta_gamma_2oE}
\begin{split}
\Delta_{\gamma,2}^{t+1} & =   \big (H(x_{t}) -  h(x_{t},\xi_t^1 ) \big )^\top (\gamma - \gamma_t)    + \big (H(x_{t}) -  h(x_{t},\xi_t^1 ) \big )^\top(\gamma_{t} - \gamma_{t+1} ) \\
& \leq    \big (  H(x_{t}) -  h(x_{t},\xi_t^1 ) \big )^\top (\gamma - \gamma_t)   + \frac{ \|H(x_{t}) -  h(x_{t},\xi_t^1 )\|^2 }{2\beta_{t} }+ \frac{ \beta_{t} }{2}  \| \gamma_{t} - \gamma_{t+1} \|^2  . 
\end{split}
\end{equation}
By the Lipschitz continuity of $H$ and  some simple computations, we have 
\begin{equation}\label{eq:delta_gamma_3}
\begin{split}
  \Delta_{\gamma,3}^{t+1}   & \leq    \| \gamma\|   \|  H(x_{t+1}) - H(x_{t}) \|   \\
& \leq   C_h  \| \gamma\|   \| x_{t+1} - x_{t}\| 
\leq  \frac{3 \| \gamma\|^2 C_h^2 }{2\eta_{t}} + \frac{ \eta_{t} }{6} \| x_{t+1} - x_{t}\|^2 .
\end{split}
\end{equation}
Thus, it holds from \eqref{eq:delta_gamma}, \eqref{eq:delta_gamma_1oE}, \eqref{eq:delta_gamma_2oE} and \eqref{eq:delta_gamma_3} that for any $\gamma\in \Re_+^{m_1}$,
\begin{equation}
    \label{eq:delta_gamma_sum}
    \begin{aligned}
        \sum_{t=0}^{K-1} \Delta_{\gamma}^{t+1} + \frac{\beta_0}{2}\|\gamma_K - \gamma\|^2 \le {}& \frac{\beta_0}{2}\|\gamma_0 - \gamma\|^2 + \sum_{t=0}^{K-1} \big (  H(x_{t}) -  h(x_{t},\xi_t^1 ) \big )^\top (\gamma - \gamma_t) \\
        {} & +\sum_{t=0}^{K-1} \Big( \frac{ \|H(x_{t}) -  h(x_{t},\xi_t^1 )\|^2 }{2\beta_{t} } +   \frac{3 \| \gamma\|^2 C_h^2 }{2\eta_{t}} + \frac{ \eta_{t} }{6} \| x_{t+1} - x_{t}\|^2 \Big ).
    \end{aligned}
\end{equation}
Summing \eqref{eq:sumdeltax} and \eqref{eq:delta_gamma_2oE}, we obtain the desired inequality and complete the proof. \QED

\subsection{Proof of Lemma \ref{lem:EUV}}\label{sec:apen_UtVt}
Recall the definition of $U_t$ in \eqref{eq:Uxgamma}:
\begin{equation*}
\begin{aligned}
    U_t(x,\gamma) ={}& \big( \widetilde \nabla_x L(x_{t},y_{t},\gamma_{t+1},\omega_t^1,\xi_t^2) -  \widetilde \nabla_x \cL(x_{t},y_{t},\gamma_{t+1}) \big)^\top ( x -x_{t})  + \frac{3 \| \widetilde \nabla_x L(x_{t},y_{t},\gamma_{t+1},\omega_t^1,\xi_t^2)\|^2}{2\eta_t} \\
    {}&+\big (  H(x_{t}) -  h(x_{t},\xi_t^1 ) \big )^\top (\gamma - \gamma_t)   + \frac{ \|H(x_{t}) -  h(x_{t},\xi_t^1 )\|^2 }{2\beta_{t} }.
\end{aligned}
\end{equation*}

We first note from the independence between $ \widetilde \nabla_x L(x_{t},y_{t},\gamma_{t+1},\omega_t^1,\xi_t^2) -  \widetilde \nabla_x \cL(x_{t},y_{t},\gamma_{t+1}) $ and $x_t$ that 
\begin{equation}\label{eq:delta_x1}
	\EE \Big [ \big (  \widetilde \nabla_x L(x_{t},y_{t},\gamma_{t+1},\omega_t^1,\xi_t^2) -  \widetilde \nabla_x \cL(x_{t},y_{t},\gamma_{t+1}) \big )^\top x_{t} \Big ]= 0
\end{equation} 
Meanwhile, for any $x\in\cX$ satisfying $\EE[\|x\|^2] \le +\infty$, we know from Lemma \ref{lemma:sum_martingale} (b) that
\begin{equation}\label{eq:delta_x4}
	\begin{split}
	& \EE \Big [\sum_{t=0}^{K-1}  \big (  \widetilde \nabla_x L(x_{t},y_{t},\gamma_{t+1},\omega_t^1,\xi_t^2) -  \widetilde \nabla_x \cL(x_{t},y_{t},\gamma_{t+1}) \big )^\top x \Big ] \\
	& \leq \frac{ \eta_0 }{2}\EE[ \| x\|^2] + \sum_{t=0}^{K-1} \frac{1}{2\eta_t} \EE [ \| \widetilde \nabla_x L(x_{t},y_{t},\gamma_{t+1},\omega_t^1,\xi_t^2) -  \widetilde \nabla_x \cL(x_{t},y_{t},\gamma_{t+1}) \|^2] 
	\\
	& \leq  \frac{ \eta_0 }{2}\EE[ \| x\|^2] + \sum_{t=0}^{K-1} \frac{1 }{\eta_t}  \EE \Big [ \|  \widetilde \nabla f(x_t, y_t, \omega_t^1)   - \widetilde \nabla F(x_{t},y_{t} )  \|^2 +  \|  ( \widetilde \nabla h(x,\xi_t^2) - \widetilde \nabla H(x_t) ) \gamma_{t+1}  \|^2 \Big ] 
 \\
	&\leq \frac{ \eta_0 }{2}\EE[ \| x\|^2] + \sum_{t=0}^{K-1} \frac{1}{\eta_t} (C_f^2 + C_h^2\EE[ \| \gamma_{t+1}\|^2]),
	\end{split}
	\end{equation}
	where we use the facts that $\eta_t \equiv \eta_0$, $\| a+b\|^2 \leq 2\| a \|^2 + 2 \| b \|^2$, the independence between $\gamma_{t+1}$ and $\nabla h(x_t,\xi_t^2)$ in the update of Algorithm \ref{alg:1}, Assumption \ref{assumption:01} that $\EE[ \|  \widetilde \nabla f(x_t, y_t, \omega_t^1)   - \widetilde \nabla F(x_{t},y_{t} )  \|^2 ] \leq C_f^2$,  and Assumption \ref{assumption:02} that $\EE[ \|\widetilde \nabla h(x,\xi_t^2) - \widetilde \nabla H(x_t)    \|^2] \leq C_h^2$ in the last inequality.
Assumptions \ref{assumption:01} and \ref{assumption:02} also imply
 \begin{equation}\label{eq:delta_x3}
\begin{split}
\EE \Big [ \| \widetilde \nabla_x L(x_{t},y_{t},\gamma_{t+1},\omega_t^1,\xi_t^2)  \|^2 \Big ] 
& \leq  2 C_f^2 + 2 C_h^2 \EE  [\| \gamma_{t+1} \|^2] .
\end{split}
\end{equation}

Next, we focus on terms in $U_t(x,\gamma)$ involving $\gamma$.
By using the independency between $\gamma_{t}$ and $ H(x_{t}) -  h(x_{t},\xi_t^1 ) $ and Lemma \ref{lemma:sum_martingale} (a), we have for all $\gamma$ satisfying $\EE[\|\gamma\|^2] < +\infty$ that 
\begin{equation}
\label{eq:hgamma}
\EE\Big[ 
\sum_{t=0}^{K-1} \big (  H(x_{t}) -  h(x_{t},\xi_t^1 ) \big )^\top (\gamma - \gamma_t) 
\Big] \le \sqrt{K} \EE[\| \gamma\|] \sigma_h.
\end{equation}
Moreover, Assumption \ref{assumption:02} implies that
\begin{equation}\label{eq:Hh}
    \EE \big[ \|H(x_t) - h(x_t,\xi_t^1)\|^2 \big] \le \sigma_h^2.
\end{equation}
Combining \eqref{eq:delta_x1}, \eqref{eq:delta_x4}, \eqref{eq:delta_x3}, \eqref{eq:hgamma} and \eqref{eq:Hh} with the definition of $U_t$, we obtain the following inequality 
\[\EE\Big[ \sum_{t=0}^{K-1} U_t(x,\gamma) \Big] \le \frac{\eta_0}{2}\EE[\|x\|^2] + \sum_{t=0}^{K-1} \frac{4}{\eta_t}\big(C_f^2 + C_h^2 \EE[\|\gamma_{t+1}\|^2]\big) + \sqrt{K}\EE[\|\gamma\|]\sigma_h + \frac{K\sigma_h^2}{2\beta_0}.\]
The desired inequality of $V_t$ follows in the similar way.
\QED


\subsection{Proof of Proposition \ref{prop:boundedness} }\label{sec:app_A2}

We present a technical result from Lemma 2.8 of \cite{boob2019stochastic}.
\begin{lemma}\label{lemma:recursive_bound}
Let $\{a_t\}_{t\geq 0}$ be a nonnegative sequence and $b_1,b_2 \geq 0 $ be two constants such that $a_0 \leq b_1$. Suppose for all $K\geq 1$, it holds that 
$$
a_K \leq b_1 + b_2 \sum_{t=0}^{K-1}a_t.
$$
Then we have $a_K \leq b_1 (1+b_2)^{K}$ for all $K \geq 1$.
\end{lemma}

 \textit{Proof of Proposition \ref{prop:boundedness}:}
By setting $x= x^*$, $y = y^*$, $\gamma = \gamma^*$ and $\lambda = \lambda^*$, and
using the minimax relationship \eqref{eq:min_max},  we have 
$$
0  \leq \EE[\sum_{t=0}^{K-1} Q(z_{t+1},z^*)] = \EE \Big [ \sum_{t=0}^{K-1}  \big(\cL(x_{t+1}, y^*,\gamma^*,\lambda_{t+1}) -  \cL(x^*, y_{t+1}, \gamma_{t+1},\lambda^*) \big) \Big ], \quad K = 1,\ldots, N.
 $$
Combining the above inequality with Theorem \ref{thm:1}, we have 
  \begin{equation*}
\begin{split}
& \frac{\beta_0}{2} \EE[ \| \gamma^* - \gamma_K\|^2]  + \frac{\eta_0}{2}  \EE [ \| \lambda_K - \lambda^*\|^2 ] + \frac{\kappa_0}{2}\EE[\|x^* - x_K\|^2] + \frac{\eta_0}{2}\EE[\|y^* - y_K\|^2]\\
& \leq  \frac{3K C_f^2 }{2\eta_0}   + \frac{\beta_0}{2} \EE [ \| \gamma_0 - \gamma^*\|^2  ]    + \sqrt{K} \|\gamma^*\| \sigma_h
 + \frac{ K \sigma_h^2 }{2\beta_0 } + \frac{3K \|\gamma^*\|^2 C_h^2 }{2\eta_0}  +  \frac{\eta_0}{2}  \| x_0 -x^*\|^2  +  \frac{\eta_0}{2}\| x^*\|^2   \\
 & \quad + \frac{3K C_f^2 }{2\kappa_0}  + \frac{\alpha_0}{2}  \| \lambda_0 - \lambda^*\|^2  + \sqrt{K} \|\lambda^*\| \sigma_g
 + \frac{ K \sigma_g^2 }{2\alpha_0 } + \frac{3K \|\lambda^*\|^2 C_g^2 }{2\kappa_0}  + \frac{\kappa_0}{2} \| y_0 -y^*\|^2  + \frac{\kappa_0}{2}  \| y^*\|^2   \\
 & \quad +  \sum_{t=0}^{K-1} \frac{4}{\eta_t} (C_f^2 + C_h^2\EE[ \| \gamma_{t+1}\|^2] )      + \sum_{t=0}^{K-1} \frac{4}{\kappa_t} (C_f^2 + C_g^2\EE[ \| \lambda_{t+1}\|^2] )   .
\end{split}
\end{equation*}
By using the facts that $\| a\|^2 \leq 2 \| a - b\|^2 + 2 \| b\|^2$, we further obtain 
  \begin{equation}
  \label{eq:boundxy}
\begin{split}
& \frac{\beta_0}{4} \EE [ \| \lambda_K \|^2 ] + \frac{\alpha_0}{4} \EE [ \| \gamma_K \|^2 ] \\
&\leq \frac{\beta_0}{4} \EE [ \| \lambda_K \|^2 ] + \frac{\alpha_0}{4} \EE [ \| \gamma_K \|^2 ] + \frac{\eta_0}{4} \EE [ \| x_K \|^2 ] + \frac{\kappa_0}{4} \EE [ \| y_K \|^2 ] 
\\
& \leq  \frac{3K C_f^2 }{2\eta_0}   + \frac{\beta_0}{2}  \| \gamma_0 - \gamma^*\|^2     + \sqrt{K} \|\gamma^*\| \sigma_h
 + \frac{ K \sigma_h^2 }{2\beta_0 } + \frac{3K \|\gamma^*\|^2 C_h^2 }{2\eta_0}  +  \frac{\eta_0}{2}  \| x_0 -x^*\|^2  +  \eta_0 \| x^*\|^2   \\
 & \quad + \frac{3K C_f^2 }{2\kappa_0}  + \frac{\alpha_0}{2}  \| \lambda_0 - \lambda^*\|^2  + \sqrt{K} \|\lambda^*\| \sigma_g
 + \frac{ K \sigma_g^2 }{2\alpha_0 } + \frac{3K \|\lambda^*\|^2 C_g^2 }{2\kappa_0}  + \frac{\kappa_0}{2} \| y_0 -y^*\|^2  + \kappa_0 \| y^*\|^2   \\
 & \quad +  \sum_{t=0}^{K-1} \frac{4}{\eta_t} (C_f^2 + C_h^2\EE[ \| \gamma_{t+1}\|^2] )      + \sum_{t=0}^{K-1} \frac{4}{\kappa_t} (C_f^2 + C_g^2\EE[ \| \lambda_{t+1}\|^2] )   + \frac{\beta_0}{2}  \| \lambda^* \|^2  + \frac{\alpha_0}{2}  \| \gamma^* \|^2 .
\end{split}
\end{equation}
By setting $\beta_t = \alpha_t =4\sqrt{N} $, $  \eta_t =  4C_h^2  \sqrt{N}$, and $\kappa_t = 4C_g^2 \sqrt{N}$, and dividing $\sqrt{N}$ on both sides of the above inequality, we further have 
  \begin{equation} \label{eq:gamma_lambda_01}
\begin{split}
  \EE [ \| \lambda_K \|^2 ] +  \EE [ \| \gamma_K \|^2 ]  & \le \EE [ \| \lambda_K \|^2 ] +  \EE [ \| \gamma_K \|^2 ] + C_h^2 \EE[\|x_K\|^2] + C_g^2 \EE[\|y_K\|^2] \\
  &\leq R_K +\frac{1}{N} \sum_{t=0}^{K-1} (\EE[ \| \gamma_{t+1} \|^2] +  \EE [ \| \lambda_{t+1}\|^2])
\end{split}
\end{equation}
where 
\begin{equation*}
\begin{split}
R_K & := \frac{11K C_f^2 }{8 C_h^2 N }   +2 \| \gamma_0 - \gamma^*\|^2     + \frac{\sqrt{K} \|\gamma^* \| }{\sqrt{N} }\sigma_h
 + \frac{ K \sigma_h^2 }{8N } + (\frac{3K  }{8N} + 2) \|\gamma^*\|^2  +  2 C_h^2 \| x_0 -x^*\|^2  +  4 C_h^2  \| x^*\|^2   \\
 & \quad + \frac{11K C_f^2 }{8 C_g^2 N}  + 2  \| \lambda_0 - \lambda^*\|^2  + \frac{ \sqrt{K} \| \lambda^*\| }{\sqrt{N}}\sigma_g
 + \frac{ K \sigma_g^2 }{8N } + (\frac{3K  }{8N} + 2)  \|\lambda^*\|^2  + 2 C_g^2 \| y_0 -y^*\|^2  + 4 C_g^2  \| y^*\|^2.
\end{split}
\end{equation*}
Furthermore, let $R$ be the constant defined in \eqref{def:R}, we observe that 
$R_t \leq R$ for $t=1,2,\cdots,N-1$. Therefore, by rearranging the terms within \eqref{eq:gamma_lambda_01}, we have 
  \begin{equation*}
\begin{split}
  (1- \frac{1}{N}) \Big ( \EE [ \| \lambda_K \|^2 ] +  \EE [ \| \gamma_K \|^2 ] \Big )   \leq R +\frac{1}{N} \sum_{t=1}^{K-1} \Big  ( \EE[ \| \gamma_{t} \|^2] +  \EE [ \| \lambda_{t}\|^2] \Big ).
\end{split}
\end{equation*}
For $N \geq 2$, the above inequality further implies 
  \begin{equation*}
\begin{split}
 \EE [ \| \lambda_K \|^2 ] +  \EE [ \| \gamma_K \|^2 ]  & \leq \Big (1- \frac{1}{N} \Big  )^{-1} \left ( R +\frac{1}{N} \sum_{t=1}^{K-1} \Big  ( \EE[ \| \gamma_{t} \|^2] +  \EE [ \| \lambda_{t}\|^2] \Big ) \right ) \\
& \leq 2 R +\frac{2}{N} \sum_{t=1}^{K-1} \Big  ( \EE[ \| \gamma_{t} \|^2] + \EE [ \| \lambda_{t}\|^2] \Big ) .
\end{split}
\end{equation*}
By using Lemma \ref{lemma:recursive_bound}, setting $a_t =   \EE [ \| \lambda_t \|^2 ] +  \EE [ \| \gamma_t \|^2 ] $, $b_1 = 2R$, and $b_2 = \tfrac{2}{N}$, 
 we conclude that 
 \begin{equation*}
 \begin{split}
 \EE [ \| \lambda_K \|^2 ] +  \EE [ \| \gamma_K \|^2 ]     \leq 2R \prod_{t=1}^{K-1}\big (1 +  \frac{2}{N} \big )  = 2R (1+ \frac{2}{N})^{K-1} \leq 2R e^{2},
\end{split}
 \end{equation*}
 where the last inequality uses the fact that $1\leq K \leq N$. Then, \eqref{eq:gamma_lambda_01} further implies that
 \[
\EE [ \| \lambda_K \|^2 ] +  \EE [ \| \gamma_K \|^2 ] + C_h^2 \EE[\|x_K\|^2] + C_g^2 \EE[\|y_K\|^2] \le (2e^2 + 1)R, \quad \forall \, 1 \le K \le N. \]
This completes the proof. \QED


\subsection{Proof of Theorem~\ref{thm:rates_basic}}
\textit{Proof:}
Denote 
$\bar \lambda_N = \frac{1}{N} \sum_{t=1}^N \lambda_t$, and $\bar \gamma_N = \frac{1}{N} \sum_{t=1}^N \gamma_t$. 
We start by proving inequality \eqref{eq:objieq}. 
By using Theorem \ref{thm:boundsQ} and setting $\gamma = 0$ and $\lambda = 0$, we have for any $(x,y)\in \cX\times\cY$ satisfying $\EE[\|x\|^2] < +\infty$ and $\EE[\|y\|^2]<+\infty$ that
 \begin{equation*}
 \begin{split}
&   \frac{1}{N} \EE \Big [ \sum_{t=1}^N  \big(\cL(x_{t}, y,0,\lambda_{t}) -  \cL(x, y_{t}, \gamma_{t},0) \big) \Big ]  =  \frac{1}{N} \EE\Big[\sum_{t=1}^N Q(z_{t},(x,y,0,0))\Big]\\
   \leq{}& \frac{1}{\sqrt{N}}  \Big ( 2Re^2 +  \frac{11C_f^2 }{8 C_h^2  }   +2  \| \gamma_0 \|^2  
 + \frac{  \sigma_h^2 }{8 }  +  2 C_h^2 \EE [\| x_0 -x \|^2]  +  2 C_h^2 \EE [ \| x \|^2 ]       \Big ) 
 \\
 & + \frac{1}{\sqrt{N} } \Big ( \frac{11 C_f^2 }{8 C_g^2 }  + 2 \|  \lambda_0  \|^2   + \frac{  \sigma_g^2 }{8 }  + 2  C_g^2 \EE [ \| y_0 -y \|^2]  + 2 C_g^2 \EE [  \| y \|^2     ]     \Big ).
\end{split}
\end{equation*}
Since  $\cL(x,y,\gamma,\lambda)$ is convex in $x, \lambda$ and concave in $y, \gamma$, we have 
\begin{equation*}
\begin{split}
&\frac{1}{N} \sum_{t=1}^N\Big ( \cL(x_{t}, y, 0,\lambda_{t}) -  \cL(x, y_{t}, \gamma_{t},0) \Big )  \\
  \geq {}&  \cL(\bar x_N, y,0,\bar \lambda_N) -  \cL(x, \bar y_N, \bar \gamma_N,0)   \\
  = {}& F(\bar x_N,y) - G(y)^\top \bar \lambda_N - F(x,\bar y_N) - H(x)^\top \bar \gamma_N   \geq  F(\bar x_N,y)  - F(x,\bar y_N),
\end{split}
\end{equation*}
where the last inequality holds since $ G(y)^\top \bar \lambda_N \le 0$ and $ H(x)^\top \bar \gamma_N\le 0$ for any feasible $(x,y)$. Then we conclude that 
\begin{equation}\label{eq:upperF}
\begin{split}
& \EE \Big [ F(\bar x_N,y )  - F(x,\bar y_N) \Big ] \\
 \leq {} & \frac{1}{\sqrt{N}}  \Big ( 2Re^2 +  \frac{11C_f^2 }{8 C_h^2  }   +2  \| \gamma_0 \|^2    
 + \frac{  \sigma_h^2 }{8 }  +  2 C_h^2 \EE[\| x_0 -x \|^2]  +  2 C_h^2 \EE [ \| x \|^2 ]       \Big ) 
 \\
 & + \frac{1}{\sqrt{N} } \Big ( \frac{11 C_f^2 }{8 C_g^2 }  + 2 \|  \lambda_0  \|^2    + \frac{  \sigma_g^2 }{8 }  + 2  C_g^2 \EE [ \| y_0 -y \|^2]  + 2 C_g^2 \EE [  \| y \|^2     ]     \Big ).
\end{split}
\end{equation}

Next, we derive the upper bounds for feasibility residuals. Let $\tilde \gamma = ( \| \gamma^*\|_2 +1)\frac{  H(\bar x_N)_+ }{ \| H(\bar x_N)_+\|_2 }$ and $\tilde \lambda  = (\| \lambda^*\|_2 + 1) \frac{  G(\bar y_N)_+ }{ \| G(\bar y_N)_+\|_2 } $ and consider the reference point $(x^*,y^*,\tilde \gamma, \tilde \lambda)$, we obtain  
\begin{equation}\label{eq:feasibility_1}
\begin{split}
& \frac{1}{N}\sum_{t=1}^N  \Big ( \cL(x_{t}, y^*,\tilde \gamma,\lambda_{t}) -  \cL(x^*, y_{t}, \gamma_{t},\tilde \lambda) \Big )  \\
\geq {}& \cL(\bar x_N, y^*,\tilde \gamma,\bar \lambda_N) -  \cL(x^*, \bar y_N, \bar \gamma_N,\tilde \lambda) \\
 = {}& F(\bar x_N,y^*)  + 
 H(\bar x_N)^\top \tilde \gamma - G(y^*)^\top \bar \lambda_N 
 - \Big[ F(x^*,\bar y_N) +  H(x^*)^\top  \bar \gamma_N
 - G(\bar y_N)^\top \tilde \lambda \Big ] \\
 \geq {}& F(\bar x_N,y^*)  + H(\bar x_N)^\top \tilde \gamma - F(x^*,\bar y_N)  + G(\bar y_N)^\top \tilde \lambda  \\
 = {}& F(\bar x_N,y^*)  + \big  (\| \gamma^*\|_2 + 1 \big ) \| H(\bar x_N)_+ \|_2 - F(x^*,\bar y_N)  + \big (\| \lambda^*\|_2  + 1 \big )\| G(\bar y_N)_+ \|_2 ,
\end{split}
\end{equation}
where the last equality follows from the facts that $H(\bar x_N)^\top H(\bar x_N)_+ = \| H(\bar x_N)_+ \|_2^2$ and $G(\bar y_N)^\top G(\bar y_N)_+ = \| G(\bar y_N)_+ \|_2^2$.
Meanwhile, 
from the minimax relationship \eqref{eq:min_max}
$$ \cL(\bar x_N, y^*, \gamma^*,\lambda^*) \geq  \cL(x^*, y^*,\gamma^*,\lambda^*)  \geq \cL(x^*, \bar y_N, \gamma^*, \lambda^*),$$
it holds that 
\begin{equation}\label{eq:feasibility_2}
\begin{split}
0 & \leq \ \cL(\bar x_N, y^*,\gamma^*,\lambda^*) -  \cL(x^*, \bar y_N, \gamma^*,\lambda^*) \\
 & = F(\bar x_N,y^*)  +  H(\bar x_N)^\top \gamma^*  - F(x^*,\bar y_N)  +  G(\bar y_N)^\top \lambda^*  \\
 & \leq F(\bar x_N,y^*)  +  H(\bar x_N)_+^\top \gamma^*  - F(x^*,\bar y_N)  + G(\bar y_N)_+^\top  \lambda^* \\
&  \leq F(\bar x_N,y^*)  +  \| \gamma^*\|_2 \| H(\bar x_N)_+ \|_2 - F(x^*,\bar y_N)  + \| \lambda^*\|_2  \| G(\bar y_N)_+ \|_2,
\end{split}
\end{equation}
where the second inequality holds since $\gamma^* \geq 0, H(\bar x_N) \leq H(\bar x_N)_+$, and $\lambda^* \geq 0, G(\bar y_N) \leq G(\bar y_N)_+$. 
Substituting \eqref{eq:feasibility_2} into \eqref{eq:feasibility_1} and then taking expectation, we have 
\begin{equation*}
\begin{split}
\EE [ \| H(\bar x_N)_+\|_2 ]  +     \EE [ \| G(\bar y_N)_+ \|_2 ]  \leq \frac{1}{N}   \EE \Big[\sum_{t=1}^N \big ( \cL(x_{t}, y^*,\tilde \gamma,\lambda_{t}) -  \cL(x^*, y_{t}, \gamma_{t},\tilde \lambda) \big ) \Big]
.
\end{split}
\end{equation*}
Since $\|\tilde \lambda \| = \|\lambda^*\| + 1$ and $\| \tilde \gamma\| =  \|\gamma^*\| + 1$, it holds from Theorem \ref{thm:boundsQ} that
\begin{equation*}
\begin{split}
& \EE [ \| H(\bar x_N)_+\|_2 ]  +     \EE [ \| G(\bar y_N)_+ \|_2 ]  \\
 \leq {}&  \frac{1}{\sqrt{N}} \Big ( 2Re^2 +  \frac{11C_f^2 }{8 C_h^2  }   + 4  \| \gamma_0 \|^2 + \frac{35}{8}  (\| \lambda^* \| +1)^2   +(\| \lambda^* \| +1) \sigma_h
 + \frac{  \sigma_h^2 }{8 }   +  2 C_h^2 \| x_0 -x^* \|^2  +  2 C_h^2 \| x^* \|^2       \Big ) 
 \\
 & + \frac{1}{\sqrt{N}} \Big ( \frac{11 C_f^2 }{8 C_g^2 }  + 4  \| \lambda_0\|^2 + \frac{35}{8}  (\| \lambda^* \| +1)^2  + (\| \lambda^* \| +1) \sigma_g
 + \frac{  \sigma_g^2 }{8 }  + 2  C_g^2  \| y_0 -y^* \|^2  + 2 C_g^2   \| y^* \|^2       \Big )
 ,
\end{split}
\end{equation*}
where the last inequality follows from the fact that $ \| \gamma_0 - \tilde \gamma\|^2 \leq 2 \| \gamma_0 \|^2 + 2 \| \tilde \gamma\|^2 = 2 \| \gamma_0 \|^2 + 2 (\| \gamma^*\|+1)^2 $ and $\| \lambda_0 - \tilde\lambda\|^2  \leq 2 \| \lambda_0 \|^2 + 2 (\| \lambda^*\|+1)^2 $.

Lastly, by setting $(x,y) = (x^*,y^*)$ and combining the above inequality with \eqref{eq:upperF} and the lower bound provided in Lemma \ref{lemma:obj_lower_bound}, we conclude that there exist two constants $C_1, C_2 >0$ such that 
\begin{equation*}
-\frac{C_1}{\sqrt{N}} \leq \EE \Big [ F(\bar x_N,y^* )  - F(x^*,\bar y_N) \Big ] \leq \frac{C_2}{\sqrt{N}}.
\end{equation*}
This completes the proof. 
\QED

\subsection{Proof of Corollary \ref{cor:strong_gap_basic}}
\textit{Proof:}
The first inequality can be easily obtained by combining Theorem \ref{thm:rates_basic} with the fact that 
$$
 F(\bar x_N, y^{\star}(\bar x_N) )  - F(x^{\star}(\bar y_N),\bar y_N)  \geq F(\bar x_N, y^* )  - F(x^*,\bar y_N),
$$
and the second inequality can be derived by setting $(x,y) = (x^*(\bar y_N), y^*(\bar x_N))$ in Theorem \ref{thm:rates_basic} and using the boundedness of  $\widetilde \cX$ and $ \widetilde \cY$. 
\QED

\section{Proof of Results in Section \ref{sec:adp_alg} }

\subsection{Proof of Lemma~\ref{lem:delta_gamma_diminish}}
\textit{Proof:}
Recall the decompostion of $\Delta_x^{t+1}$ in \eqref{eq:delta_x_split}. Under the update rule of $x_{t+1}$ in Algorithm \ref{alg:2}, for any $x \in \cX$, we obtain from the three-point lemma \ref{lemma:three_point}  that
\begin{equation*}
\begin{split}
&\cH(x_{t},y_{t},\gamma_{t+1},\omega_t^1,\xi_t^2 )^\top(x_{t+1}  -x) \\
 \leq {}& \frac{\eta_{t}}{2} \| x_{t} -x\|^2 - \frac{\eta_{t} }{2} \| x_{t} - x_{t+1}\|^2 - \frac{\eta_{t} + \rho_t  }{2} \| x_{t+1} - x\|^2 
- \frac{\rho_t}{2} \|x_{t+1} - x_0 \|^2 + \frac{\rho_t}{2} \|x - x_0 \|^2.
\end{split}
\end{equation*}
Similar to \eqref{eq:sumdeltax}, it holds for all $x\in \cX$ that 
\begin{align*}
    \Delta_x^{t+1} + \frac{\eta_t}{3} \norm{x_t  - x_{t+1}}^2 \le {}& \frac{\eta_{t}}{2} \| x_{t} -x\|^2  - \frac{\eta_{t} + \rho_t}{2} \| x_{t+1} - x\|^2 + \frac{\rho_t}{2}\norm{x - x_0}^2
    \\ {}& + \big( \widetilde \nabla_x L (x_{t},y_{t},\gamma_{t+1},\omega_t^1,\xi_t^2) -  \widetilde \nabla_x \cL(x_{t},y_{t},\gamma_{t+1}) \big)^\top ( x -x_{t}) \nonumber \\
    {}& + \frac{3 \| \widetilde \nabla_x L(x_{t},y_{t},\gamma_{t+1},\omega_t^1,\xi_t^2)  \|^2}{2\eta_{t}}.
\end{align*}
Summing the above inequality over $t=0,1,\ldots,K-1$ and noting that $\eta_{t+1} \le \eta_t + \rho_t$, we know that 
\begin{align}\label{eq:sum_deltax_app}
    &\sum_{t=0}^{K-1} \big(\Delta_x^{t+1} + \frac{\eta_t}{3} \norm{x_t  - x_{t+1}}^2 \big) + \frac{\eta_K}{2}\norm{x_K - x}^2 \nonumber \\ 
    \le {}& (\frac{\eta_{0}}{2} + \sum_{t=0}^{K-1} \frac{\rho_t}{2}) \| x_{0} -x\|^2  + \sum_{t=0}^{K-1} \big( \widetilde \nabla_x L(x_{t},y_{t},\gamma_{t+1},\omega_t^1,\xi_t^2) -  \widetilde \nabla_x \cL(x_{t},y_{t},\gamma_{t+1}) \big)^\top ( x -x_{t})\\ {}&
    + \sum_{t=0}^{K-1} \frac{3 \| \widetilde \nabla_x L(x_{t},y_{t},\gamma_{t+1},\omega_t^1,\xi_t^2)  \|^2}{2\eta_{t}}, \quad \forall\, x\in \cX. \nonumber
\end{align}

Recall the decompostion of $\Delta_\gamma^{t+1}$ in \eqref{eq:delta_gamma}. From the update rule of $\gamma_t$ in \eqref{eq:gamma_diminishing}, we know from Lemma \ref{lemma:three_point} that for all $\gamma\in\RR_+^{m_1}$, 
\begin{equation*}
\begin{split}
& \Delta_{\gamma,1}^{t+1}=  -  h(x_{t},\xi_t^1 )^\top (  \gamma_{t+1} - \gamma  )  \\
 \leq{}& \frac{\beta_{t}}{2}  \| \gamma_{t} - \gamma\|^2 -  \frac{\beta_{t}}{2}  \| \gamma_{t} - \gamma_{t+1}\|^2 -  \frac{\beta_{t}+ \tau_t }{2}  \| \gamma_{t+1} - \gamma\|^2 -\frac{\tau_t}{2}\| \gamma_{t+1} - \gamma_0\|^2 + \frac{\tau_t}{2} \| \gamma - \gamma_0\|^2. 
\end{split}
\end{equation*} 
Substituting the above inequality into \eqref{eq:delta_gamma}, and using \eqref{eq:delta_gamma_2oE} and \eqref{eq:delta_gamma_3}, we see that
\begin{equation*}
    \begin{aligned}
       \Delta_{\gamma}^{t+1}  \le {}& \frac{\beta_t}{2}\|\gamma_t - \gamma\|^2 - \frac{\beta_t + \tau_t}{2}\|\gamma_{t+1} - \gamma\|^2  - \frac{\tau_t}{2} \|\gamma_{t+1} - \gamma_0\|^2 + \frac{\tau_t}{2}\|\gamma - \gamma_0\|^2 \\
       & + \big (  H(x_{t}) -  h(x_{t},\xi_t^1 ) \big )^\top (\gamma - \gamma_t) +  \frac{ \|H(x_{t}) -  h(x_{t},\xi_t^1 )\|^2 }{2\beta_{t} } \\
       & +   \frac{3 \| \gamma\|^2 C_h^2 }{2\eta_{t}} + \frac{ \eta_{t} }{6} \| x_{t+1} - x_{t}\|^2.
    \end{aligned}
\end{equation*}
Summing the above inequality over $t=0,1,\ldots,K-1$ and noting that $\beta_{t+1} \le \beta_t + \tau_t$, we have that 
\begin{equation}\label{eq:sum_deltagamma_app}
    \begin{aligned}
       &\sum_{t=0}^{K-1}\big(\Delta_{\gamma}^{t+1} -  \frac{ \eta_{t} }{6} \| x_{t+1} - x_{t}\|^2 \big) + \frac{\beta_K}{2}\|\gamma_K - \gamma\|^2 \\
       \le {}& \big(\frac{\beta_0}{2} + \sum_{t=0}^{K-1} \frac{\tau_t}{2}\big) \|\gamma_0 - \gamma\|^2  - \sum_{t=0}^{K-1} \frac{\tau_t}{2} \|\gamma_{t+1} - \gamma_0\|^2  + \sum_{t=0}^{K-1} \big (  H(x_{t}) -  h(x_{t},\xi_t^1 ) \big )^\top (\gamma - \gamma_t) 
       \\
       & + \sum_{t=0}^{K-1}  \frac{ \|H(x_{t}) -  h(x_{t},\xi_t^1 )\|^2 }{2\beta_{t} } + \sum_{t=0}^{K-1}  \frac{3 \| \gamma\|^2 C_h^2 }{2\eta_{t}}, \quad \forall\, \gamma \in \RR_+^{m_1}.
    \end{aligned}
\end{equation}
It then holds from \eqref{eq:sum_deltax_app} and \eqref{eq:sum_deltagamma_app} that
\begin{equation}\label{eq:sum_deltaxgamma_app}
    \begin{aligned}
        &\sum_{t=0}^{K-1} \big(\Delta_{x}^{t+1} + \Delta_{\gamma}^{t+1} + \frac{\eta_t}{6}\|x_{t+1} - x_t\|^2) + \frac{\eta_K}{2}\|x_K - x\|^2 + \frac{\beta_K}{2}\|\gamma_K - \gamma\|^2 \\
        \le {}& (\frac{\eta_{0}}{2} + \sum_{t=0}^{K-1} \frac{\rho_t}{2}) \| x_{t} -x\|^2 + \big(\frac{\beta_0}{2} + \sum_{t=0}^{K-1} \frac{\tau_t}{2}\big) \|\gamma_0 - \gamma\|^2 +  \sum_{t=0}^{K-1} \big(\frac{3 \| \gamma\|^2 C_h^2 }{2\eta_{t}} - \frac{\tau_t}{2}\|\gamma_{t+1} - \gamma_0\|^2 \big) \\
        & + \sum_{t=0}^{K-1} U_t(x,\gamma), \quad\forall\, (x,\gamma)\in \cX\times \RR_+^{m_1},
    \end{aligned}
\end{equation}
where $U_t(x,\gamma)$ is defined in \eqref{eq:Uxgamma}. 

Next, we perform similar analysis for $\Delta_y^{t+1}$ and $\Delta_\gamma^{t+1}$ and obtain
\begin{equation}\label{eq:sum_deltaylambda_app}
    \begin{aligned}
        &\sum_{t=0}^{K-1} \big(\Delta_{y}^{t+1} + \Delta_{\lambda}^{t+1} + \frac{\kappa_t}{6}\|y_{t+1} - y_t\|^2) + \frac{\kappa_K}{2}\|y_K - y\|^2 + \frac{\alpha_K}{2}\|\lambda_K - \lambda\|^2 \\
        \le {}& (\frac{\kappa_{0}}{2} + \sum_{t=0}^{K-1} \frac{\phi_t}{2}) \| y_{t} -y\|^2 + \big(\frac{\alpha_0}{2} + \sum_{t=0}^{K-1} \frac{\nu_t}{2}\big) \|\lambda_0 - \lambda\|^2 +  \sum_{t=0}^{K-1} \big(\frac{3 \| \lambda\|^2 C_g^2 }{2\kappa_{t}} - \frac{\nu_t}{2}\|\lambda_{t+1} - \lambda_0\|^2 \big) \\
        & + \sum_{t=0}^{K-1} V_t(y,\lambda), \quad\forall\, (y,\lambda)\in \cY\times \RR_+^{m_2}
    \end{aligned}
\end{equation}
with $V_t(y,\lambda)$ defined in \eqref{eq:Vylambda}.
Combining \eqref{eq:sum_deltaxgamma_app} and \eqref{eq:sum_deltaylambda_app}, we obtain the deried inequality and complete the proof. \QED

\subsection{Proof of Lemma \ref{lem:EUVadp}}
\textit{Proof:} The proof here is quite similar to the one for Lemma \ref{lem:EUV} in Section \ref{sec:apen_UtVt}. Similar to \eqref{eq:delta_x4}, for all $x\in\cX$ satisfying $\EE[\|x\|^2] < +\infty$, we know from Lemma \ref{lemma:sum_martingale} (c) that
\begin{equation}\label{eq:ut_app_4}
	\begin{split}
	& \EE \Big [\sum_{t=0}^{K-1}  \big (  \widetilde \nabla_x L(x_{t},y_{t},\gamma_{t+1},\omega_t^1,\xi_t^2) -  \widetilde \nabla_x \cL(x_{t},y_{t},\gamma_{t+1}) \big )^\top x \Big ] \\
	\leq{}& \big(\frac{ \eta_0}{2} + \sum_{t=0}^K\frac{\rho_t}{2} \big) \EE[ \| x\|^2] + \sum_{t=0}^{K-1} \frac{1}{2\eta_t} \EE [ \| \widetilde \nabla_x L(x_{t},y_{t},\gamma_{t+1},\omega_t^1,\xi_t^2) -  \widetilde \nabla_x 
	\cL(x_{t},y_{t},\gamma_{t+1}) \|^2] 
 \\
	\leq&{} \big(\frac{ \eta_0}{2} + \sum_{t=0}^K\frac{\rho_t}{2} \big)\EE[ \| x\|^2] + \sum_{t=0}^{K-1} \frac{1}{\eta_t} (C_f^2 + C_h^2\EE[ \| \gamma_{t+1}\|^2]).
	\end{split}
	\end{equation}
The desired inequality of $U_t$ then follows from \eqref{eq:ut_app_4}, \eqref{eq:delta_x3}, \eqref{eq:hgamma} and \eqref{eq:Hh} and the definition in \eqref{eq:Uxgamma}. The inequality of $V_t$ can be proved in the same way. We thus complete the proof. \QED
\subsection{Proof of Theorem \ref{thm:diminishing_1}}  \label{sec:proof_of_thm_diminishing_1}
\textit{Proof:} 
From Lemmas \ref{lem:delta_gamma_diminish} and \ref{lem:EUV}, by taking expectations and setting $\lambda_0 = {\bf 0}, \gamma_0 = {\bf 0}$, we obtain that
\begin{equation}\label{eq:sum_exp_app_4}
\begin{aligned}
&\EE\Big[\sum_{t=0}^{K-1} Q(z_{t+1},z) + \frac{\eta_{K}}{2} \norm{x_K - x}^2 + \frac{\beta_{K}}{2}\norm{\gamma_K - \gamma}^2 + \frac{\kappa_{K}}{2}\norm{y_K - y}^2 + 
\frac{\alpha_{K}}{2} \norm{\lambda_K - \lambda}^2 \Big]
\\  
\le{}& \EE\Big[(\frac{\eta_0}{2} + \sum_{t=0}^{K-1} \frac{\rho_t}{2}) (\norm{x_0 - x}^2 + \norm{x}^2) \Big] + \EE\Big[ \Big(\frac{\beta_0}{2} + \sum_{t=0}^{K-1} \big( \frac{\tau_t}{2} +  \frac{3 C_h^2}{2\eta_t} \big) \Big) \norm{ \gamma}^2 \Big] + \sqrt{K}\EE[\|\gamma\|]\sigma_h \\
& + \EE\Big[\sum_{t=0}^{K-1} \big(\frac{4C_h^2}{\eta_t}- \frac{\tau_t}{2}\big)\|\gamma_{t+1}\|^2 \Big] + \sum_{t=0}^{K-1} 
\big( \frac{4C_f^2}{\eta_t} + \frac{\sigma_h^2}{2\beta_t} \big) \\
& + \EE\Big[(\frac{\kappa_0}{2} + \sum_{t=0}^{K-1} \frac{\phi_t}{2}) (\norm{y_0 - y}^2 + \norm{y}^2) \Big] + \EE\Big[ \Big(\frac{\alpha_0}{2} + \sum_{t=0}^{K-1} \big(\frac{\nu_t}{2} + \frac{3C_g^2}{2\kappa_t} \big)\norm{\lambda}^2  \Big) \Big]  + \sqrt{K}\EE[\|\lambda\|]\sigma_g
\\
& + \EE\Big[\sum_{t=0}^{K-1} \big(\frac{4C_g^2}{\kappa_t}- \frac{\nu_t}{2}\big)\|\gamma_{t+1}\|^2 \Big]
+ \sum_{t=0}^{K-1} 
\big( \frac{4C_f^2}{\kappa_t} + \frac{\sigma_g^2}{2\alpha_t} \big)
\end{aligned}
\end{equation}
for all $(x,y,\gamma, \lambda) \in \cX\times\cY\times\RR_+^{m_1} \times \RR_+^{m_2}$ with bounded second moments.
Meanwhile, it holds from \eqref{eq:adaptive_stepsize} that
\begin{equation}
\label{eq:inequality_stepsize}
\begin{aligned}
& \frac{\eta_0}{2} + \sum_{t=0}^{K-1} \frac{\rho_t}{2} = 8\sqrt{K+2}, \quad  \frac{\beta_0}{2} + \sum_{t=0}^{K-1} \frac{\tau_t}{2}  = \frac{C_h^2\sqrt{K+1}}{
	2}, \quad \sum_{t=0}^{K-1} \frac{1}{\eta_t} = \sum_{t=0}^{K-1} \frac{1}{\kappa_t} \le \frac{\sqrt{K+2}}{8}, \\
& \sum_{t=0}^{K-1} \frac{1}{\alpha_t} \le \frac{2\sqrt{K+1}}{C_g^2}, \quad \sum_{t=0}^{K-1} \frac{1}{\beta_t} \le \frac{2\sqrt{K+1}}{C_h^2}, \quad \frac{4C_h^2}{\eta_t} - \frac{\tau_t}{2} \le 0, \quad \frac{4C_g^2}{\kappa_t}- \frac{\nu_t}{2}  \le 0.
\end{aligned}
\end{equation}
The desired inequality then follows by noting \eqref{eq:adaptive_stepsize}, and substituting \eqref{eq:inequality_stepsize} into \eqref{eq:sum_exp_app_4}.\QED
%
%

\subsection{Proof of Theorem \ref{thm:rates_adp}}
\textit{Proof:} To establish the bound for the objective optimality gap, 
we set $\gamma = {\bf 0}$ and $\lambda = {\bf 0}$ in Theorem \ref{thm:diminishing_1} and adopt similar analysis as in Theorem \ref{thm:rates_basic} to obtain 
\begin{equation*}
\begin{split}
& \EE [F(\bar x_N,y)  - F(x,\bar y_N)] \\
\leq {} & \frac{1}{N} \EE \Big[ \sum_{t=0}^{N-1}  \big (\cL(x_{t+1}, y,0,\lambda_{t+1}) -  \cL(x, y_{t+1}, \gamma_{t+1},0)  \big ) \Big]  \\
\leq {} & \frac{ \sqrt{N+1} }{N}    \Big ( \frac{ \sigma_h^2 }{ C_h^2 }  +  \frac{ \sigma_g^2 }{ C_g^2 } \Big ) 
 +  \frac{ \sqrt{N+2}  }{N}\Big ( 8\EE [ \|x - x_0 \|^2 +  \| x\|^2] +  \frac{11   C_f^2     }{16}  +  8 \EE [ \|y - y_0 \|^2  + \| y \|^2 ]+  \frac{11   C_f^2     }{16}  \Big )
 \end{split}
\end{equation*} 
for all $(x,y)\in\cX\times\cY$ with bounded second moments.

For the bound of feasibility residuals, we choose $\tilde \gamma = ( \| \gamma^*\|_2 +1)\frac{  H(\bar x_N)_+ }{ \| H(\bar x_N)_+\|_2 }$ and $\tilde \lambda  = (\| \lambda^*\|_2 + 1) \frac{  G(\bar y_N)_+ }{ \| G(\bar y_N)_+\|_2 } $. Again by adopting a similar analysis to Theorem \ref{thm:rates_basic}, we have from Theorem \ref{thm:diminishing_1}  that
\begin{equation*}
\begin{split}
& \EE [ \| H(\bar x_N)_+\|_2 ]  +     \EE [ \| G(\bar y_N)_+ \|_2 ] \\
 \leq {}&  \frac{1}{N}\EE \Big[ \sum_{t=0}^{N-1} \big(\cL(x_{t+1}, y^*,\tilde \gamma,\lambda_{t+1}) -  \cL(x^*, y_{t+1}, \gamma_{t+1},\tilde \lambda) \big) \Big]  \\
\leq {}& \frac{ \sqrt{N+2} }{N}\Big (  \frac{11C_h^2}{16}  \EE[\| \tilde \gamma\|^2]  
+ \frac{11C_g^2}{16}  \EE[ \| \tilde \lambda\|^2 ] \Big )  + \frac{1}{ \sqrt{N} } \Big ( \EE[ \| \tilde \gamma \|] \sigma_h + 
\EE[\| \tilde \lambda \|]\sigma_g \Big ) + \frac{\sqrt{N+1}}{N}     \Big ( \frac{ \sigma_h^2 }{ C_h^2 }  +  \frac{ \sigma_g^2 }{ C_g^2 } \Big ) 
\\
& 
\quad +  \frac{ \sqrt{N+2} }{N} \Big ( 8 \|x^* - x_0 \|^2  +  8 \| x^* \|^2 +  \frac{11   C_f^2     }{16}  +  8 \|y^* - y_0 \|^2  + 8 \| y^* \|^2 +  \frac{11   C_f^2     }{16}  \Big ).
\end{split}
\end{equation*} 
Noting that $\| \tilde \gamma \| = \| \gamma^*\|+1$ and $\| \tilde \lambda \| = \| \lambda^*\|+1$, we arrive at the desired inequality. 

The above two inequalities, together with Lemma \ref{lemma:obj_lower_bound}, imply that there exist constants $C_1, C_2 >0$ such that 
\begin{equation*}
-\frac{C_1}{\sqrt{N}} \leq \EE \Big [ F(\bar x_N,y^* )  - F(x^*,\bar y_N) \Big ] \leq \frac{C_2}{\sqrt{N}}.
\end{equation*}
This completes the proof. 
\QED


\section{Additional Numerical Results}\label{app:numerics}
\begin{figure}[t]
    \centering
    \includegraphics[width=0.4\linewidth]{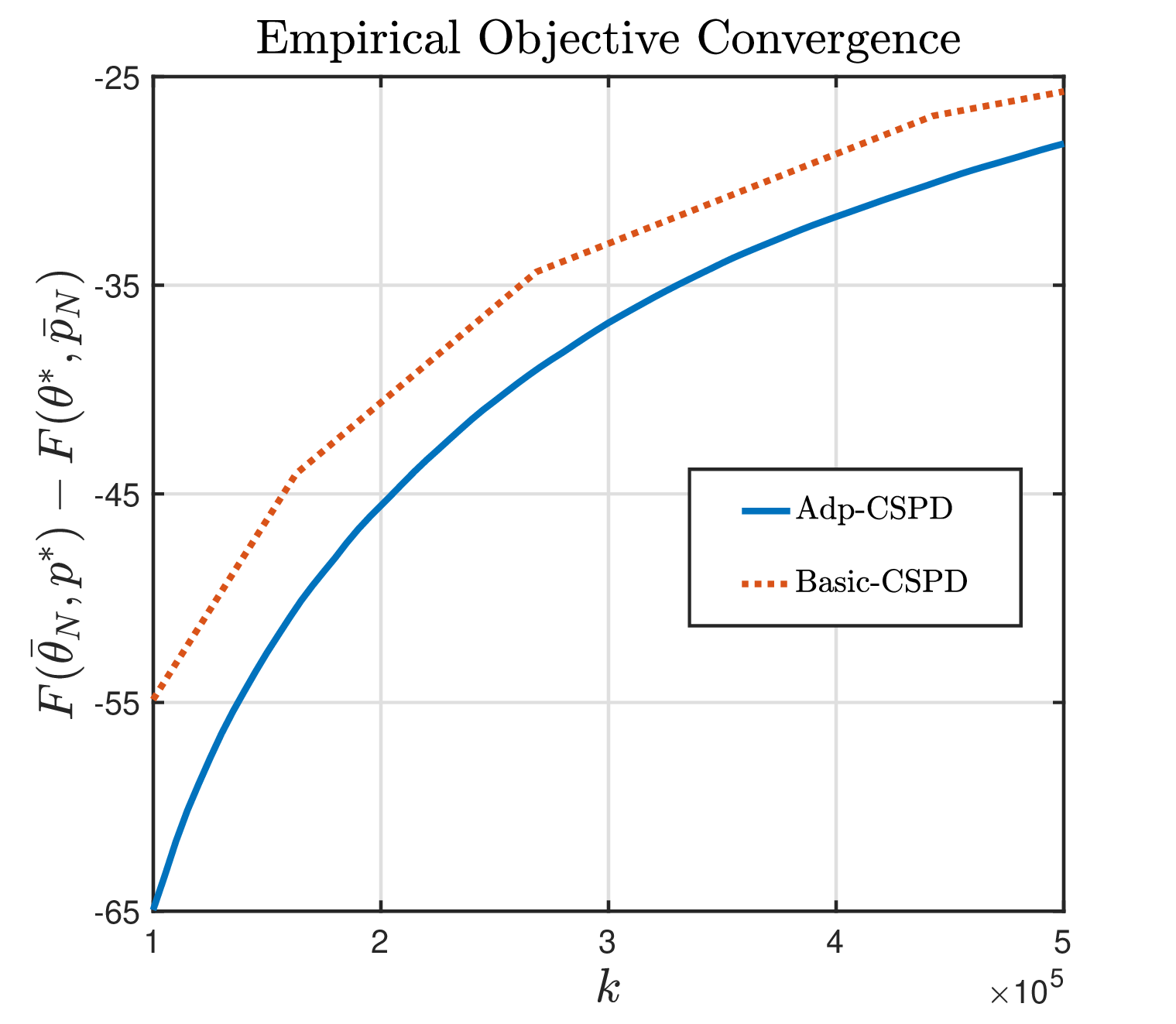}
    \includegraphics[width=0.4\linewidth]{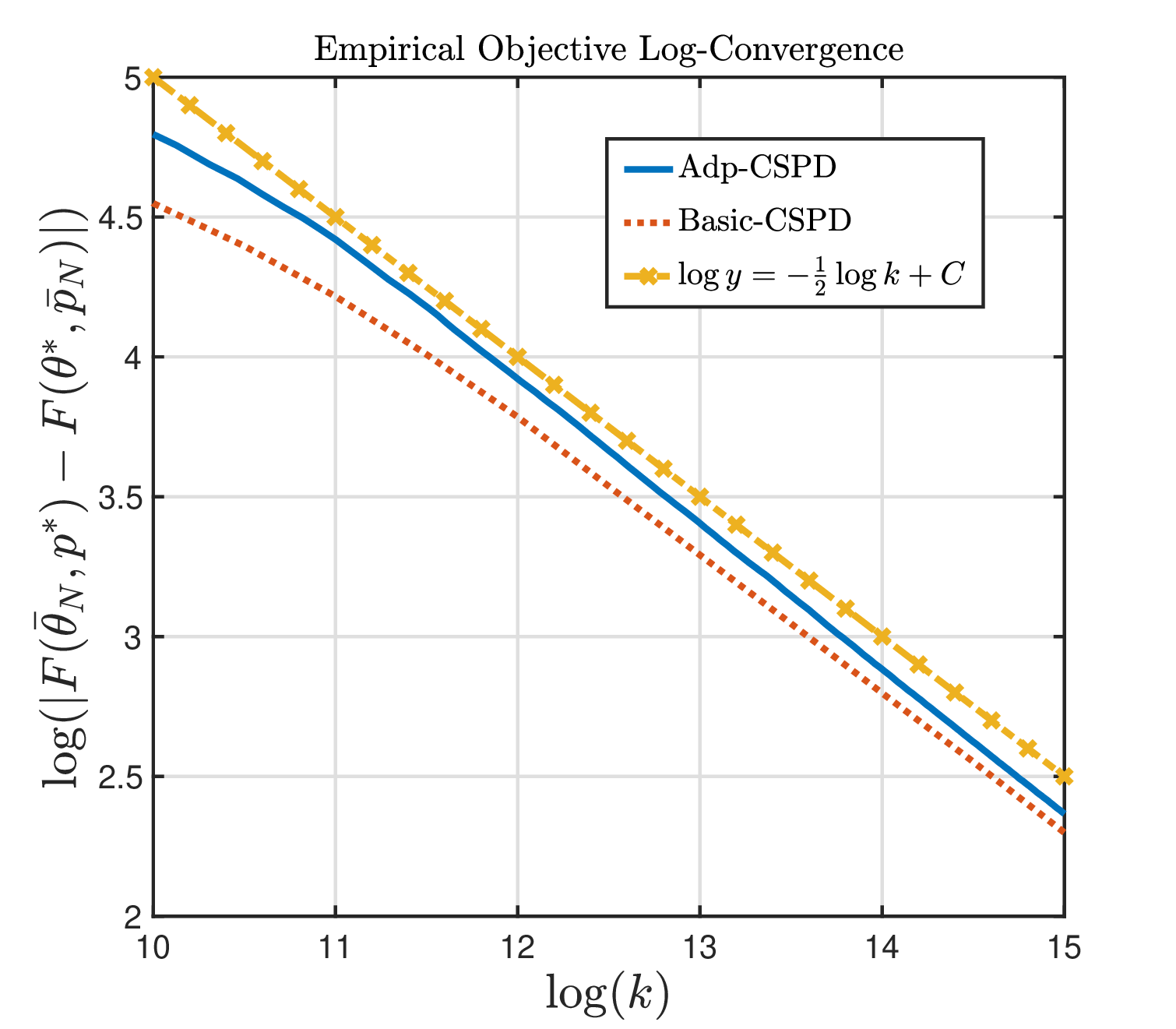}
    \caption{Empirical convergence rate of the objective gap  $ F(\bar \theta_N, p^*) - F(\theta^*,\bar p_N) $ for the robust optimal pricing under the normal design. }
    \label{fig:pricing_obj_NormalDesign}
\end{figure}

\begin{figure}[t]
    \centering
    \includegraphics[width=0.4\linewidth]{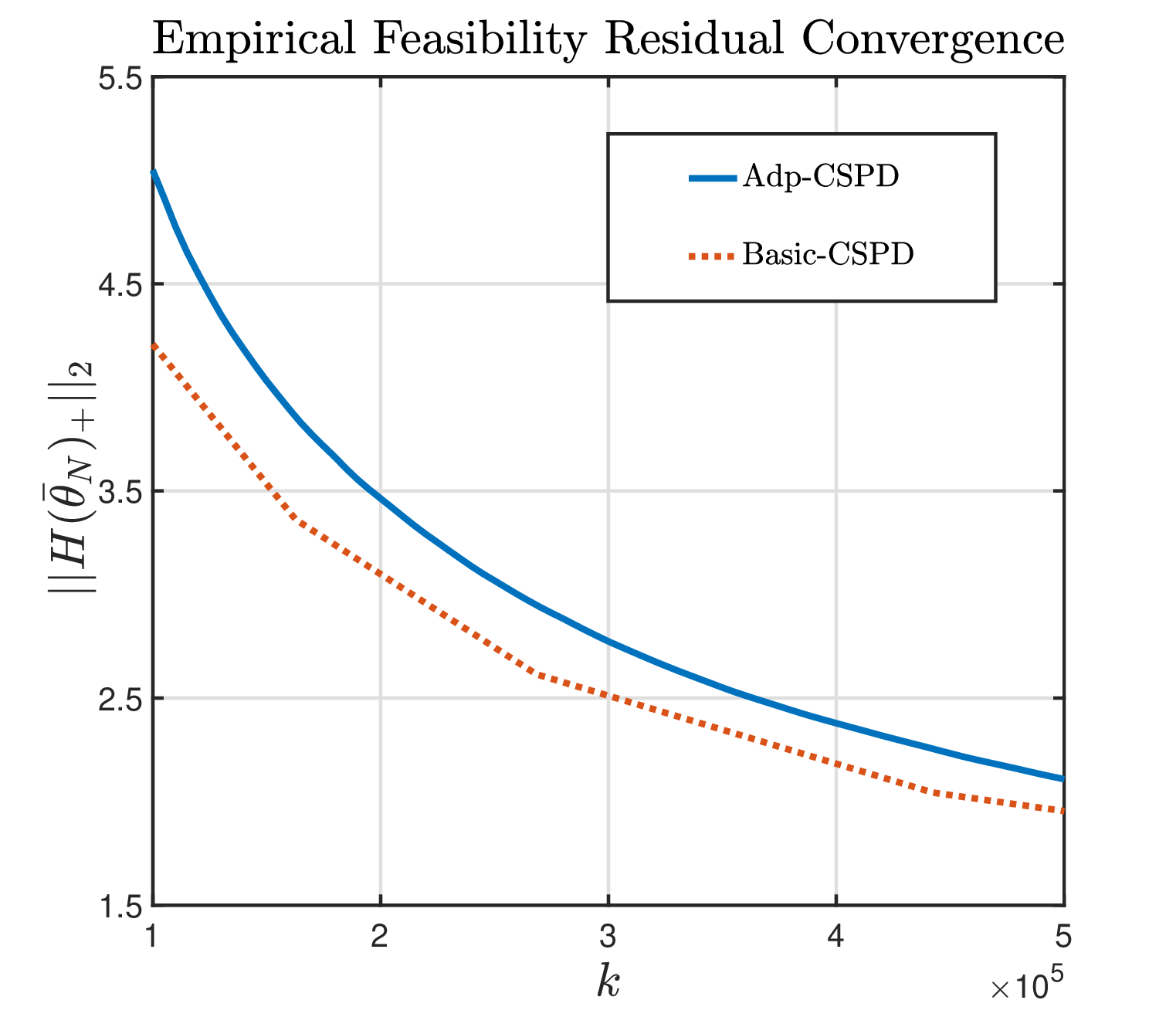}
    \includegraphics[width=0.4\linewidth]{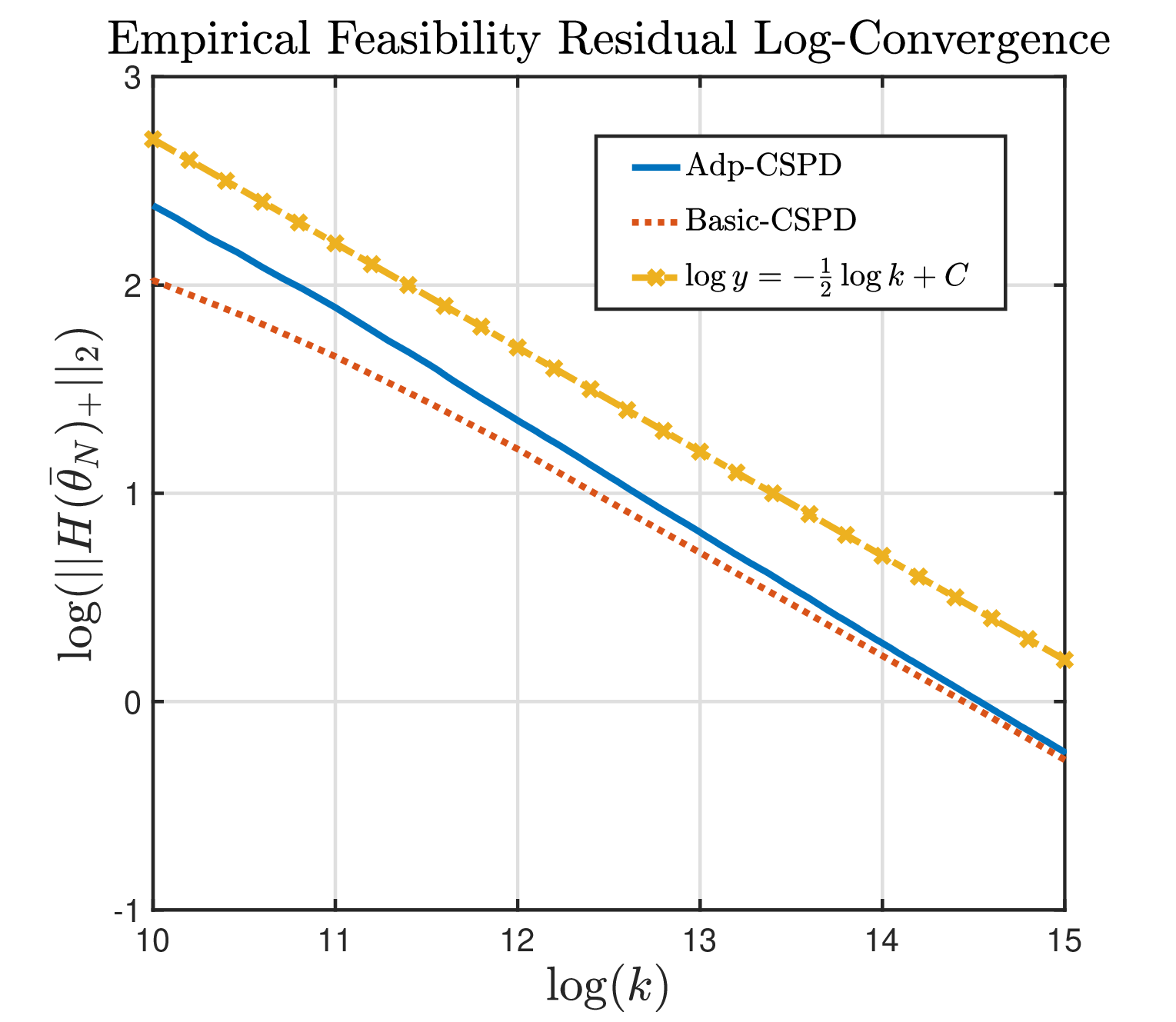}
    \caption{Empirical convergence rate of the feasibility residual $\| H(\bar \theta_N)_+ \|_2$ for robust optimal pricing Normal Design. }
    \label{fig:pricing_feasibility_NormalDesign}
\end{figure}
\begin{figure}[t]
    \centering
    \includegraphics[width=0.4\linewidth]{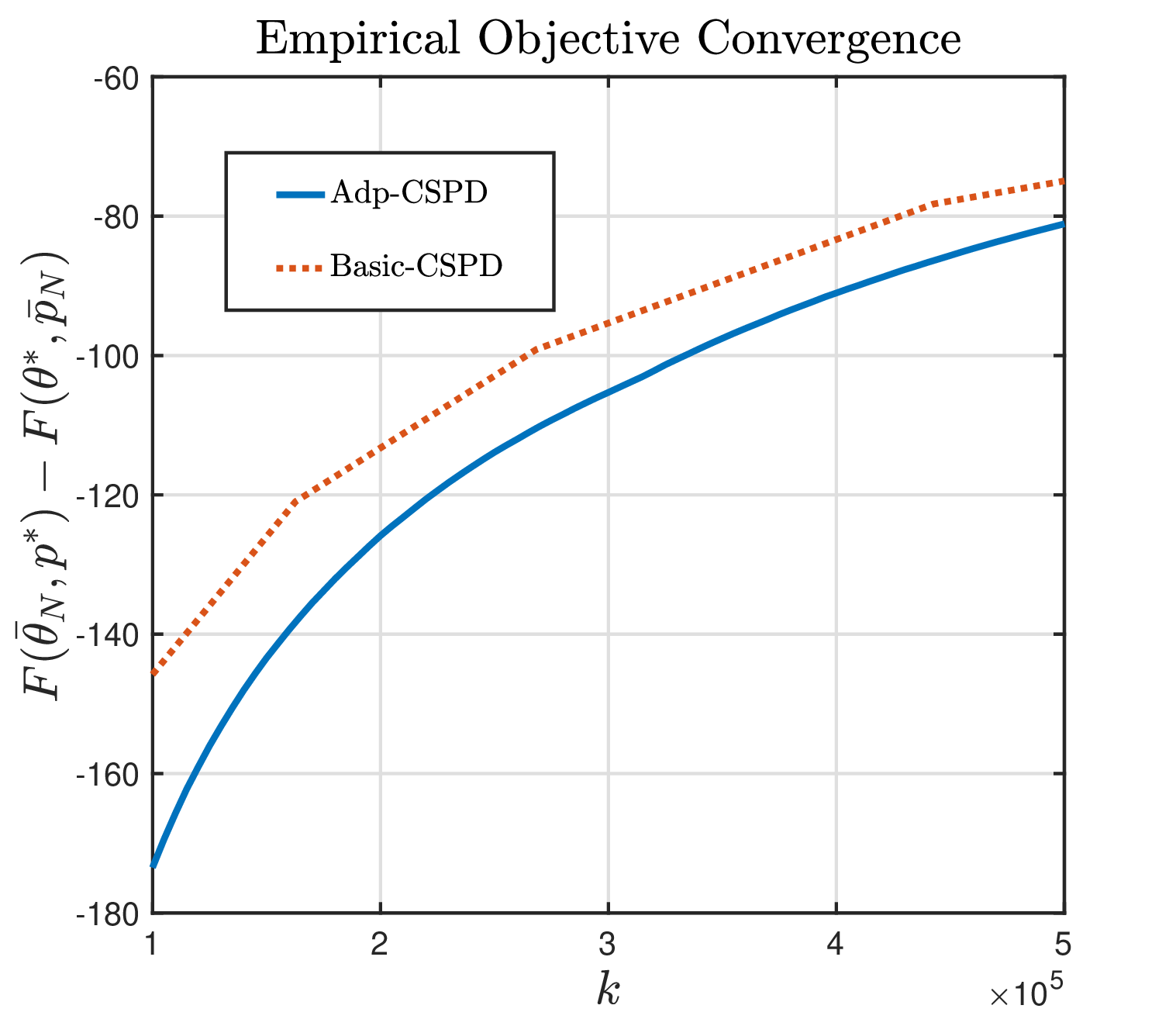}
    \includegraphics[width=0.4\linewidth]{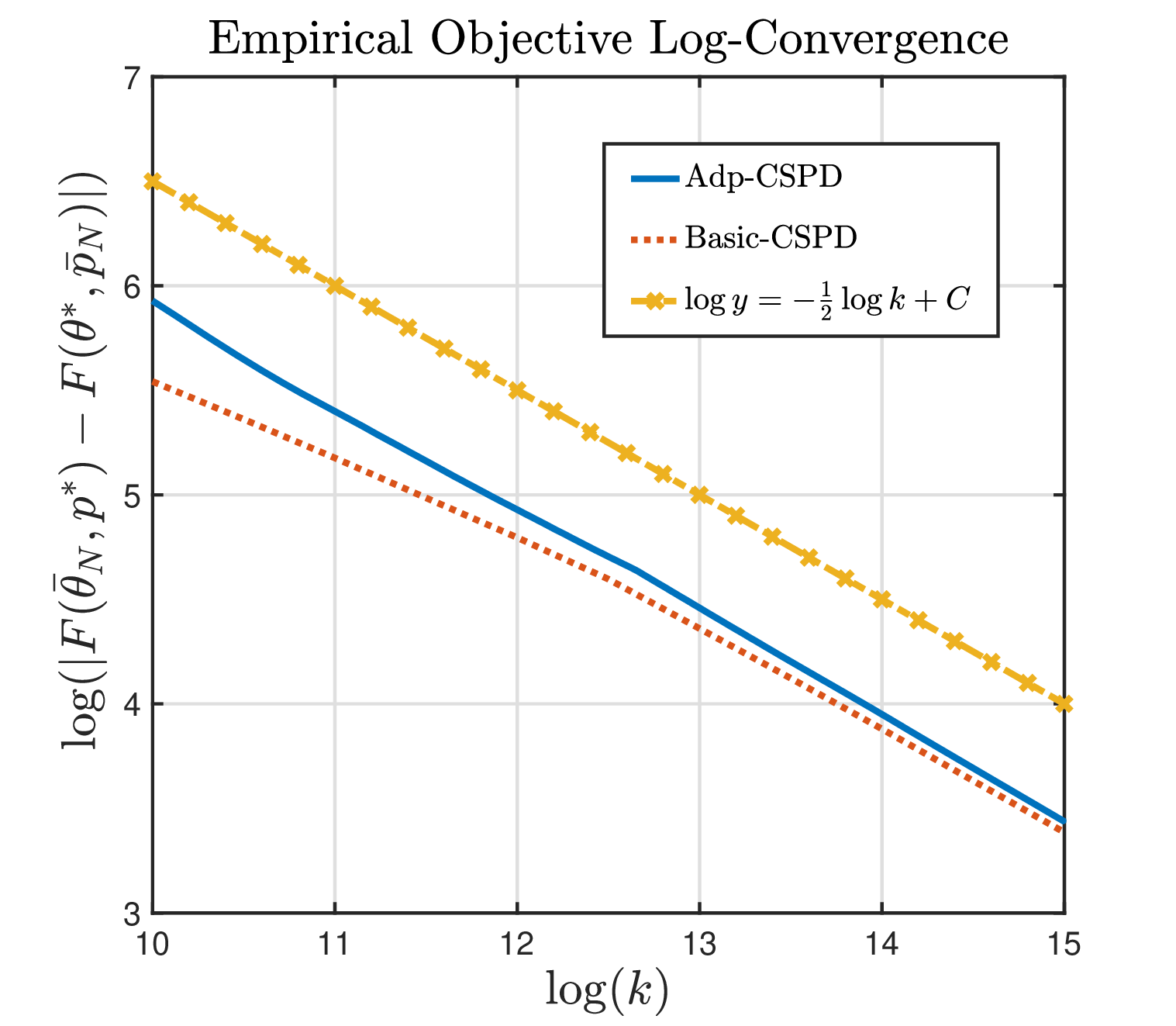}
    \caption{Empirical convergence rate of the objective gap  $ F(\bar \theta_N, p^*) - F(\theta^*,\bar p_N) $ for the robust optimal pricing under the Student design. }
    \label{fig:pricing_obj_StudentDesign}
\end{figure}
\begin{figure}[t]
    \centering
    \includegraphics[width=0.4\linewidth]{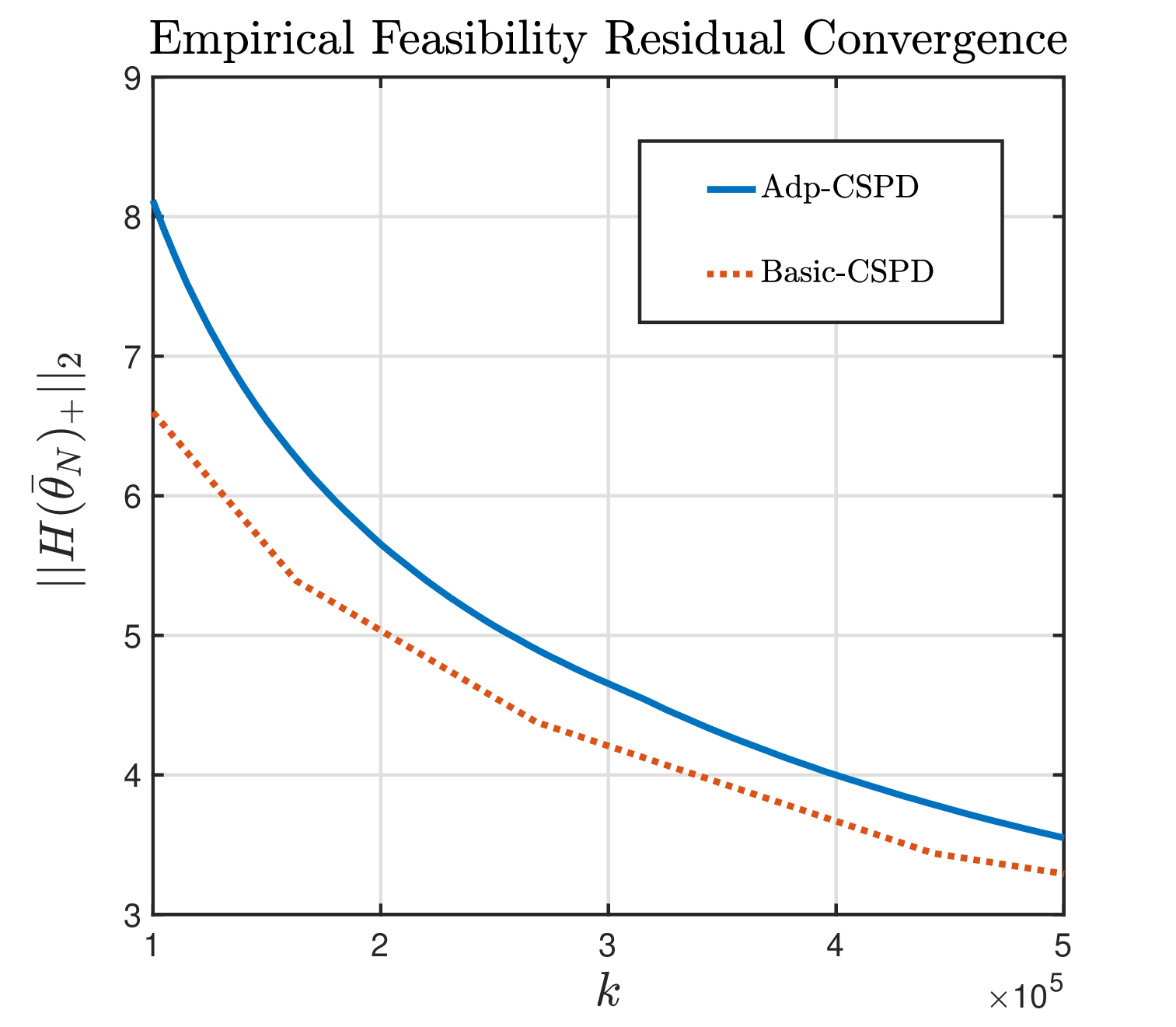}
    \includegraphics[width=0.4\linewidth]{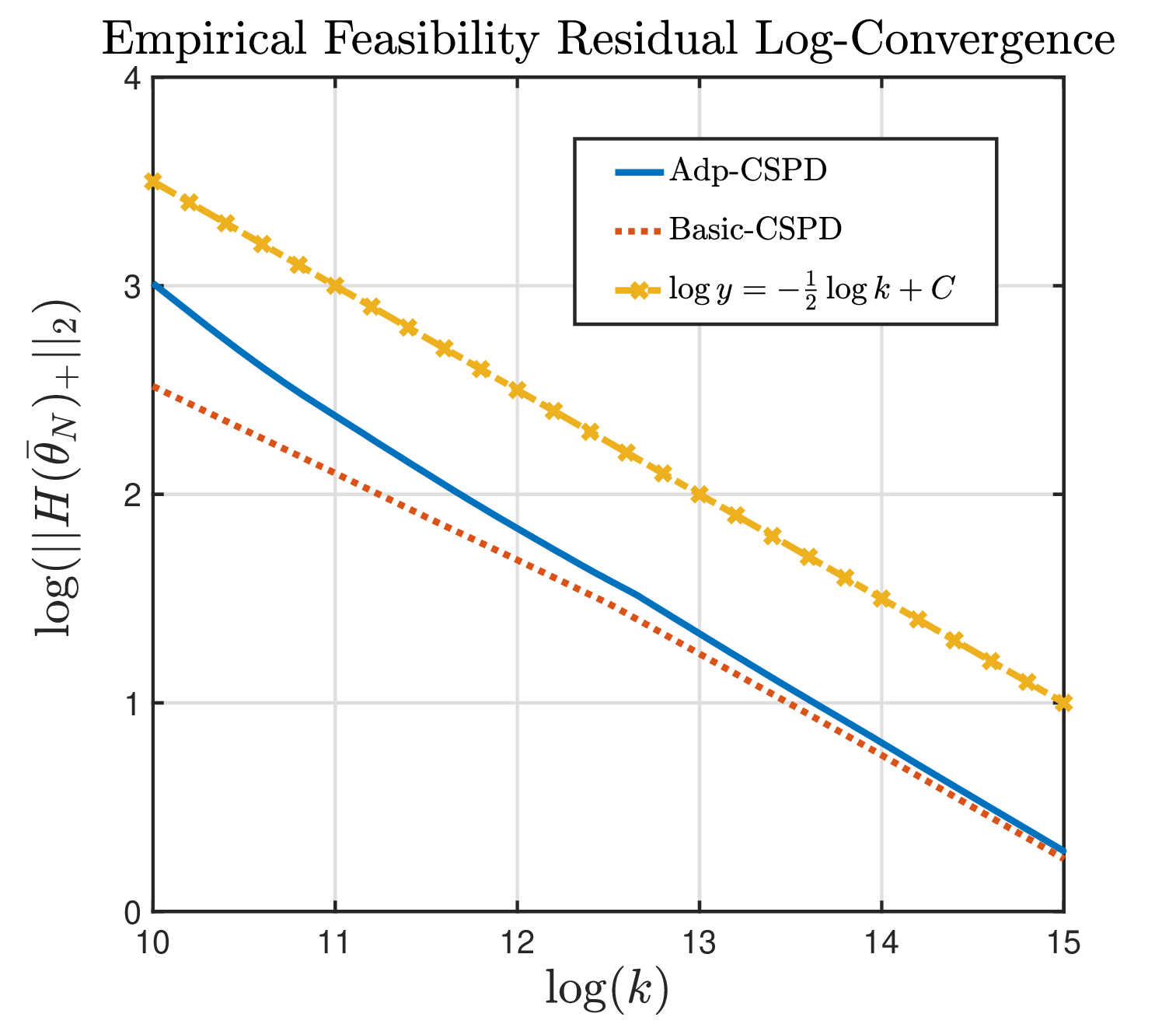}
    \caption{Empirical convergence rate of the feasibility residual $\| H(\bar \theta_N)_+ \|_2$ for the robust optimal pricing under the Student design. }
    \label{fig:pricing_feasibility_StudentDesign}
\end{figure}

Here, we conduct additional numerical experiments for the robust optimal pricing problem in Section~\ref{subsec:rop}. Specifically, we consider two settings where the features $\tilde s_i$ are generated using normal and student distributions in the following way.
\begin{itemize}
\item Normal Design: For each $i=1,\cdots,m$, each entry of the feature $\tilde s_i \in \RR^d$ is independently generated from a normal distribution that $\tilde s_{i,j} \sim \cN(2,0.5)$ for  $j = 1,\cdots, d$. Each entry of the feature $s \in \RR^d$ in the objective is also independently generated under normal distribution $\cN(2,0.5)$. 
\item Student Design: For each $i=1,\cdots,m$, each entry of the feature $\tilde s_i \in \RR^d$ is independently generated through $\tilde s_{i,j} = 2 + \nu_i$, where $\nu_i$ follows a heavy-tailed Student distribution $t_4$. Each entry of the feature $s \in \RR^d$ in the objective is also generated under Student distribution similarly.
\end{itemize}
The rest parts of the simulation environment are set the same as the uniform setting in Section~\ref{subsec:rop}. To solve these problems, we run Algorithms~\ref{alg:1} and \ref{alg:2} for 100 independent simulations, with the total number of iterations and stepsizes being the same as in Section~\ref{subsec:rop}. We report the numerical results for normal setup in Figures~\ref{fig:pricing_obj_NormalDesign} and \ref{fig:pricing_feasibility_NormalDesign}, and report the result for the student setup in Figures~\ref{fig:pricing_obj_StudentDesign} and \ref{fig:pricing_feasibility_StudentDesign}.


From these experiments, we observe that our Basic-CSPD and Adp-CSPD algorithms can efficiently solve the robust optimal pricing problem under various distribution settings. In addition, both the objective gap and feasibility residual converge to zero at the rate of $\cO(1/\sqrt{N})$, matching our theoretical convergence rate claims.

\newpage 
\bibliographystyle{plain}
\bibliography{SCGD}

\begin{thebibliography}{10}

\bibitem{BenTalElGhaouiNemirovski2009}
Aharon Ben-Tal, Laurent~El Ghaoui, and Arkadi Nemirovski.
\newblock {\em Robust Optimization}.
\newblock Princeton University Press, 2009.

\bibitem{ben2001approximate}
Aharon Ben-Tal and Arkadi Nemirovski.
\newblock On approximate robust counterparts of uncertain semidefinite and
  conic quadratic programs.
\newblock In {\em IFIP Conference on System Modeling and Optimization}, pages
  1--22. Springer, 2001.

\bibitem{ben2002robust}
Aharon Ben-Tal and Arkadi Nemirovski.
\newblock Robust optimization--methodology and applications.
\newblock {\em Mathematical Programming}, 92(3):453--480, 2002.

\bibitem{bertsekas99nonlinear}
D.~Bertsekas.
\newblock {\em Nonlinear Programming: 2nd Edition}.
\newblock Athena Scientific, Belmont, MA., 1999.

\bibitem{bertsimas2003robust}
Dimitris Bertsimas and Melvyn Sim.
\newblock Robust discrete optimization and network flows.
\newblock {\em Mathematical Programming}, 98(1):49--71, 2003.

\bibitem{bertsimas2004price}
Dimitris Bertsimas and Melvyn Sim.
\newblock The price of robustness.
\newblock {\em Operations Research}, 52(1):35--53, 2004.

\bibitem{boob2019stochastic}
Digvijay Boob, Qi~Deng, and Guanghui Lan.
\newblock Stochastic first-order methods for convex and nonconvex functional
  constrained optimization.
\newblock {\em Mathematical Programming}, 197(1):215--279, 2023.

\bibitem{chambolle2011first}
Antonin Chambolle and Thomas Pock.
\newblock A first-order primal-dual algorithm for convex problems with
  applications to imaging.
\newblock {\em Journal of Mathematical Imaging and Vision}, 40(1):120--145,
  2011.

\bibitem{chen2014optimal}
Yunmei Chen, Guanghui Lan, and Yuyuan Ouyang.
\newblock Optimal primal-dual methods for a class of saddle point problems.
\newblock {\em SIAM Journal on Optimization}, 24(4):1779--1814, 2014.

\bibitem{chen2017accelerated}
Yunmei Chen, Guanghui Lan, and Yuyuan Ouyang.
\newblock Accelerated schemes for a class of variational inequalities.
\newblock {\em Mathematical Programming}, 165(1):113--149, 2017.

\bibitem{danskin2012theory}
John~M Danskin.
\newblock {\em The Theory of Max-min and Its Application to Weapons Allocation
  Problems}, volume~5.
\newblock Springer Science \& Business Media, 2012.

\bibitem{dempe2002foundations}
Stephan Dempe.
\newblock {\em Foundations of Bilevel Programming}.
\newblock Springer Science \& Business Media, 2002.

\bibitem{dempe2013bilevel}
Stephan Dempe and Alain~B Zemkoho.
\newblock The bilevel programming problem: {R}eformulations, constraint
  qualifications and optimality conditions.
\newblock {\em Mathematical Programming}, 138(1):447--473, 2013.

\bibitem{di2020efficient}
Luca Di~Liello, Pierfrancesco Ardino, Jacopo Gobbi, Paolo Morettin, Stefano
  Teso, and Andrea Passerini.
\newblock Efficient generation of structured objects with constrained
  adversarial networks.
\newblock {\em Advances in Neural Information Processing Systems}, 33, 2020.

\bibitem{adultincome1996}
Dheeru Dua and Casey Graff.
\newblock {UCI} machine learning repository, 2017.

\bibitem{fan2014challenges}
Jianqing Fan, Fang Han, and Han Liu.
\newblock Challenges of big data analysis.
\newblock {\em National science review}, 1(2):293--314, 2014.

\bibitem{gallego2019revenue}
Guillermo Gallego and Huseyin Topaloglu.
\newblock {\em Revenue Management and Pricing Analytics}, volume 209.
\newblock Springer, 2019.

\bibitem{ghosh2021new}
Soham Ghosh, Mamta Sahare, and Sujay Mukhoti.
\newblock A new generalized newsvendor model with random demand and cost
  misspecification.
\newblock In {\em Strategic Management, Decision Theory, and Decision Science},
  pages 211--245. Springer, 2021.

\bibitem{goodfellow2014generative}
Ian Goodfellow, Jean Pouget-Abadie, Mehdi Mirza, Bing Xu, David Warde-Farley,
  Sherjil Ozair, Aaron Courville, and Yoshua Bengio.
\newblock Generative adversarial nets.
\newblock {\em Advances in Neural Information Processing Systems}, 27, 2014.

\bibitem{cvx}
Michael Grant and Stephen Boyd.
\newblock {CVX}: Matlab software for disciplined convex programming, version
  2.1.
\newblock \url{http://cvxr.com/cvx}, mar 2014.

\bibitem{hamedani2018primal}
Erfan~Yazdandoost Hamedani and Necdet~Serhat Aybat.
\newblock A primal-dual algorithm with line search for general convex-concave
  saddle point problems.
\newblock {\em SIAM Journal on Optimization}, 31(2):1299--1329, 2021.

\bibitem{hanley1982meaning}
James~A Hanley and Barbara~J McNeil.
\newblock The meaning and use of the area under a receiver operating
  characteristic (roc) curve.
\newblock {\em Radiology}, 143(1):29--36, 1982.

\bibitem{he2015mirror}
Niao He, Anatoli Juditsky, and Arkadi Nemirovski.
\newblock Mirror prox algorithm for multi-term composite minimization and
  semi-separable problems.
\newblock {\em Computational Optimization and Applications}, 61(2):275--319,
  2015.

\bibitem{heim2019constrained}
Eric Heim.
\newblock Constrained generative adversarial networks for interactive image
  generation.
\newblock In {\em Proceedings of the IEEE/CVF Conference on Computer Vision and
  Pattern Recognition}, pages 10753--10761, 2019.

\bibitem{jiang13solution}
H.~Jiang and U.~V. Shanbhag.
\newblock On the solution of stochastic optimization problems in imperfect
  information regimes.
\newblock In {\em Winter Simulation Conference (WSC)}, pages 821--832, Dec
  2013.

\bibitem{jiang2016solution}
H.~Jiang and U.~V Shanbhag.
\newblock On the solution of stochastic optimization and variational problems
  in imperfect information regimes.
\newblock {\em SIAM Journal on Optimization}, 26(4):2394--2429, 2016.

\bibitem{lan2020first}
Guanghui Lan.
\newblock {\em First-order and Stochastic Optimization Methods for Machine
  Learning}.
\newblock Springer Nature, 2020.

\bibitem{lan2013iteration}
Guanghui Lan and Renato~DC Monteiro.
\newblock Iteration-complexity of first-order penalty methods for convex
  programming.
\newblock {\em Mathematical Programming}, 138(1):115--139, 2013.

\bibitem{Lan2020ConditionalGM}
Guanghui Lan, E.~Romeijn, and Zhiqiang Zhou.
\newblock Conditional gradient methods for convex optimization with general
  affine and nonlinear constraints.
\newblock {\em SIAM Journal on Optimization}, 31(3):2307--2339, 2021.

\bibitem{lan2016algorithms}
Guanghui Lan and Zhiqiang Zhou.
\newblock Algorithms for stochastic optimization with functional or expectation
  constraints.
\newblock {\em Computational Optimization and Applications}, 76:461--498, 2020.

\bibitem{lemarechal1995new}
Claude Lemar{\'e}chal, Arkadii Nemirovskii, and Yurii Nesterov.
\newblock New variants of bundle methods.
\newblock {\em Mathematical Programming}, 69(1):111--147, 1995.

\bibitem{lin2020data}
Qihang Lin, Selvaprabu Nadarajah, Negar Soheili, and Tianbao Yang.
\newblock A data efficient and feasible level set method for stochastic convex
  optimization with expectation constraints.
\newblock {\em Journal of Machine Learning Research}, 2020.

\bibitem{loridan1996weak}
Pierre Loridan and Jacqueline Morgan.
\newblock Weak via strong {S}tackelberg problem: {N}ew results.
\newblock {\em Journal of Global Optimization}, 8(3):263--287, 1996.

\bibitem{mitsos2008global}
Alexander Mitsos, Panayiotis Lemonidis, and Paul~I Barton.
\newblock Global solution of bilevel programs with a nonconvex inner program.
\newblock {\em Journal of Global Optimization}, 42(4):475--513, 2008.

\bibitem{myerson1997game}
Roger~B Myerson.
\newblock {\em Game Theory: Analysis of Conflict}.
\newblock Harvard University Press, 1997.

\bibitem{nemirovski1995information}
Arkadi Nemirovski.
\newblock Information-based complexity of convex programming.
\newblock {\em Lecture notes}, 1994.

\bibitem{Nemirovski2005}
Arkadi Nemirovski.
\newblock Prox-method with rate of convergence $o(1/t)$ for variational
  inequalities with {L}ipschitz continuous monotone operators and smooth
  convex-concave saddle point problems.
\newblock {\em SIAM Journal on Optimization}, page 229–251, January 2005.

\bibitem{nemirovski2009robust}
Arkadi Nemirovski, Anatoli Juditsky, Guanghui Lan, and Alexander Shapiro.
\newblock Robust stochastic approximation approach to stochastic programming.
\newblock {\em SIAM Journal on Optimization}, 19(4):1574--1609, 2009.

\bibitem{Nemirovsky83}
A.S.\ Nemirovsky and D.B.\ Yudin.
\newblock {\em Problem Complexity and Method Efficiency in Optimization}.
\newblock John Wiley \& Sons, Ltd., Great Britain, 1983.

\bibitem{nesterov2005smooth}
Yu~Nesterov.
\newblock Smooth minimization of non-smooth functions.
\newblock {\em Mathematical Programming}, 103(1):127--152, 2005.

\bibitem{nesterov1998introductory}
Yurii Nesterov.
\newblock {\em Introductory Lectures on Convex Programming volume i: Basic
  course}.
\newblock 1998.

\bibitem{nouiehed2019solving}
Maher Nouiehed, Maziar Sanjabi, Tianjian Huang, Jason~D Lee, and Meisam
  Razaviyayn.
\newblock Solving a class of non-convex min-max games using iterative first
  order methods.
\newblock {\em Advances in Neural Information Processing Systems}, 32, 2019.

\bibitem{oliveira2022sample}
Roberto~I Oliveira and Philip Thompson.
\newblock Sample average approximation with heavier tails i: {N}on-asymptotic
  bounds with weak assumptions and stochastic constraints.
\newblock {\em Mathematical Programming}, pages 1--48, 2022.

\bibitem{oliveira2017sample}
Roberto~I Oliveira and Philip Thompson.
\newblock Sample average approximation with heavier tails ii: {L}ocalization in
  stochastic convex optimization and persistence results for the {L}asso.
\newblock {\em Mathematical Programming}, 199(1-2):49--86, 2023.

\bibitem{outrata1988note}
Ji{\v{r}}{\'\i}~V Outrata.
\newblock A note on the usage of nondifferentiable exact penalties in some
  special optimization problems.
\newblock {\em Kybernetika}, 24(4):251--258, 1988.

\bibitem{ouyang2021lower}
Yuyuan Ouyang and Yangyang Xu.
\newblock Lower complexity bounds of first-order methods for convex-concave
  bilinear saddle-point problems.
\newblock {\em Mathematical Programming}, 185(1-2):1--35, 2021.

\bibitem{pearsall1976lagrange}
Edward~S Pearsall.
\newblock A {L}agrange multiplier method for certain constrained min-max
  problems.
\newblock {\em Operations Research}, 24(1):70--91, 1976.

\bibitem{shapiro2013sample}
Alexander Shapiro.
\newblock Sample average approximation.
\newblock {\em Encyclopedia of Operations Research and Management Science},
  3:1350--1355, 2013.

\bibitem{soland1973optimal}
Richard~M Soland.
\newblock Optimal defensive missile allocation: A discrete min-max problem.
\newblock {\em Operations Research}, 21(2):590--596, 1973.

\bibitem{von2007theory}
John Von~Neumann and Oskar Morgenstern.
\newblock {\em Theory of Games and Economic Behavior}.
\newblock Princeton University Press, 2007.

\bibitem{wiesemann2013pessimistic}
Wolfram Wiesemann, Angelos Tsoukalas, Polyxeni-Margarita Kleniati, and
  Ber{\c{c}} Rustem.
\newblock Pessimistic bilevel optimization.
\newblock {\em SIAM Journal on Optimization}, 23(1):353--380, 2013.

\bibitem{ying2016stochastic}
Yiming Ying, Longyin Wen, and Siwei Lyu.
\newblock Stochastic online {AUC} maximization.
\newblock {\em Advances in neural information processing systems}, 29, 2016.

\bibitem{yu2017online}
Hao Yu, Michael Neely, and Xiaohan Wei.
\newblock Online convex optimization with stochastic constraints.
\newblock {\em Advances in Neural Information Processing Systems}, 30, 2017.

\bibitem{zafar2019fairness}
Muhammad~Bilal Zafar, Isabel Valera, Manuel Gomez-Rodriguez, and Krishna~P
  Gummadi.
\newblock Fairness constraints: A flexible approach for fair classification.
\newblock {\em The Journal of Machine Learning Research}, 20(1):2737--2778,
  2019.

\bibitem{zhang2021stochastic}
Liwei Zhang, Yule Zhang, Xiantao Xiao, and Jia Wu.
\newblock Stochastic approximation proximal method of multipliers for convex
  stochastic programming.
\newblock {\em Mathematics of Operations Research}, 48(1):177--193, 2023.

\bibitem{zhang2020optimal}
Zhe Zhang and Guanghui Lan.
\newblock Optimal algorithms for convex nested stochastic composite
  optimization.
\newblock {\em arXiv preprint arXiv:2011.10076}, 2020.

\bibitem{zhang2022solving}
Zhe Zhang and Guanghui Lan.
\newblock Solving convex smooth function constrained optimization is as almost
  easy as unconstrained optimization.
\newblock {\em arXiv preprint arXiv:2210.05807}, 2022.

\end{thebibliography}
\end{document}